\pdfoutput=1
\documentclass{article}

\PassOptionsToPackage{round}{natbib}
\usepackage[main, final]{neurips_2026}
\usepackage{times}

\usepackage[utf8]{inputenc}
\usepackage[T1]{fontenc}
\usepackage{microtype}

\usepackage{amsmath,amssymb,amsthm,amsfonts}

\usepackage{graphicx}
\usepackage{xcolor}
\usepackage{tikz}
\usetikzlibrary{arrows.meta,positioning,patterns,calc}
\usepackage{pgfplots}
\pgfplotsset{compat=1.18}

\usepackage{booktabs}
\usepackage{colortbl}
\usepackage{array}

\usepackage{enumitem}
\usepackage{needspace}
\usepackage{placeins}

\usepackage{url}
\usepackage{hyperref}
\hypersetup{hidelinks}

\usepackage[most]{tcolorbox}
\usepackage{pifont}
\newcommand{\cmark}{\ding{51}}
\newcommand{\xmark}{\ding{55}}

\definecolor{regimeHelpful}{HTML}{DDEEE9}
\definecolor{regimeNoise}{HTML}{F5DDDA}
\definecolor{rowAlt}{HTML}{F6F7F5}
\definecolor{hiBad}{HTML}{B85C55}
\definecolor{hiWarn}{HTML}{D3922D}
\definecolor{cellGood}{HTML}{D7EEDF}
\definecolor{cellGoodL}{HTML}{EBF6EF}
\definecolor{cellBad}{HTML}{EFCFCB}
\definecolor{cellBadL}{HTML}{F7E4E2}

\newcommand{\vneg}[1]{\textcolor{hiBad}{\textbf{#1}}}
\newcommand{\vwarn}[1]{\textcolor{hiWarn}{\textbf{#1}}}
\newcommand{\cellg}[1]{\cellcolor{cellGoodL}#1}
\newcommand{\cellR}[1]{\cellcolor{cellBad}#1}
\newcommand{\cellr}[1]{\cellcolor{cellBadL}#1}
\newcommand{\cellbest}[1]{\cellcolor{cellGood}\textbf{#1}}
\newcommand{\cellsec}[1]{\cellcolor{cellGoodL}\underline{#1}}
\newcommand{\cellder}[1]{\cellcolor{cellGoodL}\textit{#1}}

\newtcolorbox{resultbox}{
  colback=black!3,
  colframe=black!45,
  boxrule=0.45pt,
  arc=1pt,
  left=4pt, right=4pt, top=3pt, bottom=3pt,
  before skip=6pt, after skip=6pt
}
\newtcolorbox[auto counter,number within=section]{casebox}[2][]{%
  colback=gray!4, colframe=gray!50, fonttitle=\small\bfseries,
  title=Case~\thetcbcounter: #2, #1,
  boxrule=0.4pt, arc=1.5pt,
  left=4pt, right=4pt, top=2pt, bottom=2pt,
  before upper={\small}
}

\usepackage{titletoc}

\makeatletter
  \def\ttl@Hy@steplink#1{%
    \Hy@MakeCurrentHrefAuto{#1*}%
    \edef\ttl@Hy@saveanchor{%
      \noexpand\Hy@raisedlink{%
        \noexpand\hyper@anchorstart{\@currentHref}%
        \noexpand\hyper@anchorend
        \def\noexpand\ttl@Hy@SavedCurrentHref{\@currentHref}%
        \noexpand\ttl@Hy@PatchSaveWrite
      }%
    }%
  }%
  \def\ttl@Hy@PatchSaveWrite{%
    \begingroup
      \toks@\expandafter{\ttl@savewrite}%
      \edef\x{\endgroup
        \def\noexpand\ttl@savewrite{%
          \let\noexpand\@currentHref\noexpand\ttl@Hy@SavedCurrentHref
          \the\toks@
        }%
      }%
    \x
  }%
  \def\ttl@Hy@refstepcounter#1{%
    \let\ttl@b\Hy@raisedlink
    \def\Hy@raisedlink##1{%
      \def\ttl@Hy@saveanchor{\Hy@raisedlink{##1}}%
    }%
    \refstepcounter{#1}%
    \let\Hy@raisedlink\ttl@b
  }%
  \def\ttl@gobblecontents#1#2#3#4{\ignorespaces}%
\makeatother

\newcommand\DoAppendixContents{%
  \startcontents[appendix]
  \printcontents[appendix]{}{1}{%
    \section*{Appendix Contents}
    \vskip2pt\hrule\vskip4pt
  }%
  \vskip2pt\hrule\vskip8pt
}
\title{Bayes-Sufficient Compression Is Not Enough: \\
How Does Communication Help Multi-Agent Systems? }

\author{%
  Yi Xie$^{1}$\quad Zhanke Zhou$^{2}$\quad Yi Fan$^{3}$\footnotemark[3]\quad Yong Ge$^{4}$\quad Bo Han$^{2}$\quad Bo Liu$^{1}$\footnotemark[2] \\
  $^{1}$Department of Electrical \& Computer Engineering, University of Arizona \\
  $^{2}$TMLR Group, Department of Computer Science, Hong Kong Baptist University \\
  $^{3}$Amazon Web Services \\
  $^{4}$Department of Management Information Systems, \\
  Eller College of Management, the University of Arizona \\
  \texttt{yix@arizona.edu}, \texttt{cszkzhou@comp.hkbu.edu.hk}, \texttt{fnyi@amazon.com}, \\
  \texttt{yongge@arizona.edu}, \texttt{bhanml@comp.hkbu.edu.hk}, \texttt{boliu@arizona.edu} \\
}
\newtheorem{theorem}{Theorem}
\newtheorem{proposition}[theorem]{Proposition}
\newtheorem{definition}{Definition}[section]
\newtheorem{lemma}[definition]{Lemma}
\newtheorem{corollary}[definition]{Corollary}

\newtheorem{remark}[definition]{Remark}

\newcommand{\E}{\mathbb{E}}
\newcommand{\argmax}{\mathop{\mathrm{argmax}}}

\newcommand{\safeincludegraphics}[2][]{\includegraphics[#1]{#2}}

\newcommand{\coderepo}{\url{https://github.com/Yydc/Communication}}

\begin{document}
\raggedbottom

\maketitle
{\renewcommand{\thefootnote}{\fnsymbol{footnote}}%
\footnotetext[2]{Corresponding author.}%
\footnotetext[3]{This work is not related to the author's position at Amazon Inc.}}
\suppressfloats[t]

\begin{abstract}
Multi-agent LLM systems pair a sender with broad context and an executor with a limited local view. We study when a short message improves the executor's next decision, when raw context is preferable, and when a stronger sender helps. Our framework, \emph{receiver-relative bounded coordination}, expresses message utility as receiver gain minus protocol tax. Compression beats raw context when tax savings exceed losses from omitted information and decoder mismatch. Even \emph{Bayes-sufficient} compression can fail when a bounded executor cannot use its surface form. A three-stage decomposition separates externalization, absorption, and \emph{action closure}, explaining how errors remain after the correct content reaches the receiver. Under a single-crossing condition, sender upgrades help above a receiver-burden threshold. Across six benchmarks, the same Qwen protocol raises ContextBench joint accuracy from $0.633$ to $0.775$ but lowers ToolSandbox from $0.889$ to $0.653$. Fixed-message replay reveals closure failures despite correct artifact recovery. These results guide an inference-time selector that improves the accuracy-cost frontier on the evaluated communication regimes. Evaluation artifacts: \coderepo.

\end{abstract}

\section{Introduction}
\label{sec:intro}

Multi-agent LLM systems increasingly split work across agents with
asymmetric contexts, capabilities, and budgets~\citep{yi2025debate}. In coding, web, and
tool-use workflows, a main agent may inspect repository state, prior
failures, tool traces, or verifier outputs~\citep{dong2025survey}. The executor, by contrast,
must edit one file, choose one tool, or execute the next action under
tighter context and latency limits \citep{openai2025agents,
openai2025dev,anthropic2025research,anthropic2025harness}. This split
makes communication a central protocol decision. Exposing the full
context forces the executor to rediscover relevance within budget; a
short message can expose the missing target directly, but may also
omit useful information or present it in a form the executor cannot
easily use~\citep{shi2026intrinsic}.

\begin{figure}[t]
\centering
\begin{minipage}[c]{0.56\linewidth}
\centering
\includegraphics[width=\linewidth]{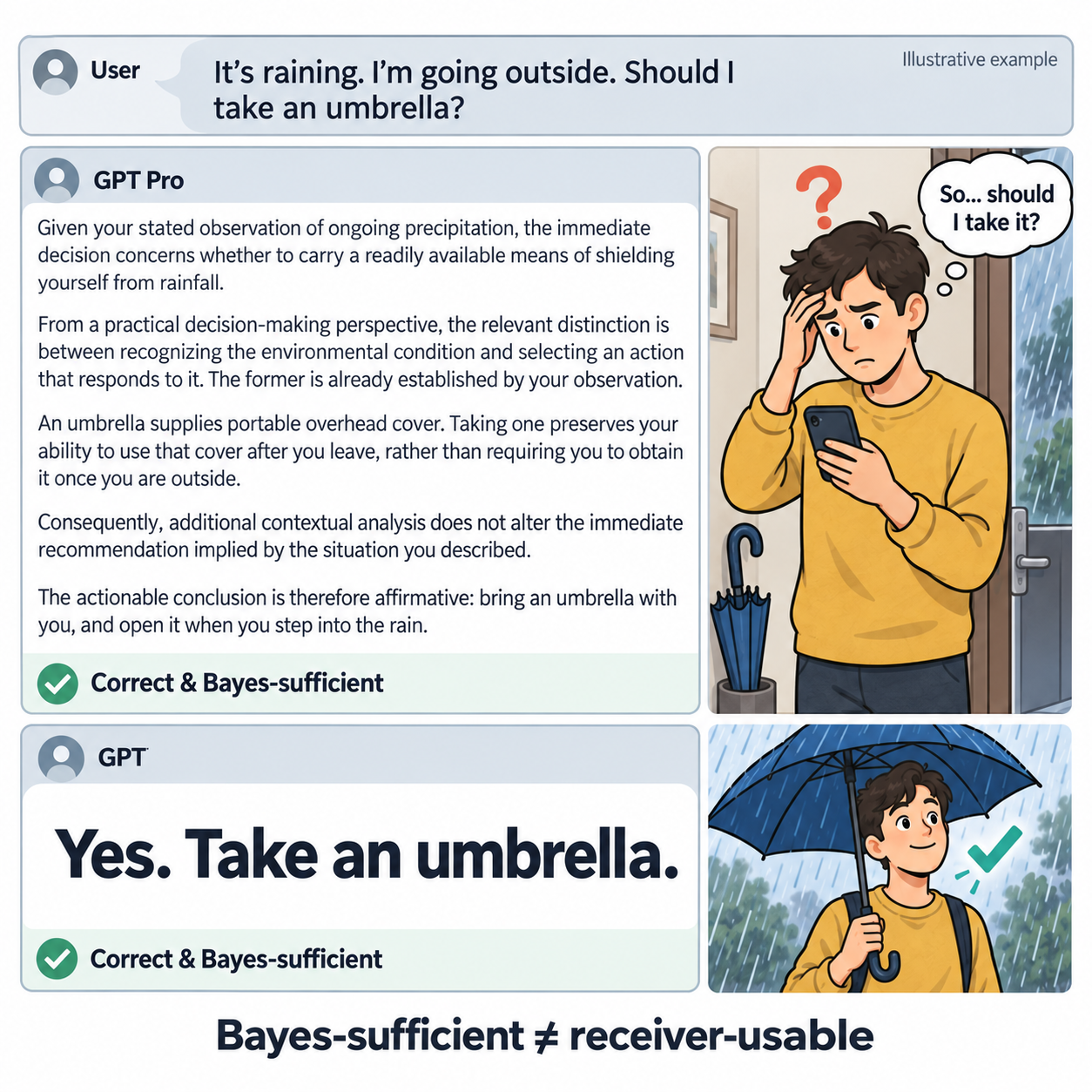}
\end{minipage}\hfill
\begin{minipage}[c]{0.41\linewidth}
\setlength{\abovecaptionskip}{0pt}
\caption[Bayes-sufficient is not the same as receiver-usable.]{\raggedright\textbf{Bayes-sufficient is not the same as receiver-usable.}
Illustrative example, constructed for exposition rather than taken from model logs.
Both replies are correct and Bayes-sufficient: each carries the information that determines the right action.
The verbose reply leaves the user unsure what to do, whereas the short reply leads directly to the action.
The framework attributes this difference to the receiver's bounded decoder and the protocol tax of the message (Section~\ref{sec:theory}); Theorem~\ref{thm:sufficiency-not-enough} shows that Bayes sufficiency alone does not make a message receiver-optimal.}
\label{fig:teaser}
\end{minipage}
\end{figure}

Recent evidence suggests that multi-agent LLM performance depends less
on the number of agents than on how agents coordinate~\citep{anthropic2025harness}.
Matched-budget studies report coordination tax and topology-dependent
error amplification; failure analyses trace breakdowns to specification
gaps, inter-agent misalignment, and weak verification; production
harnesses depend heavily on routing, aggregation, and context control
\citep{kim2025towards,cemri2025mast,anthropic2025research}. These observations raise a step-level question: \textit{when does a message improve the executor's next decision rather than adding
burden?} Aggregate task metrics conflate these effects. A final failure may come from an uninformative message, a useful message the receiver fails to use, or an error carried across steps. Protocol design therefore needs to evaluate each message at the receiver's next decision.

We study this question through \emph{receiver-relative bounded
coordination}. The main agent, or sender, has a broader context and sends
a bounded message to the executor, or receiver, which acts from a
limited local view. Three commitments distinguish this framework from
aggregate analyses: the step is the unit of analysis, the
\emph{action-sufficient target} rather than total information defines what is useful, and the executor is modeled as a bounded
decoder, so an information-sufficient message need not be usable.
Message value decomposes into the executor's next-step gain minus the protocol tax induced. 

The framework connects four aspects of communication. Compression trades tax savings against omitted information and decoder mismatch. Bayes sufficiency preserves information, while receiver compatibility determines whether the executor can use it (Figure~\ref{fig:teaser}). Fixed-message replay separates content recovery from \emph{action closure}, the conversion of recovered content into an emitted action. Finally, under single crossing, a stronger sender becomes useful above a receiver-burden threshold. 

We tested these relationships with paired receiver-only, full-context, and multi-agent runs; replay with a fixed sender message; and sender swaps. The six benchmarks cover function calling, stateful tools, code context, retrieval, software repair, and customer service. Communication helped when the local view lacked useful context, while its effect varied across receiver interfaces on locally informative tasks. These observations also guide a selector that chooses communication actions by their estimated gain, protocol burden, and cost.

\textbf{Contributions.}
\begin{itemize}[leftmargin=1.5em,topsep=2pt,itemsep=2pt]
\item We introduce a step-level theory of receiver-relative bounded coordination, with formal results on representation-aware compression, information sufficiency versus executor usability, post-message action-closure loss, and sender-upgrade burden thresholds (Section~\ref{sec:theory}).

\item We develop an inference-time protocol selector that scores communication actions using the same gain, tax, decoder-mismatch, sender-upgrade, and residual-risk quantities (Section~\ref{sec:implications}).

\item We provide empirical tests across six benchmark families, showing compression sign flips, receiver-side action-closure failures, sender-upgrade thresholds, and accuracy-cost improvements from adaptive protocol selection (Section~\ref{sec:experiments}).
\end{itemize}
\section{A Theoretical Framework for Bounded Coordination}
\label{sec:theory}

We formalize bounded communication at the receiver's next decision. The framework is \emph{receiver-relative}: communication is useful only to the extent that it improves the deployed receiver's next-step decision. The same quantities are extended to a discrete-action control rule in Section~\ref{sec:implications}. Full proofs and theoretical supplements are in Appendix~\ref{app:theory}.

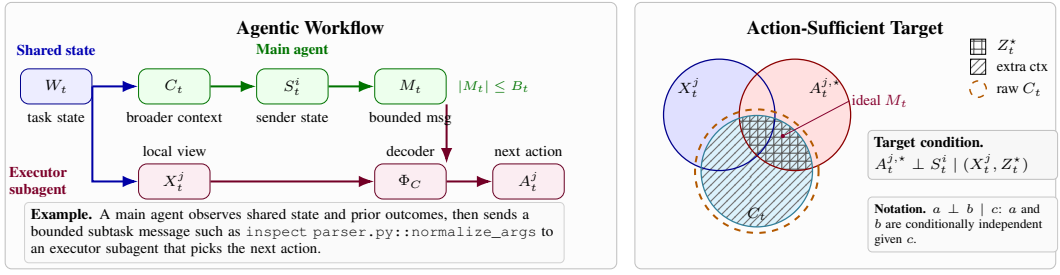
\begin{figure}[t]
\centering
\resizebox{\linewidth}{!}{%
\begin{tikzpicture}[font=\scriptsize, >=Latex]
\tikzset{
  panel/.style={draw=black!25, rounded corners=3pt, fill=black!1},
  head/.style={font=\bfseries\small},
  group/.style={font=\bfseries\scriptsize},
  statebox/.style={draw=blue!55!black, fill=blue!7, rounded corners=4pt,
                   minimum width=1.18cm, minimum height=0.58cm, inner sep=2pt},
  senderbox/.style={draw=green!45!black, fill=green!7, rounded corners=4pt,
                    minimum width=1.18cm, minimum height=0.58cm, inner sep=2pt},
  recvbox/.style={draw=purple!55!black, fill=purple!7, rounded corners=4pt,
                  minimum width=1.18cm, minimum height=0.58cm, inner sep=2pt},
  note/.style={align=center, text width=1.65cm, inner sep=1pt, font=\scriptsize},
  chip/.style={draw=black!22, rounded corners=2pt, fill=black!3,
               inner sep=2.5pt, align=left},
  casebox/.style={draw=black!18, rounded corners=2pt, fill=black!2,
                  inner sep=3pt, align=left},
  legend/.style={font=\scriptsize, align=left, inner sep=1pt}
}
\begin{scope}[shift={(0,0)}]
\node[panel, minimum width=10.10cm, minimum height=4.40cm,
      anchor=south west] (L) at (0,0) {};

\node[anchor=north, head] at (5.05,4.22)
  {Agentic Workflow};
\node[group, text=blue!70!black]  at (0.85,3.62) {Shared state};
\node[group, text=green!45!black] at (4.75,3.62) {Main agent};
\node[group, text=purple!65!black, anchor=center, align=center]
  at (0.55,1.45) {Executor\\subagent};
\node[statebox]  (W)   at (0.85,3.02) {$W_t$};

\node[senderbox] (C)   at (2.80,3.02) {$C_t$};
\node[senderbox] (S)   at (4.75,3.02) {$S_t^i$};
\node[senderbox] (M)   at (6.70,3.02) {$M_t$};

\node[recvbox]   (X)   at (2.80,1.45) {$X_t^j$};
\node[recvbox]   (Phi) at (6.70,1.45) {$\Phi_C$};
\node[recvbox]   (A)   at (8.65,1.45) {$A_t^j$};
\draw[->, line width=0.95pt, blue!65!black]   (W) -- (C);                 %
\draw[->, line width=0.95pt, blue!65!black]   (W.east) |- (X.west);       %

\draw[->, line width=0.95pt, green!45!black]  (C) -- (S);                 %
\draw[->, line width=0.95pt, green!45!black]  (S) -- (M);                 %

\draw[->, line width=0.95pt, purple!60!black] (M.south east) -- (Phi.north east); %
\draw[->, line width=0.95pt, purple!60!black] (X) -- (Phi);                       %
\draw[->, line width=0.95pt, purple!60!black] (Phi) -- (A);                       %
\node[note, anchor=north] at ($(W.south)+(0,-0.08)$) {task state};
\node[note, anchor=north] at ($(C.south)+(0,-0.08)$) {broader context};
\node[note, anchor=north] at ($(S.south)+(0,-0.08)$) {sender state};
\node[note, anchor=north] at ($(M.south)+(0,-0.08)$) {bounded msg};
\node[font=\tiny, anchor=west, color=green!45!black]
  at ($(M.east)+(0.06, 0)$) {$|M_t|\le B_t$};
\node[note, anchor=south] at ($(X.north)+(0,0.08)$)   {local view};
\node[note, anchor=south] at ($(Phi.north)+(0,0.08)$) {decoder};
\node[note, anchor=south] at ($(A.north)+(0,0.08)$)   {next action};
\node[casebox, anchor=south west, text width=8.70cm] at (0.32,0.10)
  {\textbf{Example.} A main agent observes shared state and prior outcomes, then sends a bounded subtask message such as \texttt{inspect parser.py::normalize\_args} to an executor subagent that picks the next action.};
\end{scope}
\begin{scope}[shift={(10.40,0)}]
\node[panel, minimum width=6.95cm, minimum height=4.40cm,
      anchor=south west] (R) at (0,0) {};

\node[anchor=north, head] at (3.48,4.22)
  {Action-Sufficient Target};
\def\r{0.92}
\coordinate (Xv) at (1.40,2.55);
\coordinate (Av) at (2.65,2.55);
\coordinate (Cv) at (2.02,1.62);

\fill[blue!12] (Xv) circle (\r);
\fill[red!12]  (Av) circle (\r);
\fill[cyan!12] (Cv) circle (\r);
\begin{scope}
  \clip (Cv) circle (\r);
  \begin{scope}[even odd rule]
    \clip (0,0) rectangle (5.6,4.3) (Av) circle (\r);
  \end{scope}
  \fill[pattern=north east lines, pattern color=black!55]
    (0,0) rectangle (5.6,4.3);
\end{scope}
\begin{scope}
  \clip (Cv) circle (\r);
  \clip (Av) circle (\r);
  \begin{scope}[even odd rule]
    \clip (0,0) rectangle (5.6,4.3) (Xv) circle (\r);
  \end{scope}
  \fill[pattern=grid, pattern color=black!65]
    (0,0) rectangle (5.6,4.3);
\end{scope}

\draw[semithick, blue!55!black] (Xv) circle (\r);
\draw[semithick, red!55!black]  (Av) circle (\r);
\draw[semithick, cyan!55!black] (Cv) circle (\r);
\node at (0.90,3.00) {$X_t^j$};
\node at (3.15,3.00) {$A_t^{j,\star}$};
\node at (2.02,0.92) {$C_t$};
\draw[line width=0.95pt, orange!70!black, dashed]
  (Cv) circle (\r+0.10);
\node[circle, fill=purple!70!black, inner sep=0.7pt] (pZ)
  at (2.45,2.20) {};
\draw[thin, purple!70!black] (pZ) -- (3.55,2.78);
\node[purple!70!black, anchor=west, inner sep=1pt, font=\scriptsize]
  at (3.55,2.78) {ideal $M_t$};
\fill[pattern=grid, pattern color=black!65]
  (5.55,3.55) rectangle (5.80,3.80);
\draw[thin] (5.55,3.55) rectangle (5.80,3.80);
\node[legend, anchor=west] at (5.94,3.68)
  {$Z_t^\star$};

\fill[pattern=north east lines, pattern color=black!55]
  (5.55,3.22) rectangle (5.80,3.47);
\draw[thin] (5.55,3.22) rectangle (5.80,3.47);
\node[legend, anchor=west] at (5.94,3.34)
  {extra ctx};

\draw[orange!70!black, dashed, line width=0.7pt]
  (5.675,2.975) circle (0.135);
\node[legend, anchor=west] at (5.94,2.97)
  {raw $C_t$};
\node[draw=black!25, rounded corners=2pt, fill=black!3,
      anchor=north west, text width=2.92cm, inner sep=3pt,
      font=\scriptsize, align=left]
  at (3.85,2.30)
  {\textbf{Target condition.}\\[1pt]
   $A_t^{j,\star}\perp S_t^i\mid(X_t^j,Z_t^\star)$};
\node[draw=black!22, rounded corners=2pt, fill=black!2,
      anchor=north west, text width=2.92cm, inner sep=3pt,
      font=\tiny, align=left]
  at (3.85,1.20)
  {\textbf{Notation.} $a\perp b\mid c$: $a$ and $b$ are conditionally independent given $c$.};
\end{scope}

\end{tikzpicture}%
}
\caption{\textbf{Setup and action-sufficient target.} \emph{Left:} one decision step in an agentic workflow; $\Phi_C$ is the bounded decoder class the executor can implement. \emph{Right:} the Venn view of $Z_t^\star$.}
\label{fig:setup}
\end{figure}

\subsection{Setup}
\label{subsec:setup}

At step $t$, the team occupies a shared task state $W_t$. Receiver
$j$ acts from local view $X_t^j=g_t^j(W_t,\xi_t^j)$, while sender $i$
has broader context $C_t$ and sender state $S_t^i=s_t(C_t)$; let
$M_t$ denote the sender-to-receiver message. We compare three
endpoint protocols:
\[
\mathrm{RO}: M_t=\varnothing,\qquad
\mathrm{MA}: M_t=e_t(S_t^i),\ |M_t|\le B_t,\qquad
\mathrm{SF}: M_t=C_t,
\]
where RO is receiver-only, MA is bounded multi-agent communication,
and SF is the raw-context comparator.
The receiver then chooses
$A_t^j$ from $(X_t^j,M_t)$ through a bounded decoder class $\Phi_C$
defined below. Here and below, the subscript $C$ in $\Phi_C,
A_{C,\ell}^\star, R_C^\star$ denotes the receiver's bounded compute
class (not the broader context $C_t$), and $a\perp b\mid c$ denotes
conditional independence.

\begin{definition}[Action-Sufficient Coordination Target]
\label{def:action-sufficient-target}
A variable $Z_t^\star\in\mathcal Z_t$ is an
\emph{action-sufficient coordination target} for receiver $j$ if
$A_t^{j,\star}\perp S_t^i \mid (X_t^j,Z_t^\star)$, where
$A_t^{j,\star}$ is the optimal next action.

\end{definition}

Intuitively, $Z_t^\star$ is the sender-side state sufficient for the
receiver's next action beyond $X_t^j$ (not necessarily unique or
minimal); in our experiments,
$Z_t^\star=(Z_t^{\mathrm{art}},Z_t^{\mathrm{act}})$ is the next
target artifact and action type. The decoder class $\Phi_C$ consists
of admissible maps from $(X_t^j,M_t)$ to internal decision
representations; we use a larger probe class
$\Phi_{\bar C}\supseteq\Phi_C$ to separate message-side recoverability
from receiver-side usability, and assume drop-message closure: if
$\phi\in\Phi_C$ can process $(X_t^j,M_t)$, then $\Phi_C$ also contains
an admissible map that ignores $M_t$.
\subsection{One-Step Utility}
\label{subsec:benefit-tax}

Fix a step and let $\delta\in\{0,1\}$ indicate whether the message is
exposed: the receiver conditions on $(X_t^j,M_t)$ when $\delta=1$ and
on $X_t^j$ when $\delta=0$. Let $R_C^\star(\delta)$ be the
lowest risk attainable by the bounded receiver class $\Phi_C$ on
target $Z_t^\star$, and $R_\rho(\delta)$ be the corresponding
risk of the deployed protocol $\rho$.

The receiver-relative gain $A_{C,\ell}^\star(M_t;Z_t^\star)$ is the
class-optimal risk reduction from seeing the message; for any other
side information $Y_t$, $A_{C,\ell}^\star(Y_t;Z_t^\star)$ is defined
by replacing $M_t$ with $Y_t$. The protocol tax
$\tau_{\rho,\ell}(M_t)$ is the increase in the gap between the
deployed risk and the class-optimal risk after the message is
exposed. It is signed: positive tax means the message moves the
deployed protocol farther from the best decoder in the same bounded
class. In multi-agent LLM systems, positive tax can reflect extra
parsing burden, distraction from irrelevant content, and prompt-interface
mismatch, or other protocol-induced errors. A message, therefore,
helps exactly when its receiver-side gain exceeds the extra protocol
tax it creates.

\begin{resultbox}
\begin{proposition}[Communication-Tax Identity]
\label{thm:communication-tax}
For any deployed protocol $\rho$ and step loss $\ell$,
\begin{equation}
U_{\rho,\ell}(M_t;Z_t^\star)
=
R_\rho(0)-R_\rho(1)
=
A_{C,\ell}^\star(M_t;Z_t^\star)-\tau_{\rho,\ell}(M_t).
\label{eq:tax-identity}
\end{equation}
\end{proposition}
\end{resultbox}

Proposition~\ref{thm:communication-tax} is the accounting identity
used throughout the theory. Later results ask when the gain grows,
when the tax shrinks, and when their difference changes sign. See
Appendix~\ref{app:one-step-proofs} for the proof.
\subsection{Representation-Aware Compression}
\label{subsec:compression}

Compression is useful only when the protocol tax it saves exceeds the information and usability costs it introduces. We separate two
terms: omitted target information $\mathcal I_{\rm omit}$, the
target-relevant information in raw context $C_t$ but not
recovered from $(X_t^j,M_t)$; and decoder mismatch
$\Lambda_C(M_t\!\to\!C_t)$, the difference between how easily the
bounded class $\Phi_C$ can use the compressed message and how easily it can use raw context (positive $\Lambda_C$ favors raw; negative mismatch favors compression). Log-loss definitions are in Appendix~\ref{app:one-step-proofs}.

\begin{resultbox}
\begin{theorem}[Compression Wins by Saving More Tax Than It Loses]
\label{thm:compression-dominance}
Under log-loss, if $M_t=f(C_t)$, then
\begin{equation}
\Delta U_{\rm cmp-raw}
=
\Delta\tau_{\rm raw\to cmp}
-
\mathcal I_{\rm omit}
-
\Lambda_C(M_t\!\to\!C_t).
\label{eq:cmp-vs-raw-gap}
\end{equation}
\end{theorem}
\end{resultbox}

Here $\Delta U_{\rm cmp-raw}$ is the deployed utility of the
compressed message minus that of the raw context, and
$\Delta\tau_{\rm raw\to cmp}$ is the tax saved by not exposing raw
context. Thus, an action-sufficient message wins immediately when it
reduces tax and is no harder for the executor to use than raw
context. Appendix~\ref{app:one-step-proofs} provides the proof, the
expanded entropy form of Eq.~\eqref{eq:cmp-vs-raw-gap}, and an
equivalent receiver-class certificate in terms of the representation
gap $B_C(M_t,C_t)$.

\emph{Bayes sufficiency} is an information-level property, not a
usability guarantee. In an LLM harness, the deployed executor is a
fixed model with a fixed prompt, parser, context window, and tool
interface, and the corresponding bounded receiver class is
\emph{representation-sensitive}: it is not closed under arbitrary recodings of the message surface. We therefore require compressed
messages to preserve target-relevant information while using a
strictly smaller budget than the raw context; the formal condition
excludes degenerate copies and same-size recodings.

\begin{definition}[Proper Bayes-Sufficient Compression]
\label{def:alpha-proper}
For $0<\alpha<1$, $M_t=T(C_t)$ is an
\emph{$\alpha$-proper Bayes-sufficient compression} of $C_t$ for
target $Z_t^\star$ and receiver view $X_t^j$ if
(i) $I(Z_t^\star;C_t\mid X_t^j,M_t)=0$ and
(ii) $H(M_t)\le \alpha H(C_t)$.
\end{definition}

Even a proper Bayes-sufficient compression can fail when it is hard for the bounded executor to use.

\begin{resultbox}
\begin{theorem}[Bayes-Sufficient Compression Need Not Be Receiver-Optimal]
\label{thm:sufficiency-not-enough}
For any $0<\alpha<1$ and $\varepsilon>0$, there exist a task, an
$\alpha$-proper Bayes-sufficient compression
(Definition~\ref{def:alpha-proper}), and a representation-sensitive
receiver class on which the raw context attains zero error, while the
compressed message incurs an error of at least $1/2-\varepsilon$.
\end{theorem}
\end{resultbox}

Concretely, a message may contain the correct artifact and action
information, yet present it in a surface form that the executor fails to
use. Appendix~\ref{app:sufficiency-not-enough} gives the
construction, the finite-alphabet entropy version of the
proper-compression condition, and the corresponding
receiver-compatible sufficiency condition. It also shows that
sufficiency becomes useful when tax savings exceed the receiver-class
representation gap.
\subsection{Where Post-Message Loss Remains}
\label{subsec:loss-decomp}

A useful message need not become a correct action. To localize this failure, we expose one internal bottleneck of the executor. A receiver
is \emph{bottleneck-factorized} if it first maps $(X_t^j,M_t)$ to an
internal target representation $\widehat Z_t$, and then maps
$(X_t^j,\widehat Z_t)$ to the emitted action $A_t^j$. Conditioned on the local view, this gives the chain
$Z_t^\star \to C_t \to M_t \to \widehat Z_t \to A_t^j$, where arrows
denote conditional-independence structure rather than causal timing.

\emph{Externalization} is sender-side loss: useful content in $C_t$
does not reach $M_t$. \emph{Absorption} is receiver-side uptake loss:
useful content in $M_t$ does not enter $\widehat Z_t$.
\emph{Action closure} is receiver-side decision loss: recovered content in $\widehat Z_t$ does not become the correct final action.

\begin{resultbox}
\begin{theorem}[Post-Message Loss Decomposes into Three Stages]
\label{thm:bottleneck-chain}
For any bottleneck-factorized receiver,
\begin{equation}
L_{\rm total}=L_{\rm ext}+L_{\rm abs}+L_{\rm close}.
\label{eq:loss-decomp-identity}
\end{equation}
Each term is nonnegative. The terms correspond respectively to
externalization, absorption, and action closure.
\end{theorem}
\end{resultbox}

The theorem identifies where post-message loss remains, not merely
its magnitude; mutual-information definitions are in
Appendix~\ref{app:bottleneck-proofs}. This motivates the replay experiment: holding the sender message fixed across receivers removes
sender-side variation and isolates receiver-side losses.

\begin{resultbox}
\begin{proposition}[Fixed-Message Replay Isolates Receiver-Side Loss]
\label{prop:replay-isolation}
For two receivers evaluated on the same local input, broader context,
and the same message generated by a fixed sender and held fixed during replay, the sender-side externalization term cancels. The paired
difference, therefore, depends only on receiver-side absorption and
action-closure terms.
\end{proposition}
\end{resultbox}

We instantiate
$Z_t^\star=(Z_t^{\mathrm{art}},Z_t^{\mathrm{act}})$: absorption
appears as missing content, while action closure appears as a failure to convert recovered content into the correct action. On
non-replay protocols, content-missing metrics conflate absorption
with sender-side externalization. Proofs in
Appendix~\ref{app:bottleneck-proofs}.
\subsection{When Stronger Upstream Agents Help}
\label{subsec:alignment}

A stronger upstream agent is not automatically better. It may produce
a more informative message, but it can also introduce additional
protocol tax through longer messages, less compatible surface forms,
or an increased reasoning burden for the executor. For a fixed
executor $R$, compare two upstream agents $s\preceq s'$. Let
$\Delta A_{R,\ell}(s,s')$ denote the extra receiver-relative gain
from using $s'$ instead of $s$, and let
$\Delta\tau_{R,\ell}(s,s')$ denote the extra protocol tax.

\begin{definition}[Burden-Indexed Receiver Family]
\label{def:burden-family}
A receiver family $\{R_b:b\in\mathcal B\}$ is
\emph{burden-indexed} if each receiver's local insufficiency is
summarized by a scalar burden $b$, such as
$H(Z_t^\star\mid X_t^{R_b})$ or an empirical receiver-only proxy.
\end{definition}

\begin{resultbox}
\begin{theorem}[Sender-Upgrade Burden Threshold under Single Crossing]
\label{prop:alignment-threshold}
For any step loss $\ell$ and fixed executor $R$,
\[
U_{s',R,\ell}-U_{s,R,\ell}
=
\Delta A_{R,\ell}(s,s')-\Delta\tau_{R,\ell}(s,s').
\]
For a burden-indexed family, let $\Psi_{s,s'}(b)$ be this net value
at burden $b$. If $\Psi_{s,s'}$ is continuous and single-crossing,
with negative value at low burden and positive value at high burden,
then there is a unique threshold $b^\star$ at which the stronger
upstream agent changes from harmful to helpful.
\end{theorem}
\end{resultbox}

The theorem makes sender choice depend on the receiver's local burden (Figure~\ref{fig:sender-threshold}).
Section~\ref{sec:experiments} estimates this threshold; the proof is
in Appendix~\ref{app:alignment-proofs}.

\section{Dynamic Coordination and Protocol Control}
\label{sec:implications}

We turn the step-level theory of Section~\ref{sec:theory} into an
inference-time control rule that scores discrete protocol actions by
the same gain, tax, mismatch, sender-upgrade, and residual-risk
quantities.

\subsection{Residual Risk in Shared-State Workflows}
\label{subsec:propagation}

A one-step communication error does not always have the same cost. In loosely coupled tasks, later observations may resynchronize the agents. In shared-state tasks, however, an incorrect intermediate
artifact, a stale plan state, or a workspace inconsistency may persist
and affect subsequent steps. We represent this with a scalar
$D_t\ge 0$: the unresolved task-relevant discrepancy in the shared state.

In our clipped residual model, let $k_t$ be the fresh discrepancy
at step $t$, $q_t$ the resolved discrepancy, and $\lambda_t$ the
carry coefficient measuring how strongly residual persists through
the shared state. For a counterfactual trajectory with residual
$\widetilde D_t$, the dynamics and residual gap are:
\begin{equation}
D_{t+1}=k_t+\lambda_t[D_t-q_t]_+,
\qquad
\Delta_t:=\widetilde D_t-D_t.
\label{eq:accumulation}
\end{equation}
The scalar notation is for readability; vector residuals follow componentwise. A residual gap multiplies forward while both
trajectories remain in the carry regime; later observations or repairs
can shrink the gap, and the gap vanishes only when the perturbed residual is fully resolved. The following proposition compares trajectories with the same discrepancy, repair, and carry sequence.

\begin{resultbox}
\begin{proposition}[Conditional Propagation]
\label{prop:conditional-propagation}
Suppose two trajectories share the same $(k_t,q_t,\lambda_t)$
sequence and differ only through an initial residual gap $\Delta_t$.
If $D_u>q_u$ and $\widetilde D_u>q_u$ for all
$u=t,\ldots,t+h-1$, then
\[
\Delta_{t+h}
=
\Bigl(\prod_{u=t}^{t+h-1}\lambda_u\Bigr)\Delta_t.
\]
\end{proposition}
\end{resultbox}

Proposition~\ref{prop:conditional-propagation} is the dynamic
complement to the one-step theory. Weak coupling makes local mistakes
wash out; persistent coupling turns them into future risk. See
Appendix~\ref{app:state-prop-proofs} for the detailed proof.
\subsection{Analysis-Guided Protocol Control}
\label{subsec:selector}

The resulting protocol selector chooses a communication action at
each step. The action set is discrete, with budgets and model choices
treated as action parameters:
\[
\mathcal A_t
\subseteq
\{\texttt{skip},\texttt{compress}(B),\texttt{raw},\texttt{escalate}(C'),\texttt{refresh}\}.
\]
The actions correspond to the theory as follows. \texttt{skip} uses
no message and implements the RO endpoint. \texttt{compress}(B) sends
a bounded message under budget $B$ and is evaluated by the
compression criterion. \texttt{raw} exposes full context and remains
necessary because Bayes-sufficient compression can still fail for a
bounded executor. \texttt{escalate}(C') directly uses a stronger receiver
on the compressed message, scored by the corresponding gain--tax comparison.
\texttt{refresh} triggers shared-state resynchronization and is
scored by residual propagation.

For an action $\alpha$ at step $t$, let $\Phi_\alpha$ be the deployed
receiver class, $\rho_\alpha$ the induced protocol, and $M_t(\alpha)$
the message. Let $c(\alpha)$ be the immediate cost, and $r_t(\alpha)$ the estimated finite-horizon
residual risk under the clipped propagation model of
Proposition~\ref{prop:conditional-propagation}. The selector scores
each candidate by receiver-relative utility minus immediate and
future costs, with $\eta_c,\eta_r\ge 0$ tuned on a
held-out split:

\begin{resultbox}
\begin{equation}
J_t(\alpha)
=
A_{\Phi_\alpha,\ell}^\star(M_t(\alpha);Z_t^\star)
-
\tau_{\rho_\alpha,\ell}(M_t(\alpha))
-
\eta_c c(\alpha)
-
\eta_r r_t(\alpha),
\qquad
\alpha_t^\star\in\argmax_{\alpha\in\mathcal A_t}J_t(\alpha).
\label{eq:control-value}
\end{equation}
\end{resultbox}

The gain term measures target recovery, tax measures protocol burden, and the remaining terms charge immediate cost and future residual risk. Eq.~\eqref{eq:control-value} guides the deployed score $\widehat J_t$ through the following procedure.

\textbf{Offline calibration.} We split records by \texttt{instance\_id}, keeping the receiver interface and action set fixed. Paired RO/SF/MA calibration records supply an action-level offset $\widehat\tau$ that captures protocol burden together with gain effects. Held-out RO records also supply a local-sufficiency rate for an early \texttt{skip} branch. The offsets, skip threshold, and cost weights are fixed before evaluation; the selector reads instance-level signals at test time.

\textbf{Scoring and execution.} The static ContextBench study ranks stored \texttt{skip}, \texttt{raw}, and \texttt{compress} outcomes. The online ToolSandbox gate selects \texttt{skip} before sender calls. Otherwise, the five-action path (\texttt{refresh/skip/raw/compress/refresh})
constructs and caches real \texttt{compress} and \texttt{refresh} messages before nine receiver probes, capped at four tokens. \texttt{raw} reads full upstream context; \texttt{escalate} reuses the compressed message with a stronger receiver. The gain surrogate $\hat A$ is the clipped leading-token log-probability difference. The selected action executes once using its cached candidate. Appendix~\ref{app:tax-controller} specifies these paths and their call counts.

The static one-step ablation uses $\eta_c=10^{-4}$ on raw token counts and $\eta_r=0$. Relative costs in tables are normalized for display. For shared-state workflows, logged discrepancy features inform the residual score; \texttt{refresh} regenerates the shared-state summary. The perturbation and repair studies evaluate this mechanism separately. Appendix~\ref{app:selector-regret} gives the value-regret bound under bounded score-estimation error.
\section{Experiments}
\label{sec:experiments}

We tested representation sensitivity (Theorem~\ref{thm:sufficiency-not-enough}), receiver-side loss (Theorem~\ref{thm:bottleneck-chain}), sender upgrades (Theorem~\ref{prop:alignment-threshold}), and the selector score (Eq.~\eqref{eq:control-value}). Appendix~\ref{app:exp-details} provides the protocols, paired uncertainty analyses, and supplementary results.

\begin{table}[!t]
\centering
\caption{\textbf{Protocol performance by receiver.} Regime labels describe the Qwen condition.
$\mathrm{hit\_joint}$ for RO/SF/MA. In the Qwen rows, MA uses
Qwen3-Next-80B-A3B as sender; in the OpenAI extension, the receiver is shown
at left, and MA uses \texttt{gpt-5.4} as sender. $^{\dagger}$ denotes the
matched-sender Qwen3.5-9B condition. Released supports for the displayed Qwen
rows are $480$ steps for ContextBench, $72$ for $\tau$-bench and ToolSandbox,
and $100$ for BFCL. OpenAI cell counts ($60$--$300$) are in Appendix~\ref{app:cross-family}.}
\label{tab:main-results}
\footnotesize
\renewcommand{\arraystretch}{1.05}
\setlength{\tabcolsep}{1.6pt}
\begin{tabular}{@{}l@{\hspace{6pt}}ccc@{\hspace{7pt}}ccc@{\hspace{7pt}}ccc@{\hspace{7pt}}ccc@{\hspace{7pt}}ccc@{}}
\toprule
 & \multicolumn{3}{c}{\textbf{ContextBench}}
 & \multicolumn{3}{c}{\textbf{ToolSandbox}}
 & \multicolumn{3}{c}{\textbf{$\tau$-bench}}
 & \multicolumn{3}{c}{\textbf{BFCL}}
 & \multicolumn{3}{c}{\textbf{Derived summaries}} \\
 & \multicolumn{3}{c}{\emph{\footnotesize helpful}}
 & \multicolumn{3}{c}{\emph{\footnotesize noise}}
 & \multicolumn{3}{c}{\emph{\footnotesize boundary}}
 & \multicolumn{3}{c}{\emph{\footnotesize noise}}
 & \multicolumn{3}{c}{\footnotesize $\overline{\Delta}_{\rm comm}\;\;\overline{\Delta}_{\rm cmp}\;\;\Delta_{\rm comm}^{\rm Ctx}$} \\
\cmidrule(lr){2-4}\cmidrule(lr){5-7}\cmidrule(lr){8-10}\cmidrule(lr){11-13}\cmidrule(lr){14-16}
Receiver / Setup & RO & SF & MA & RO & SF & MA & RO & SF & MA & RO & SF & MA & \multicolumn{3}{c}{} \\
\midrule
\rowcolor{regimeHelpful}
\multicolumn{16}{@{}l}{\textbf{Qwen bounded protocol} (Qwen3.5-9B receiver, Qwen3-Next-80B-A3B sender)} \\
Qwen3.5-9B
& .633 & \cellR{.294} & \cellbest{.775}
& \cellg{.889} & \cellsec{.903} & \cellR{.653}
& .139 & \cellbest{.167} & .139
& \cellsec{.980} & \cellbest{1.00} & \cellR{.750}
& $-$.08 & $-$.01 & +.14 \\
Qwen matched$^{\dagger}$
& \cellder{.633} & \cellder{.294} & .565
& \cellder{.889} & \cellder{.903} & \cellg{.917}
& \cellder{.139} & \cellder{.167} & \cellg{.167}
& \cellder{.980} & \cellder{1.00} & \cellg{.910}
& $-$.02 & +.05 & $-$.07 \\
\midrule
\rowcolor{regimeHelpful}
\multicolumn{16}{@{}l}{\textbf{OpenAI model-family extension} (MA sender = \texttt{gpt-5.4}; receiver shown at left)} \\
\texttt{gpt-4o-mini}
& .650 & \cellg{.743} & \cellbest{.787}
& .803 & \cellg{.957} & \cellbest{1.00}
& .253 & \cellR{.020} & \cellbest{.373}
& \cellsec{.878} & \cellR{.722} & \cellr{.833}
& +.10 & +.14 & +.14 \\
\texttt{gpt-5.4-nano}
& .493 & \cellg{.637} & \cellbest{.783}
& .757 & \cellg{.927} & \cellbest{1.00}
& .313 & \cellr{.233} & \cellr{.300}
& \cellbest{.883} & \cellr{.811} & \cellr{.839}
& +.12 & +.08 & +.29 \\
\texttt{gpt-5.4-mini}
& .613 & \cellg{.770} & \cellbest{.787}
& .697 & \cellg{.890} & \cellbest{1.00}
& .293 & \cellr{.233} & \cellbest{.327}
& .839 & \cellr{.833} & \cellbest{.844}
& +.13 & +.06 & +.17 \\
\texttt{gpt-5.4}
& .600 & \cellg{.777} & \cellbest{.853}
& .653 & \cellg{1.00} & \cellbest{.993}
& .240 & \cellr{.180} & \cellR{.147}
& .861 & \cellr{.833} & \cellbest{1.00}
& +.16 & +.05 & +.25 \\
\texttt{o3}
& .223 & \cellR{.133} & \cellbest{.493}
& .740 & \cellsec{.980} & \cellbest{1.00}
& \cellsec{.200} & .173 & \cellr{.160}
& \cellbest{1.00} & .983 & \cellbest{1.00}
& +.12 & +.10 & +.27 \\
\texttt{o4-mini}
& .500 & \cellsec{.617} & \cellbest{.743}
& .677 & \cellsec{.733} & \cellbest{1.00}
& \cellsec{.273} & \cellbest{.280} & \cellR{.160}
& .844 & \cellsec{.889} & \cellbest{1.00}
& +.15 & +.10 & +.24 \\
\bottomrule
\end{tabular}%

\end{table}
\subsection{Experimental Setup}
\label{sec:exp-setup}
We evaluated six benchmark families (BFCL \citep{patil2023gorilla},
ToolSandbox \citep{lu2024toolsandbox}, ContextBench \citep{zhu2026contextbench},
RepoBench \citep{liu2024repobench}, SWE-bench Lite \citep{jimenez2024swebench},
and $\tau$-bench \citep{yao2024tau}) covering high-local-sufficiency,
context-heavy, capability-transfer, and boundary regimes; LongCLI-Bench
provides a shared-state stress test (Appendix~\ref{app:shared-state}).
The default sender is Qwen3-Next-80B-A3B, the receiver Qwen3.5-9B,
the message budget $B{=}128$ tokens (SWE-Lite uses Qwen3-Coder-Next-FP8 as sender), and matched-sender
($\mathrm{MA}_{\rm match}$, 9B$\to$9B), stronger-sender ($\mathrm{MA}_{\rm main}$),
and fixed-message replay contrasts are paired at the (instance, step) level.
For scoring, we map each benchmark's native supervision into a set of
acceptable artifact/action targets $G_t=\{(Z_{t,k}^{\rm art},Z_{t,k}^{\rm act})\}_{k=1}^{K_t}$; these labels are never exposed to the sender, receiver, or selector.
Appendix~\ref{app:benchmark-summary} gives mapping rules and traces. One receiver instruction template is shared across the no-message and with-message paths; the user-turn context varies by protocol.
Primary accuracy is \texttt{hit\_joint} ($h_j$), computed directly from the
paired target. For the content/closure diagnostic, we report an artifact
accuracy $h_a$, conditional action-conversion $h_{\mathrm{action}}$, content
loss $L_{\rm content}$, and closure loss $L_{\rm closure}$; when all quantities
are measured on the same step support, these correspond to the empirical identity
$1-h_j=\mathbb{E}[1-h_a]+\mathbb{E}[h_a(1-h_{\mathrm{action},t})]$.
Replay tables with non-uniform coverage report $h_j$ directly and treat
the loss columns as diagnostic component means rather than reconstructing
$h_j$ from marginal products. The three theory-aligned contrasts are
$\Delta_{\rm comm}{:=}\mathrm{Acc}(\mathrm{MA}_{\rm main}){-}\mathrm{Acc}(\mathrm{RO})$,
$\Delta_{\rm cmp}{:=}\mathrm{Acc}(\mathrm{MA}_{\rm main}){-}\mathrm{Acc}(\mathrm{SF})$, and
$\Delta_{\rm strong}{:=}\mathrm{Acc}(\mathrm{MA}_{\rm main}){-}\mathrm{Acc}(\mathrm{MA}_{\rm match})$,
with gain surrogate $\hat A{:=}\mathbb{E}[\texttt{surprisal\_no\_msg}-\texttt{surprisal\_bits}]$, whose two terms are hit-based proxy scores rather than measured log-probabilities (Appendix~\ref{app:claim1-extra}), and token cost taken as mean tokens per evaluated step. Evaluation-set definitions and protocol-variant controls in
Appendices~\ref{app:benchmark-summary} and~\ref{app:cross-family}.

\subsection{Benchmark-Wide Numerical Summary}
\label{sec:exp-summary}
Table~\ref{tab:main-results} compares RO, SF, and MA; Table~\ref{tab:replay-summary} fixes the sender message. Figures~\ref{fig:regimes}--\ref{fig:tax-knobs-main} connect these contrasts to protocol choices. Qwen comparisons are paired within each evaluation set. OpenAI cells use the counts in Appendix~\ref{app:cross-family}, including unequal protocol coverage. The derived columns summarize displayed rates: $\overline{\Delta}_{\rm comm}$ and $\overline{\Delta}_{\rm cmp}$ average the four benchmarks, and $\Delta_{\rm comm}^{\rm Ctx}$ gives the ContextBench contrast.

Instance-clustered bootstrap analyses support the helpful and harmful regimes on separately stated paired supports (Appendix~\ref{app:benchmark-summary}). ContextBench MA exceeds RO by $0.1711$ (95\% CI $[0.1417,0.2012]$; $2{,}390$ steps, $799$ instances), and RepoBench by $0.1611$ ($[0.1340,0.1891]$; $2{,}396$ steps, $800$ instances). The $\tau$-bench interval includes zero. These paired estimates complement the main table and retain their own sample counts.

\begin{table}[!t]
\centering
\caption{\textbf{Fixed-message replay on ContextBench.}
Receivers process the same DeepSeek-V3.1 messages, isolating receiver-side effects. Component coverage is non-uniform across receivers; $h_j$ is computed directly; the content and closure columns are diagnostic component means.}
\label{tab:replay-summary}
\footnotesize
\setlength{\tabcolsep}{2.4pt}
\renewcommand{\arraystretch}{1.00}
\begin{tabular}{@{}l c c c c c c c c@{}}
\toprule
\multicolumn{1}{c}{\textbf{Receiver}} &
\multicolumn{3}{c}{\textbf{Replay accuracy}} &
\multicolumn{3}{c}{\textbf{Observable post-message error}} &
\multicolumn{2}{c}{\textbf{Summary}} \\
\cmidrule(lr){2-4}\cmidrule(lr){5-7}\cmidrule(l){8-9}
 &
$h_j\uparrow$ & $h_a\uparrow$ & $h_{\rm action}\uparrow$ &
$L_{\rm content}\downarrow$ & $L_{\rm closure}\downarrow$ & frac.\,closure &
$\eta\uparrow$ & Regret$\downarrow$ \\
\midrule
\rowcolor{regimeHelpful}
DeepSeek-R1         & \textbf{0.731} & 0.732 & 0.989 & \vwarn{0.268} & 0.008 & \phantom{0}3\% & 1.00 & 0.000 \\
\rowcolor{rowAlt}
Qwen3.5-9B         & 0.728 & 0.728 & 0.989 & \vwarn{0.272} & 0.008 & \phantom{0}3\% & 1.00 & 0.003 \\
\rowcolor{regimeHelpful}
Llama-70B        & 0.714 & 0.714 & 0.997 & \vwarn{0.286} & 0.002 & \phantom{0}1\% & 1.00 & 0.017 \\
\rowcolor{rowAlt}
DeepSeek-V3.1       & 0.707 & 0.730 & 0.977 & \vwarn{0.270} & 0.017 & \phantom{0}6\% & 0.97 & 0.024 \\
\rowcolor{regimeHelpful}
Qwen2.5-7B         & 0.702 & 0.702 & 0.975 & \vwarn{0.298} & 0.018 & \phantom{0}6\% & 1.00 & 0.029 \\
\rowcolor{rowAlt}
Qwen3.5-397B-A17B  & 0.685 & 0.717 & 0.926 & \vwarn{0.283} & 0.053 & 16\% & 0.95 & 0.046 \\
\midrule
\rowcolor{regimeNoise}
GLM-5          & 0.382 & 0.741 & 0.545 & 0.259 & \vneg{0.337} & \vneg{57\%} & 0.52 & \vneg{0.349} \\
\rowcolor{regimeNoise}
MiniMax-M2.5       & 0.036 & 0.613 & 0.053 & 0.387 & \vneg{0.581} & \vneg{60\%} & 0.06 & \vneg{0.695} \\
\bottomrule
\end{tabular}%
\end{table}

\subsection{When Compression Beats Raw Context}
\label{sec:exp-claim1}

\begin{figure}[t]
\centering
\safeincludegraphics[width=0.95\columnwidth]{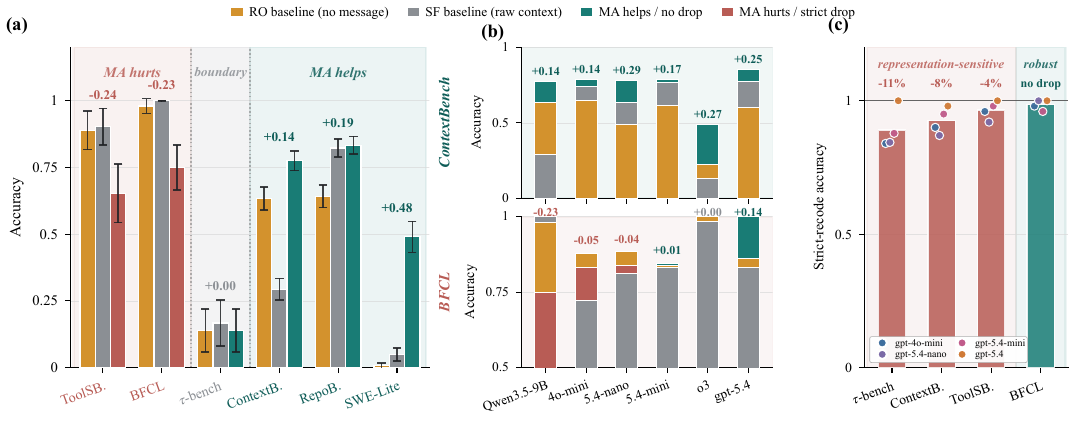}
\caption{\textbf{Helpful, noisy, and boundary regimes for compression.}
(a) Six benchmarks under the Qwen3.5-9B protocol, grouped by communication effect.
(b) ContextBench (top) and BFCL (bottom) protocol comparison for Qwen and OpenAI receivers.
(c) Controlled representation-sensitivity probe motivated by Theorem~\ref{thm:sufficiency-not-enough}: direct vs.\ strict-recoded forms across four benchmarks.}
\label{fig:regimes}
\end{figure}

Communication value depends on the receiver's local view and interface. In Figure~\ref{fig:regimes}(a), the default Qwen protocol hurts BFCL and ToolSandbox, helps ContextBench ($0.633{\to}0.775$) and RepoBench ($0.642{\to}0.833$), and has little effect on $\tau$-bench. ContextBench also improves across the OpenAI receivers in panel (b), including \texttt{gpt-5.4} ($0.60{\to}0.85$). ToolSandbox behaves differently across families: \texttt{gpt-4o-mini} improves from $0.803$ to $1.000$, whereas the default Qwen receiver drops from $0.889$ to $0.653$ (Table~\ref{tab:main-results}). Thus the regime belongs to the receiver and protocol condition as well as the task.

ContextBench gains persist when message truncation is controlled. On the $480$-step audit, MA exceeds RO by $10.64$ percentage points among $404$ capped messages and by $32.89$ points among $76$ uncapped messages. The respective paired 95\% intervals are $[5.7,15.8]$ and $[22.4,43.4]$ points. All capped messages retain both decision fields. Appendix~\ref{app:claim1-extra} reports the split, a paired budget increase, and gain-surrogate diagnostics.

Surface form matters even with the target held fixed. The strict-recoding probe in Figure~\ref{fig:regimes}(c) lowers accuracy by $13.5$ points on $\tau$-bench, $6.0$ on ContextBench, and $3.5$ on ToolSandbox; BFCL is unchanged. This controlled comparison links Theorem~\ref{thm:sufficiency-not-enough} to the deployed interface and motivates retaining \texttt{raw} in Section~\ref{sec:exp-implications}.
\subsection{Closure-Dominated Receiver-Side Loss}
\label{sec:exp-claim2}

\begin{figure}[t]
\centering
\safeincludegraphics[width=0.95\columnwidth]{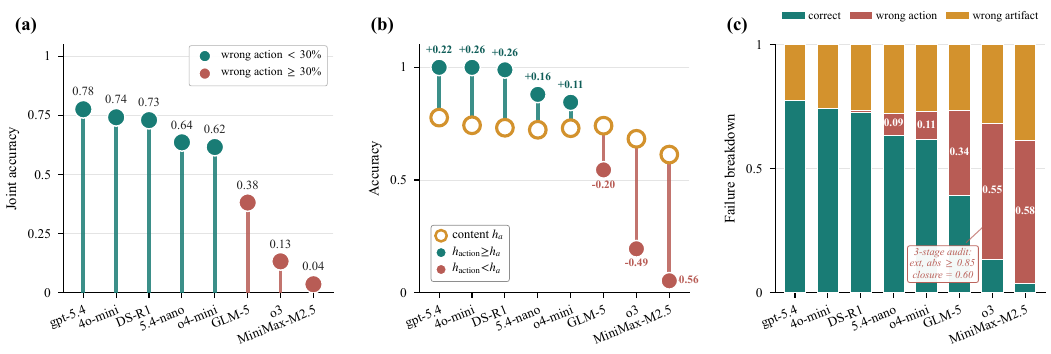}
\caption{\textbf{Receiver-side content and closure bottlenecks on ContextBench.}
(a) Joint accuracy by receiver, colored by wrong-action share.
(b) Content $h_a$ (open ring) and action $h_{\mathrm{action}}$ (filled, direction-colored) per receiver, with $h_{\mathrm{action}}{-}h_a$ annotated.
(c) Observable outcome decomposition into correct/wrong action/wrong artifact; inline three-stage diagnostic motivated by Theorem~\ref{thm:bottleneck-chain}.}
\label{fig:absorption}
\end{figure}

Correct artifact recovery can coexist with frequent action errors. Figure~\ref{fig:absorption} tests the output events motivated by Theorem~\ref{thm:bottleneck-chain}. DeepSeek-R1, \texttt{gpt-5.4}, and \texttt{gpt-4o-mini} have $h_{\mathrm{action}}{\ge}0.98$ and closure fractions at or below $3\%$, so joint accuracy tracks artifact accuracy. GLM-5, \texttt{o3}, and MiniMax-M2.5 retain $h_a{\ge}0.61$ but have action-conversion rates of $0.545$, $0.195$, and $0.053$. The separate \texttt{gpt-5.4}-sender audit in panel (c) places \texttt{o3}'s externalization and absorption success at $\ge0.85$, with conditional closure at $0.60$.

Parsing audits locate action errors after valid output and correct artifact recovery. In a \texttt{gpt-5.4}-sender replay, wrong-action rates among such outputs are $44.5\%$ for GLM-5, $94.0\%$ for MiniMax-M2.5, and $40.8\%$ for \texttt{o3}. Outputs explicitly choose \texttt{read} when the evaluated action is \texttt{edit}. On SWE-Lite, RO artifact accuracy spans $0.885$--$0.917$ across three parse-valid seed slices; the reported zero-joint slice reflects action errors (Appendix~\ref{app:benchmark-summary}). Appendices~\ref{app:claim2-extra} and~\ref{app:case-studies} provide parsing rates, evaluation supports, and traces.
\subsection{The Stronger-Sender Threshold}
\label{sec:exp-claim3}

\begin{figure}[t]
\centering
\safeincludegraphics[width=0.95\columnwidth]{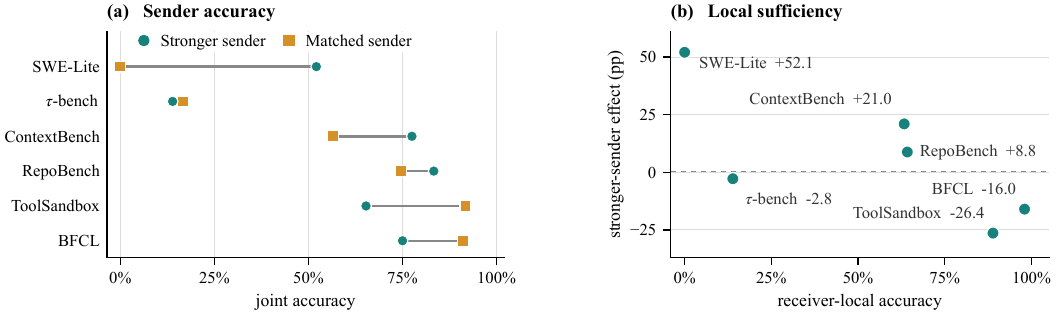}
\caption{\textbf{Stronger-sender effects are regime-dependent.}
Throughout, $\Delta_{\rm strong} := \mathrm{Acc}(\mathrm{MA}_{\rm main}) - \mathrm{Acc}(\mathrm{MA}_{\rm match})$; positive means the stronger sender helps.
(a) Stronger- and matched-sender accuracies for the six Qwen benchmark conditions.
(b) Reported net effects against receiver-only accuracy. SWE-Lite uses seeds 2 and 1 for stronger and matched senders. RepoBench's $0.088$ effect uses unrounded endpoints.}

\label{fig:alignment}
\end{figure}

Sender upgrades help on SWE-Lite, ContextBench, and RepoBench but hurt the other three Qwen conditions. Table~\ref{tab:main-results} and Figure~\ref{fig:alignment} show gains of $+0.52$, $+0.21$, and $+0.09$, compared with $-0.03$ on $\tau$-bench, $-0.26$ on ToolSandbox, and $-0.16$ on BFCL. The negative $\tau$-bench effect at low RO accuracy shows that local accuracy alone does not determine the regime; message fit and protocol burden also matter. Appendix~\ref{app:claim3-extra} extends the sender swaps to OpenAI receivers.

Expanding the paired study from $22$ to $41$ anchors preserves the crossover in RO accuracy: the estimate is $0.6289$ with bootstrap 95\% CI $[0.552,0.706]$, compared with the original $0.638$ and $[0.491,0.780]$. Of $38$ nonzero anchors, $34$ ($89.5\%$) follow a single positive-to-negative crossing as local sufficiency increases. A hierarchical analysis of $12{,}540$ paired steps favors the single-crossing structure over linear and two-crossing alternatives in all $10$ anchor-separated folds. Appendix~\ref{app:claim3-extra} reports the fits and leave-family-out analysis; Appendix~\ref{app:cross-family} gives the model-family controls.
\subsection{Inference-Time Protocol Selection in Practice}
\label{sec:exp-implications}

\begin{figure}[t]
\centering
\safeincludegraphics[width=0.95\columnwidth]{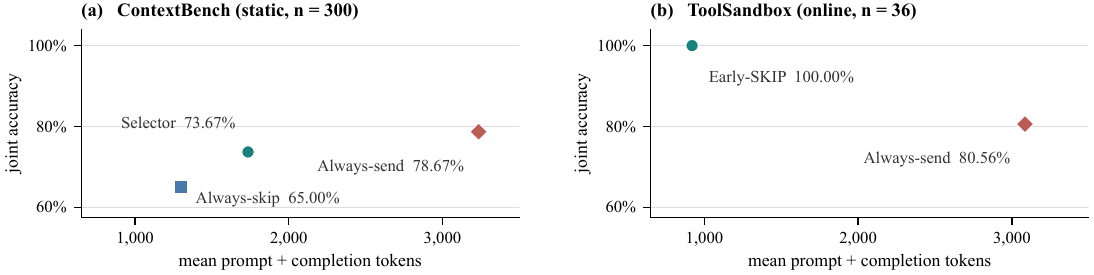}
\caption{\textbf{Accuracy and recorded token cost under protocol selection.}
(a) ContextBench static audit on 300 records: always-skip, selector, and always-send.
(b) ToolSandbox online audit on 36 steps: early-skip and always-send.
Both panels use absolute joint accuracy and mean prompt-plus-completion tokens; supports are separate.}

\label{fig:tax-knobs-main}
\end{figure}

The selector trades communication cost against accuracy according to the receiver's needs. On the 300-record ContextBench audit in Figure~\ref{fig:tax-knobs-main}(a), it reaches $0.7367$ accuracy at $1{,}735$ tokens per step, between always-skip ($0.6500$, $1{,}300$ tokens) and always-send ($0.7867$, $3{,}235.4$ tokens). It recovers $8.67$ accuracy points over always-skip while using $46.4\%$ fewer tokens than always-send. This static comparison ranks stored outcomes for the same three actions.

On the online $36$-step ToolSandbox slice, the early-skip path reached $1.000$ accuracy with $919.4$ prompt-plus-completion tokens and $6.60$ seconds per step, versus $0.8056$, $3{,}085.4$ tokens, and $9.56$ seconds for always-send. On ContextBench, instance-disjoint calibration splits gave accuracy $0.7368$--$0.7458$ across calibration fractions of $5$--$40\%$. The learned-router comparison reached utility $0.5692$, versus $0.5191$ for gradient boosting under the same cost weight. The static $\tau$-bench result retains a useful failure case: rewriting reached $0.139$ accuracy versus the selector's $0.083$ at similar input-token cost. Appendices~\ref{app:tax-controller} and~\ref{app:shared-state} give the baselines, ablations, call accounting, and residual studies.

\section{Conclusion}
\label{sec:conclusion}

Receiver-relative communication trades information loss and decoder compatibility against protocol tax and local burden. This step-level view explains why one protocol can help or hurt: compression pays off only when tax savings exceed omitted information and decoder mismatch, Bayes sufficiency does not guarantee receiver usability, post-message loss splits into externalization, absorption, and action closure, and a stronger sender helps only above a receiver-burden threshold. On six benchmarks, one fixed Qwen protocol lifts ContextBench joint accuracy from $0.633$ to $0.775$ while reducing ToolSandbox accuracy from $0.889$ to $0.653$, and fixed-message replay exposes receivers that recover the correct artifact yet choose the wrong action. The resulting selector adapts communication to receiver needs, and Appendix~\ref{app:limitations} discusses trajectory-level success and calibration across interfaces.

\bibliographystyle{plainnat}
{\small
\bibliography{reference}
}

\section*{Reproducibility statement}
The theoretical assumptions and full proofs are in Appendix~\ref{app:theory}. Appendix~\ref{app:exp-details} specifies benchmark construction, evaluation protocols, and supporting results; Appendix~\ref{app:tax-controller} gives selector settings, candidate construction, calibration splits, and call accounting. Appendix~\ref{app:case-studies} provides fixed-message traces.

\section*{AI use statement}
During manuscript revision, we used OpenAI Codex to assist with language editing, LaTeX formatting, methodological consistency checks, inspection of existing code and result records, analysis and presentation of existing results, and clarification of implementation parameters. Implementation settings were incorporated following author confirmation. This revision used existing experimental records and did not include new experimental runs. The authors are responsible for the final content of the paper.

\newpage
\appendix
\raggedbottom
\DoAppendixContents

The appendix is organized in the order body sections reference it: Appendix~\ref{app:related} expands the related-work discussion previewed in Section~\ref{sec:intro}; Appendix~\ref{app:theory} provides full proofs for Sections~\ref{sec:theory} and~\ref{sec:implications}; Appendix~\ref{app:exp-details} provides experimental details and supporting results for Section~\ref{sec:experiments}; Appendix~\ref{app:case-studies} collects trajectory-level case studies; Appendix~\ref{app:limitations} expands scope and limitations from Section~\ref{sec:conclusion}.

\section{Extended Related Work}
\label{app:related}

\paragraph{Harness engineering and multi-agent workflow design.}
A large body of recent work treats LLM-based multi-agent systems as an engineering problem of workflow decomposition, role allocation, and protocol design. Early systems introduced role-playing and conversational agent societies \citep{li2023camel}, programmable multi-agent conversations \citep{wu2023autogen}, dynamic agent-group orchestration \citep{chen2023agentverse}, standardized operating procedures for collaborative agents \citep{hong2023metagpt}, and chat-based software-development teams \citep{qian2024chatdev}. More recent software and agentic workflow settings instantiate explicit multi-agent or sub-agent structures to scale context handling, task decomposition, and execution, including self-organized coding agents for large codebases \citep{ishibashi2024soa}, modularized workflow optimization \citep{niu2025flow}, lightweight multi-supervisor orchestration \citep{sami2025nexus}, structured communication with hierarchical refinement \citep{wang2025talkhier}, agile-role software engineering agents \citep{tawosi2025almas}, protocol-driven multi-agent engineering \citep{mao2025semap}, and agentic reasoning systems that orchestrate models with tools and refine solutions over multiple rounds \citep{zhou2025alphaapollo}. These works primarily propose better harnesses or workflow templates. Harness-aware evaluation makes the complementary point on the measurement side: reported reasoning performance depends on the harness around the model as well as on the model itself \citep{zhou2026harness}. Our paper instead fixes the general sender--receiver setting and asks a different question: what information should move between components when the receiver is bounded?

\paragraph{Training and equilibrium approaches to multi-agent coordination.}
A complementary line of work improves coordination by changing the agents or their interaction structure. Equilibrium-based methods cast multi-agent LLM reasoning as a Bayesian Nash equilibrium \citep{yi2025debate} or select stable coordination conventions through an entropy-regularized quantal response equilibrium \citep{xie2026dice}. Trust-region and sequential fine-tuning methods train LLM teams with per-update improvement guarantees and plug-and-play replacement of individual agents \citep{xie2026teamtr,xie2026sat}, and in multi-agent reinforcement learning, acyclic reachability-based coordination reduces redundant communication between agents \citep{xie2025acorn}. These methods train or restructure the team, whereas our analysis holds the deployed sender and receiver fixed and asks when a single message improves the receiver's next decision.

\paragraph{Tool use, feedback, and agent-computer interfaces.}
Single-agent language-agent work also motivates our focus on next-step artifact/action decisions \citep{xiao2025visual, xiao2026not, xiao2026staying}. ReAct interleaves reasoning and environment actions \citep{yao2023react}, Toolformer studies learned API-use decisions \citep{schick2023toolformer}, Reflexion uses verbal feedback and memory to improve later actions \citep{shinn2023reflexion}, and SWE-agent shows that the interaction interface itself can materially change software-agent performance \citep{yang2024sweagent}. These systems emphasize that action spaces, tool schemas, feedback channels, and interface design shape what a model can turn into an action. Our receiver-relative view treats these concerns as part of the executor's bounded decoder and protocol tax rather than as fixed properties of a benchmark. Work on single-model reasoning visualizes reasoning trajectories by their distances to candidate answers \citep{zhou2026landscape} and reweights policy optimization by rollout confidence and question difficulty \citep{zhou2026codapo}; these methods study how one model reaches an answer, whereas our step-level diagnostics locate where a message fails between two models.

\paragraph{Communication-centric surveys, protocol taxonomies, and failure analyses.}
A second line of work organizes the MA-LLM literature around communication itself. Beyond Self-Talk surveys LLM-based multi-agent systems from a communication-centric perspective \citep{yan2025beyond}, while A Survey of AI Agent Protocols analyzes the protocol layer and compares existing designs along dimensions such as scalability and latency \citep{yang2025agentprotocols}. Most closely aligned with our motivation, MAST provides a fine-grained taxonomy of specification, misalignment, and verification failures across multiple MAS frameworks \citep{cemri2025mast}. These papers clarify the design space and its failure modes but remain primarily descriptive or taxonomic. We recast these concerns as receiver-relative quantities for analysis, diagnosis, and protocol selection.

\paragraph{Coordination benchmarks and protocol evaluation.}
Recent evaluation work has also made multi-agent coordination and protocol choice first-class benchmark targets. MultiAgentBench evaluates collaboration and competition across diverse interactive scenarios \citep{zhu2025multiagentbench}; LLM-Coordination studies pure coordination settings involving environment comprehension, theory-of-mind reasoning, and joint planning \citep{agashe2025llmcoordination}; and ProtocolBench argues that protocol choice should be evaluated not only by task success but also by latency, message overhead, and robustness \citep{du2025protocolbench}. Recent routers and resource-rational protocols, including MasRouter and auction-based communication, make related inference-time choices over collaboration modes, model assignments, or speaking rights \citep{yue2025masrouter,fan2025costeffective}. Agent evaluations in other domains likewise separate surface success from task-relevant correctness, scoring reproducible speedups rather than functional patches in software performance optimization \citep{cui2026perfopt} and semantic faithfulness rather than compilation in natural-language-to-Lean formalization \citep{zhang2026beyond}. Our use of benchmarks is narrower and identification-oriented: RO / SF / MA contrasts, same-message replay, and sender swap serve here as tools for isolating communication gain, receiver-side loss, and alignment effects rather than as a broad leaderboard target.

\paragraph{Communication under asymmetric information and protocol-sensitive collaboration.}
Communication and Verification in LLM Agents explicitly studies collaboration under information asymmetry and unequal verification capability \citep{peng2025commverify}. This is closely related to our concern that the same message may be recoverable in principle yet unusable in a deployed protocol. Our paper differs in two ways. First, we separate externalization, absorption, and action closure as distinct loci of coordination loss. Second, we tie these distinctions to same-message replay and sender-swap diagnostics, which enable us to localize receiver-side versus sender-side failures under a common step-level objective.

\paragraph{Protocol choice as a control problem.}
Across harness engineering, protocol surveys, and protocol-evaluation work, one recurring practical theme is that communication choices trade off capability, latency, and message budget \citep{du2025protocolbench,openai2025agents,anthropic2025harness}. Our closest connection to this theme is conceptual rather than architectural: we cast those choices as an inference-time control problem driven by receiver-relative gain, communication tax, and dynamic residual risk. The contribution is therefore not another routing architecture, but a common problem formulation that links modeling, diagnosis, and protocol selection for bounded receivers.

\paragraph{Error propagation and recovery.}
Agent failure studies and safety frameworks also model how local errors spread through multi-step or multi-agent interaction \citep{xiao2026layer, xiao2026adapting}. AgentDebug studies cascading failures through a modular error taxonomy and targeted feedback for recovery \citep{zhu2025agentfailures}, while GUARDIAN uses temporal graph modeling to detect and mitigate hallucination and error propagation in multi-agent collaborations \citep{zhou2025guardian}. Our conditional propagation law is complementary: it is a small analytic model for when a local residual gap carries forward or resynchronizes, rather than a full debugging or safety framework.

\section{Additional Theory and Proofs}
\label{app:theory}

This appendix provides full proofs and auxiliary results for Sections~\ref{sec:theory} and~\ref{sec:implications}; the subsections below are presented in the order in which their results are stated.

\begin{table}[!htbp]
\centering

\caption{\textbf{Notation roadmap.} Workflow primitives are shown in
Figure~\ref{fig:setup}; endpoint protocols are defined in Section~\ref{subsec:setup}. Several quantities listed here are defined in Sections~\ref{sec:theory} and~\ref{sec:implications}.}
\label{tab:notation}
\footnotesize
\setlength{\tabcolsep}{4pt}
\renewcommand{\arraystretch}{1.04}
\begin{tabular}{@{}l@{\ }l@{\hspace{10pt}}l@{\ }l@{}}
\toprule
$Z_t^\star$ & action-sufficient target &
$A_{C,\ell}^\star,\tau_{\rho,\ell},U_{\rho,\ell}$ & gain, tax, utility \\
$\Gamma_C,\mathcal I_{\rm omit},\Lambda_C$ & slack, omitted info, mismatch &
$L_{\rm ext}/L_{\rm abs}/L_{\rm close}$ & externalization / absorption / closure \\
$b,\Psi(b)$ & burden, upgrade value &
$D_t,\Delta_t,\lambda_t$ & residual, gap, carry \\
$k_t,q_t$ & fresh / resolved residual &
$\alpha,J_t(\alpha)$ & protocol action, selector value \\
$W_t, C_t, S_t^i, M_t$ & state, context, sender, message &
$X_t^j, \Phi_C, A_t^j$ & local view, decoder, next action \\

\bottomrule
\end{tabular}

\end{table}

\subsection{One-Step Utility and Representation-Aware Compression}
\label{app:one-step-proofs}

We first record a basic fact about the log-loss approximation slack.

\begin{lemma}[Slack is Nonnegative]
\label{lem:gamma-nonnegative}
For any conditioning variable $Y$ and any receiver class $\Phi_C$ of predictive distributions under log-loss,
\[
\Gamma_C(Y)
:=
\mathcal R_{C,\log}^\star(Z_t^\star\mid Y)
-
H(Z_t^\star\mid Y)
\ge 0.
\]
\end{lemma}

\begin{proof}
For any admissible predictor $q_\phi(\cdot\mid Y)$,
\begin{align*}
\E[-\log q_\phi(Z_t^\star\mid Y)]
&=
\E[-\log p(Z_t^\star\mid Y)]
+
\E\!\left[
\log \frac{p(Z_t^\star\mid Y)}{q_\phi(Z_t^\star\mid Y)}
\right] \\
&=
H(Z_t^\star\mid Y)
+
\E\!\left[
\mathrm{KL}\bigl(p(\cdot\mid Y)\,\|\,q_\phi(\cdot\mid Y)\bigr)
\right]\\
&\ge
H(Z_t^\star\mid Y).
\end{align*}
Taking the infimum over $\phi\in\Phi_C$ gives the claim.
\end{proof}

\begin{proof}[Proof of Proposition~\ref{thm:communication-tax}]
By definition,
\[
U_{\rho,\ell}(M_t;Z_t^\star)=R_\rho(0)-R_\rho(1).
\]
Add and subtract $R_C^\star(0)$ and $R_C^\star(1)$:
\begin{align*}
U_{\rho,\ell}(M_t;Z_t^\star)
&=
R_\rho(0)-R_\rho(1)
+R_C^\star(0)-R_C^\star(0)
+R_C^\star(1)-R_C^\star(1)\\
&=
\bigl(R_C^\star(0)-R_C^\star(1)\bigr)
-
\Bigl[
\bigl(R_\rho(1)-R_C^\star(1)\bigr)
-
\bigl(R_\rho(0)-R_C^\star(0)\bigr)
\Bigr].
\end{align*}
The first bracket is $A_{C,\ell}^\star(M_t;Z_t^\star)$ and the second bracket is $\tau_{\rho,\ell}(M_t)$.
\end{proof}

\begin{proof}[Proof of Theorem~\ref{thm:compression-dominance}]
By Lemma~\ref{lem:gamma-nonnegative} and the definition of $\Gamma_C$,
\[
\mathcal R_{C,\log}^\star(Z_t^\star\mid Y)
=
H(Z_t^\star\mid Y)+\Gamma_C(Y)
\]
for every conditioning variable $Y$ considered below.

Apply this identity to $Y=X_t^j$ and $Y=(X_t^j,M_t)$:
\begin{align*}
A_{C,\log}^\star(M_t;Z_t^\star)
&=
\mathcal R_{C,\log}^\star(Z_t^\star\mid X_t^j)
-
\mathcal R_{C,\log}^\star(Z_t^\star\mid X_t^j,M_t)\\
&=
H(Z_t^\star\mid X_t^j)-H(Z_t^\star\mid X_t^j,M_t)
+
\Gamma_C(X_t^j)-\Gamma_C(X_t^j,M_t)\\
&=
I(Z_t^\star;M_t\mid X_t^j)
+
\Gamma_C(X_t^j)-\Gamma_C(X_t^j,M_t).
\end{align*}

Similarly, conditioning on raw context gives
\[
A_{C,\log}^\star(C_t;Z_t^\star)
=
I(Z_t^\star;C_t\mid X_t^j)
+
\Gamma_C(X_t^j)-\Gamma_C(X_t^j,C_t).
\]
Subtracting the compressed-message gain from the raw-context gain yields
\begin{align*}
A_{C,\log}^\star(C_t;Z_t^\star)
-
A_{C,\log}^\star(M_t;Z_t^\star)
&=
I(Z_t^\star;C_t\mid X_t^j)
-
I(Z_t^\star;M_t\mid X_t^j)\\
&\quad+
\Gamma_C(X_t^j,M_t)-\Gamma_C(X_t^j,C_t).
\end{align*}
Because $M_t=f(C_t)$ is deterministic, the conditional chain rule gives
\[
I(Z_t^\star;C_t\mid X_t^j)
=
I(Z_t^\star;M_t\mid X_t^j)
+
I(Z_t^\star;C_t\mid X_t^j,M_t).
\]
Therefore
\[
A_{C,\log}^\star(C_t;Z_t^\star)
-
A_{C,\log}^\star(M_t;Z_t^\star)
=
I(Z_t^\star;C_t\mid X_t^j,M_t)
+
\Lambda_C(M_t\!\to\!C_t),
\]
where
\[
\Lambda_C(M_t\!\to\!C_t)
:=
\Gamma_C(X_t^j,M_t)-\Gamma_C(X_t^j,C_t).
\]

Apply Proposition~\ref{thm:communication-tax} to the compressed protocol and to the raw-context protocol:
\[
U_{\rho_{\rm cmp},\log}(M_t;Z_t^\star)
=
A_{C,\log}^\star(M_t;Z_t^\star)
-
\tau_{\rho_{\rm cmp},\log}(M_t),
\]
\[
U_{\rho_{\rm raw},\log}(C_t;Z_t^\star)
=
A_{C,\log}^\star(C_t;Z_t^\star)
-
\tau_{\rho_{\rm raw},\log}(C_t).
\]
Subtracting the second identity from the first gives
\begin{align*}
\Delta U_{\rm cmp-raw}
&=
U_{\rho_{\rm cmp},\log}(M_t;Z_t^\star)
-
U_{\rho_{\rm raw},\log}(C_t;Z_t^\star)\\
&=
\bigl(A_{C,\log}^\star(M_t;Z_t^\star)
-
A_{C,\log}^\star(C_t;Z_t^\star)\bigr)
+
\bigl(\tau_{\rho_{\rm raw},\log}(C_t)-\tau_{\rho_{\rm cmp},\log}(M_t)\bigr)\\
&=
\Delta\tau_{\rm raw\to cmp}
-
I(Z_t^\star;C_t\mid X_t^j,M_t)
-
\Lambda_C(M_t\!\to\!C_t),
\end{align*}
which is Eq.~\eqref{eq:cmp-vs-raw-gap}.
\end{proof}

\paragraph{Receiver-class representation gap.}
The omitted information and decoder-mismatch terms can be combined into a single bounded-receiver certificate. Define
\[
B_C(M_t,C_t)
:=
\mathcal R^\star_{C,\log}(Z_t^\star\mid X_t^j,M_t)
-
\mathcal R^\star_{C,\log}(Z_t^\star\mid X_t^j,C_t).
\]
This is the class-optimal log-loss gap between using the compressed message and using raw context.

\begin{proposition}[Receiver-Class Compression Certificate]
\label{prop:receiver-class-certificate}
Under log-loss, if $M_t=f(C_t)$, then
\[
B_C(M_t,C_t)
=
I(Z_t^\star;C_t\mid X_t^j,M_t)
+
\Lambda_C(M_t\!\to\!C_t).
\]
Consequently,
\[
\Delta U_{\rm cmp-raw}
=
\Delta\tau_{\rm raw\to cmp}
-
B_C(M_t,C_t).
\]
Thus, compressed communication dominates raw-context execution exactly when the tax saved by compression exceeds the bounded-receiver representation gap.
\end{proposition}

\begin{proof}
By the log-loss slack identity,
\begin{align*}
B_C(M_t,C_t)
&=
\Bigl[
H(Z_t^\star\mid X_t^j,M_t)+\Gamma_C(X_t^j,M_t)
\Bigr]\\
&\quad-
\Bigl[
H(Z_t^\star\mid X_t^j,C_t)+\Gamma_C(X_t^j,C_t)
\Bigr]\\
&=
H(Z_t^\star\mid X_t^j,M_t)
-
H(Z_t^\star\mid X_t^j,C_t)
+
\Lambda_C(M_t\!\to\!C_t).
\end{align*}
Because $M_t=f(C_t)$, the entropy difference equals
\[
H(Z_t^\star\mid X_t^j,M_t)
-
H(Z_t^\star\mid X_t^j,C_t)
=
I(Z_t^\star;C_t\mid X_t^j,M_t).
\]
This proves the first identity. The second identity follows by substituting this expression into Eq.~\eqref{eq:cmp-vs-raw-gap}.
\end{proof}

\begin{corollary}[Compatibility Condition for Compression Dominance]
\label{cor:compatibility}
If $B_C(M_t,C_t)\le \kappa_t$, then compressed communication dominates raw-context execution whenever
\[
\Delta\tau_{\rm raw\to cmp}>\kappa_t.
\]
Equivalently, if $\Lambda_C(M_t\!\to\!C_t)\le \bar\kappa_t$, then compression dominates whenever
\[
\Delta\tau_{\rm raw\to cmp}>
I(Z_t^\star;C_t\mid X_t^j,M_t)+\bar\kappa_t.
\]
\end{corollary}

\begin{proof}
The first statement is immediate from Proposition~\ref{prop:receiver-class-certificate}. The second follows by substituting
\[
B_C(M_t,C_t)
=
I(Z_t^\star;C_t\mid X_t^j,M_t)
+
\Lambda_C(M_t\!\to\!C_t).
\]
\end{proof}

\subsection{Bayes-Sufficient Compression and Receiver Compatibility}
\label{app:sufficiency-not-enough}

This subsection proves Theorem~\ref{thm:sufficiency-not-enough}. The result separates information sufficiency from receiver usability. A compressed message can preserve all target-relevant information for an unconstrained Bayes decoder while changing the representation geometry available to a bounded receiver class.

Operationally, a deployed LLM executor induces a bounded receiver class through an interface tuple
\[
\mathcal I=(\text{model},\text{tokenizer},\text{prompt template},
\text{schema/parser},\text{tool validator},L,\text{decoding policy}).
\]
A compression is receiver-compatible for this interface when the required artifact/action fields can be extracted and converted into a valid emitted action with small excess risk. Under this view, field names, action verbs, file paths, and tool schemas are part of the communication channel rather than cosmetic formatting. For example, a direct payload such as
\(\{\texttt{action\_type}:\texttt{edit},\texttt{target\_artifact}:\texttt{path}\}\)
may be compatible with an edit executor, while
\(\{\texttt{f1}:\texttt{A2},\texttt{f2}:\texttt{path}\}\)
can preserve the Bayes target but fail under the deployed parser or policy.

We use the following finite-alphabet proper-compression condition. A message $M=T(C)$ is an $\alpha$-proper Bayes-sufficient compression of raw context $C$ for target $Z$ if
\begin{equation}
I(Z;C\mid M)=0
\qquad\text{and}\qquad
H(M)\le \alpha H(C),
\label{eq:proper-bs-compression}
\end{equation}
for some $\alpha\in(0,1)$. The first condition is Bayes sufficiency. The second excludes both raw context and same-size recodings.

The entropy constraint is an analysis-level abstraction of the finite token caps used in the experiments. If a token sequence over vocabulary \(\mathcal V\) is hard-capped at length \(B\), its support size is at most \(\sum_{\ell=0}^{B}|\mathcal V|^\ell\), so \(H(M)\le B\log|\mathcal V|+\log(B+1)\). More generally, realized token lengths upper-bound entropy up to length-coding constants. The appendix experiments therefore report realized token counts and truncation rates; they are operational budget controls, not direct entropy estimates.

\begin{lemma}[Alternating Labels Are Hard for Thresholds]
\label{lem:alternating-threshold}
Let $M$ be uniform on $\{0,1,\ldots,2m-1\}$, and let the binary label alternate on the ordered message axis:
\[
Z(M)=
\begin{cases}
0, & M \text{ even},\\
1, & M \text{ odd}.
\end{cases}
\]
For the threshold class
\[
\Phi_{\rm thr}
=
\{h_\theta(M)=\mathbf 1\{M\ge \theta\}:\theta\in\mathbb R\},
\]
the optimal $0/1$ error is at least $(m-1)/(2m)$.
\end{lemma}

\begin{proof}
Any threshold is equivalent to choosing a cut index $k\in\{0,1,\ldots,2m\}$, predicting $0$ on the first $k$ ordered message values and $1$ on the remaining $2m-k$ values. If $k=2r$ is even, the first segment contains $r$ positive labels and the second segment contains $m-r$ negative labels, so the threshold makes $m$ mistakes. If $k=2r+1$ is odd, the first segment contains $r$ positive labels and the second segment contains $m-r-1$ negative labels, so it makes $m-1$ mistakes. The boundary cases $k=0$ and $k=2m$ each make $m$ mistakes. Therefore, each threshold incurs at least $m-1$ mistakes among the $2m$ equiprobable states.
\end{proof}

\begin{proof}[Proof of Theorem~\ref{thm:sufficiency-not-enough}]
Fix any compression ratio $\alpha\in(0,1)$ and any $\varepsilon>0$. Choose $m\ge 2$ large enough that
\[
\frac{m-1}{2m}\ge \frac12-\varepsilon.
\]
Let
\[
Y\sim \mathrm{Unif}\{0,1,\ldots,2m-1\},
\qquad
U\sim \mathrm{Unif}\{1,\ldots,K\},
\]
independently, and define the raw context
\[
C=(Y,U).
\]
The nuisance coordinate $U$ carries no target information but contributes to raw-context size. Let
\[
Z:=\mathbf 1\{Y\ge m\}.
\]

The bounded receiver class is representation-sensitive. It thresholds the visible scalar coordinate delivered by the protocol. In the raw context, the visible coordinate is the natural coordinate $Y$; on a compressed scalar message, the visible coordinate is the message value. Formally, define
\[
\operatorname{coord}(Y,U)=Y,
\qquad
\operatorname{coord}(M)=M,
\]
and let
\[
\Phi_{\rm thr}
=
\left\{
h_\theta(v)=\mathbf 1\{\operatorname{coord}(v)\ge \theta\}:\theta\in\mathbb R
\right\}.
\]
Thus, the raw context is decoded by thresholding $Y$, whereas the compressed message is decoded by thresholding the delivered surface value $M$.

On raw context, the threshold $\theta^\star=m-\frac12$ achieves zero error:
\[
h_{\theta^\star}(Y,U)=\mathbf 1\{Y\ge m\}=Z.
\]
Therefore
\[
\mathcal R_{\Phi_{\rm thr},0/1}^\star(Z\mid C)=0.
\]

Now define a recoding $f:\{0,\ldots,2m-1\}\to\{0,\ldots,2m-1\}$ by
\[
f(i)=2i,\qquad i=0,\ldots,m-1,
\]
and
\[
f(m+i)=2i+1,\qquad i=0,\ldots,m-1.
\]
The compressed message drops the nuisance coordinate and sends
\[
M:=T(C):=f(Y).
\]

Because $f$ is a bijection on $Y$, the value of $Y$ is a deterministic function of $M$. Since $Z$ is a deterministic function of $Y$, we have
\[
I(Z;C\mid M)=0.
\]
Thus $M$ is Bayes-sufficient.

The message is also properly compressed. Since $f$ is bijective on $Y$,
\[
H(M)=H(Y)=\log(2m).
\]
Because $Y$ and $U$ are independent,
\[
H(C)=H(Y,U)=\log(2m)+\log K.
\]
Choose
\[
K\ge \left\lceil (2m)^{(1-\alpha)/\alpha}\right\rceil.
\]
Then
\[
H(M)
=
\log(2m)
\le
\alpha\bigl(\log(2m)+\log K\bigr)
=
\alpha H(C).
\]
Thus $M$ is an $\alpha$-proper Bayes-sufficient compression.

It remains to be shown that the error on the compressed representation is lower-bounded. Under the ordered message values $M=0,1,\ldots,2m-1$, labels alternate:
\[
Z=0,1,0,1,\ldots,0,1.
\]
The negative states $Y=0,\ldots,m-1$ map to even message values, while the positive states $Y=m,\ldots,2m-1$ map to odd message values. By Lemma~\ref{lem:alternating-threshold},
\[
\mathcal R_{\Phi_{\rm thr},0/1}^\star(Z\mid M)
\ge
\frac{m-1}{2m}
\ge
\frac12-\varepsilon.
\]
Thus, the raw context is decoded perfectly by the bounded receiver class, while an $\alpha$-proper Bayes-sufficient compressed message has error arbitrarily close to balanced binary chance error.
\end{proof}

\paragraph{Receiver-compatible sufficiency.}
Theorem~\ref{thm:sufficiency-not-enough} shows that Bayes sufficiency alone does not imply usability for a bounded receiver. The missing condition is receiver compatibility.

\begin{definition}[Receiver-Compatible Sufficient Compression]
\label{def:receiver-compatible-sufficient}
A Bayes-sufficient compression $M_t=f(C_t)$ is $\kappa$-receiver-compatible for class $\Phi_C$ if
\[
B_C(M_t,C_t)
=
\mathcal R^\star_{C,\log}(Z_t^\star\mid X_t^j,M_t)
-
\mathcal R^\star_{C,\log}(Z_t^\star\mid X_t^j,C_t)
\le \kappa.
\]
\end{definition}

\begin{corollary}[Sufficient Compression Wins When Compatibility Beats Tax]
\label{cor:receiver-compatible-sufficiency}
If $M_t=f(C_t)$ is Bayes-sufficient for $Z_t^\star$ given $X_t^j$, then
\[
I(Z_t^\star;C_t\mid X_t^j,M_t)=0
\]
and
\[
\Delta U_{\rm cmp-raw}
=
\Delta\tau_{\rm raw\to cmp}
-
\Lambda_C(M_t\!\to\!C_t).
\]
More generally, if $M_t$ is $\kappa$-receiver-compatible, compressed communication dominates raw context whenever
\[
\Delta\tau_{\rm raw\to cmp}>\kappa.
\]
\end{corollary}

\begin{proof}
Bayes sufficiency sets the omitted-information term in Eq.~\eqref{eq:cmp-vs-raw-gap} to zero, leaving only the decoder-mismatch term. The general statement follows from Proposition~\ref{prop:receiver-class-certificate}: if $B_C(M_t,C_t)\le\kappa$, then
\[
\Delta U_{\rm cmp-raw}
=
\Delta\tau_{\rm raw\to cmp}
-
B_C(M_t,C_t)
\ge
\Delta\tau_{\rm raw\to cmp}-\kappa.
\]
Thus, compression beats raw context whenever the saved tax exceeds the receiver-class representation gap.
\end{proof}

\begin{remark}[Design implication]
The useful target is not Bayes sufficiency alone, but receiver-compatible sufficiency: the message should preserve the action-sufficient target, reduce protocol tax, and present the target in a surface form usable by the bounded receiver.
\end{remark}

\subsection{Bottleneck Decomposition and Fixed-Message Replay}
\label{app:bottleneck-proofs}

We now prove the loss decomposition and replay result. Throughout this subsection, condition on $X_t^j$ unless stated otherwise. The bottleneck-factorized receiver satisfies the conditional-independence chain
\[
Z_t^\star \to C_t \to M_t \to \widehat Z_t \to A_t^j.
\]

\begin{proof}[Proof of Theorem~\ref{thm:bottleneck-chain}]
Define the adjacent drops
\[
L_{\rm ext}
:=
I(Z_t^\star;C_t\mid X_t^j)-I(Z_t^\star;M_t\mid X_t^j),
\]
\[
L_{\rm abs}
:=
I(Z_t^\star;M_t\mid X_t^j)-I(Z_t^\star;\widehat Z_t\mid X_t^j),
\]
and
\[
L_{\rm close}
:=
I(Z_t^\star;\widehat Z_t\mid X_t^j)-I(Z_t^\star;A_t^j\mid X_t^j).
\]
By the data-processing inequality applied along
\[
Z_t^\star \to C_t \to M_t \to \widehat Z_t \to A_t^j
\]
conditional on $X_t^j$, each adjacent drop is nonnegative. Summing the adjacent drops gives the telescoping identity
\[
I(Z_t^\star;C_t\mid X_t^j)-I(Z_t^\star;A_t^j\mid X_t^j)
=
L_{\rm ext}+L_{\rm abs}+L_{\rm close}.
\]
The left-hand side is $L_{\rm total}$.
\end{proof}

\begin{proof}[Proof of Proposition~\ref{prop:replay-isolation}]
For a receiver $r$ evaluated on the fixed triple $(X_t,C_t,M_t)$, Theorem~\ref{thm:bottleneck-chain} gives
\[
I(Z_t^\star;A_t^{r}\mid X_t)
=
I(Z_t^\star;M_t\mid X_t)
-
L_{\rm abs}^{r}
-
L_{\rm close}^{r}.
\]
The term $I(Z_t^\star;M_t\mid X_t)$ is shared by all receivers in a fixed-message replay setting. Therefore, for two receivers $r,r'$,
\begin{align*}
I(Z_t^\star;A_t^{r}\mid X_t)
-
I(Z_t^\star;A_t^{r'}\mid X_t)
&=
\bigl(L_{\rm abs}^{r'}+L_{\rm close}^{r'}\bigr)
-
\bigl(L_{\rm abs}^{r}+L_{\rm close}^{r}\bigr).
\end{align*}
Thus, sender-side externalization cancels in the paired comparison, leaving receiver-side absorption and action closure.
\end{proof}

\paragraph{Observable factorization.}
When $Z_t^\star=(Z_t^{\rm art},Z_t^{\rm act})$, this motivates the observable $0/1$ factorization used in Section~\ref{sec:experiments}: a diagnostic, not an MI estimate.

\subsection{Sender-Gap Identity and Burden Threshold}
\label{app:alignment-proofs}

\begin{figure}[!htbp]
\centering
\resizebox{\linewidth}{!}{%
\begin{tikzpicture}[font=\scriptsize, >=Latex]
\tikzset{
  panel/.style={draw=black!25, rounded corners=3pt, fill=black!1},
  head/.style={font=\bfseries\small},
  sourcebox/.style={draw=purple!45!black, fill=purple!8, rounded corners=3pt,
                    align=center, inner sep=2.5pt,
                    minimum width=2.4cm, minimum height=0.50cm},
  gainbox/.style={draw=blue!55!black, fill=blue!8, rounded corners=3pt,
                  align=center, inner sep=2.5pt,
                  minimum width=2.4cm, minimum height=0.78cm},
  taxbox/.style={draw=red!60!black, fill=red!8, rounded corners=3pt,
                 align=center, inner sep=2.5pt,
                 minimum width=2.4cm, minimum height=0.78cm},
  netbox/.style={draw=black!45, fill=black!3, rounded corners=3pt,
                 align=center, inner sep=2.5pt,
                 minimum width=2.9cm, minimum height=0.50cm},
  regbad/.style={draw=red!50!black, fill=red!7, rounded corners=3pt,
                 align=center, inner sep=2.5pt,
                 minimum width=2.4cm, minimum height=0.50cm},
  reggood/.style={draw=green!40!black, fill=green!7, rounded corners=3pt,
                  align=center, inner sep=2.5pt,
                  minimum width=2.4cm, minimum height=0.50cm}
}

\begin{scope}[shift={(0,0)}]
\node[panel, minimum width=6.40cm, minimum height=4.40cm,
      anchor=south west] (L) at (0,0) {};

\node[anchor=north, head] at (3.20,4.34)
  {Stronger sender: help or hurt?};

\node[sourcebox] (src) at (3.20,3.55) {stronger sender $s'$};

\node[gainbox] (gain) at (1.75,2.55)
  {richer signal\\[1pt]$\Rightarrow\;+\Delta A(b)$ gain};
\node[taxbox]  (tax)  at (4.65,2.55)
  {harder surface\\[1pt]$\Rightarrow\;+\Delta\tau(b)$ tax};

\node[netbox] (net) at (3.20,1.45)
  {net $\Psi(b)=\Delta A{-}\Delta\tau$};

\node[regbad]  (lb) at (1.75,0.55)
  {$b{<}b^\star$: tax wins};
\node[reggood] (hb) at (4.65,0.55)
  {$b{>}b^\star$: gain wins};

\draw[->, semithick, blue!55!black] (src.south)  -| (gain.north);
\draw[->, semithick, red!60!black]  (src.south)  -| (tax.north);
\draw[->, semithick, blue!55!black] (gain.south) -- (net.north west);
\draw[->, semithick, red!60!black]  (tax.south)  -- (net.north east);
\draw[->, thin, red!50!black, dashed]   (net.south) -| (lb.north);
\draw[->, thin, green!40!black, dashed] (net.south) -| (hb.north);
\end{scope}

\begin{scope}[shift={(6.70,0)}]
\node[panel, minimum width=9.20cm, minimum height=4.40cm,
      anchor=south west] (R) at (0,0) {};

\node[anchor=north, head] at (4.60,4.34)
  {Net upgrade value $\Psi(b)$};

\begin{axis}[
  at={(0.55cm,0.45cm)},
  scale only axis=true,
  width=8.10cm, height=3.45cm,
  axis on top, axis lines=middle,
  axis line style={-Latex, line width=0.7pt, black!80},
  xlabel={burden $b$}, ylabel={value},
  xlabel style={at={(axis description cs:1.0,0.45)}, anchor=west, font=\scriptsize},
  ylabel style={at={(axis description cs:0.04,1.0)}, anchor=south, font=\scriptsize},
  xtick={0,2.5}, xticklabels={$0$,$b^\star$},
  xticklabel style={font=\scriptsize, anchor=north},
  ytick={-1.5,0,1.5}, yticklabels={$-$,$0$,$+$},
  yticklabel style={font=\scriptsize, anchor=east},
  ymin=-1.7, ymax=2.05, xmin=-0.05, xmax=6.30,
  domain=0:5.5, samples=130, smooth,
  clip mode=individual, no markers,
]
\fill[gray!9]   (axis cs:0,-1.7)   rectangle (axis cs:2.5,2.05);
\fill[cyan!11]  (axis cs:2.5,-1.7) rectangle (axis cs:6.30,2.05);
\draw[dashed, gray!55, thin] (axis cs:0,0) -- (axis cs:6.20,0);
\addplot[draw=none, fill=red!16,   domain=0:2.5]
  {(0.25+1.55/(1+exp(-0.9*(x-2.2))))-(0.65+0.85*exp(-0.45*(x-1.4)^2))} \closedcycle;
\addplot[draw=none, fill=green!22, domain=2.5:5.5]
  {(0.25+1.55/(1+exp(-0.9*(x-2.2))))-(0.65+0.85*exp(-0.45*(x-1.4)^2))} \closedcycle;
\addplot[densely dashed, blue!55!black, line width=1.0pt]
  {0.25+1.55/(1+exp(-0.9*(x-2.2)))};
\addplot[densely dotted, red!65!black, line width=1.1pt]
  {0.65+0.85*exp(-0.45*(x-1.4)^2)};
\addplot[solid, black, line width=1.7pt]
  {(0.25+1.55/(1+exp(-0.9*(x-2.2))))-(0.65+0.85*exp(-0.45*(x-1.4)^2))};
\node[font=\scriptsize, blue!55!black, anchor=west,
      fill=white, inner sep=1pt, draw=none] at (axis cs:5.55,1.73) {$\Delta A(b)$};
\node[font=\scriptsize, black, anchor=west,
      fill=white, inner sep=1pt, draw=none] at (axis cs:5.55,1.05) {$\Psi(b)$};
\node[font=\scriptsize, red!65!black,  anchor=west,
      fill=white, inner sep=1pt, draw=none] at (axis cs:5.55,0.66) {$\Delta\tau(b)$};
\draw[dashed, gray!60] (axis cs:2.5,-1.5) -- (axis cs:2.5,0);
\fill[black] (axis cs:2.5,0) circle (2.8pt);
\fill[white] (axis cs:2.5,0) circle (1.2pt);
\node[font=\scriptsize, gray!55!black, anchor=north,
      fill=white, draw=gray!30, line width=0.3pt,
      rounded corners=1.5pt, inner sep=1.8pt]
  at (axis cs:1.25,1.95) {tax wins ($\Psi{<}0$)};
\node[font=\scriptsize, cyan!40!black, anchor=north,
      fill=white, draw=cyan!30, line width=0.3pt,
      rounded corners=1.5pt, inner sep=1.8pt]
  at (axis cs:4.40,1.95) {gain wins ($\Psi{>}0$)};
\end{axis}
\end{scope}

\end{tikzpicture}%
}
\caption{\textbf{Sender-upgrade threshold.} A stronger upstream agent $s'$ raises both extra gain (richer signal, blue) and extra tax (harder surface, red); under single-crossing of $\Psi(b){=}\Delta A(b){-}\Delta\tau(b)$, the upgrade helps only above the burden threshold $b^\star$ (Theorem~\ref{prop:alignment-threshold}).}
\label{fig:sender-threshold}
\vspace{-10pt}
\end{figure}
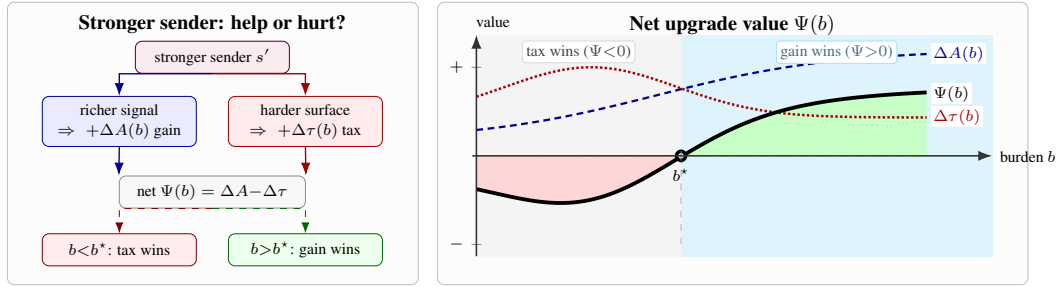

\begin{proof}[Proof of Theorem~\ref{prop:alignment-threshold}]
For a fixed executor $R$, apply Proposition~\ref{thm:communication-tax} to the sender $s$ and to the stronger sender $s'$:
\[
U_{s,R,\ell}
=
A_{R,\ell}^\star(M_t^{s};Z_t^\star)
-
\tau_{s,R,\ell}(M_t^{s}),
\]
\[
U_{s',R,\ell}
=
A_{R,\ell}^\star(M_t^{s'};Z_t^\star)
-
\tau_{s',R,\ell}(M_t^{s'}).
\]
Subtracting gives
\[
U_{s',R,\ell}-U_{s,R,\ell}
=
\Delta A_{R,\ell}(s,s')-\Delta\tau_{R,\ell}(s,s').
\]

Now let $\{R_b:b\in\mathcal B\}$ be a burden-indexed receiver family and define
\[
\Psi_{s,s'}(b)
:=
U_{s',R_b,\ell}-U_{s,R_b,\ell}.
\]
If $\Psi_{s,s'}$ is continuous and strictly single-crossing, its sign changes at most once. If it is negative at low burden and positive at high burden, the intermediate value theorem gives a crossing $b^\star$ with $\Psi_{s,s'}(b^\star)=0$, and strict single-crossing makes this crossing unique since the stronger sender helps exactly when $\Psi_{s,s'}(b)>0$, it helps exactly on the high-burden side of $b^\star$.
\end{proof}

\subsection{State Propagation}
\label{app:state-prop-proofs}

\begin{proof}[Proof of Proposition~\ref{prop:conditional-propagation}]
The residual dynamics are
\[
D_{u+1}=k_u+\lambda_u[D_u-q_u]_+,
\qquad
\widetilde D_{u+1}=k_u+\lambda_u[\widetilde D_u-q_u]_+.
\]
If $D_u>q_u$ and $\widetilde D_u>q_u$, then clipping is inactive:
\[
[D_u-q_u]_+=D_u-q_u,
\qquad
[\widetilde D_u-q_u]_+=\widetilde D_u-q_u.
\]
Therefore
\[
\Delta_{u+1}
=
\widetilde D_{u+1}-D_{u+1}
=
\lambda_u(\widetilde D_u-D_u)
=
\lambda_u\Delta_u.
\]
Iterating for $u=t,\ldots,t+h-1$ gives
\[
\Delta_{t+h}
=
\left(\prod_{u=t}^{t+h-1}\lambda_u\right)\Delta_t.
\]
If clipping becomes active for either trajectory, the difference is no longer propagated linearly; this is the resynchronization regime, as represented by the clipped residual model. For vector-valued residuals, the same argument applies componentwise when the carry condition holds componentwise.
\end{proof}

\subsection{Protocol-Control Regret Bound}
\label{app:selector-regret}

This subsection records the elementary plug-in guarantee for the deployed protocol controller. Let
\[
\alpha_t^\star\in\argmax_{\alpha\in\mathcal A_t}J_t(\alpha)
\]
be the ideal action under the analysis objective, and let
\[
\widehat\alpha_t\in\argmax_{\alpha\in\mathcal A_t}\widehat J_t(\alpha)
\]
be the deployed action chosen by calibrated estimates.

\begin{proposition}[Plug-In Selector Regret]
\label{prop:plugin-selector-regret}
Suppose that for every $\alpha\in\mathcal A_t$,
\[
|\widehat J_t(\alpha)-J_t(\alpha)|\le \epsilon.
\]
Then the one-step value regret of the deployed selector satisfies
\[
J_t(\alpha_t^\star)-J_t(\widehat\alpha_t)\le 2\epsilon.
\]
\end{proposition}

\begin{proof}
By optimality of $\widehat\alpha_t$ under $\widehat J_t$,
\[
\widehat J_t(\widehat\alpha_t)\ge \widehat J_t(\alpha_t^\star).
\]
Using the uniform error bound twice,
\[
J_t(\alpha_t^\star)
\le
\widehat J_t(\alpha_t^\star)+\epsilon
\le
\widehat J_t(\widehat\alpha_t)+\epsilon
\le
J_t(\widehat\alpha_t)+2\epsilon.
\]
Rearranging gives the claim.
\end{proof}

\begin{corollary}[Termwise Calibration Error]
\label{cor:termwise-selector-regret}
Suppose the deployed score uses
\[
\widehat J_t(\alpha)
=
\widehat A_t(\alpha)
-
\widehat\tau_t(\alpha)
-
\eta_c \widehat c(\alpha)
-
\eta_r \widehat r_t(\alpha),
\]
and for every $\alpha\in\mathcal A_t$,
\[
|\widehat A_t(\alpha)-A_t(\alpha)|\le \epsilon_A,\qquad
|\widehat\tau_t(\alpha)-\tau_t(\alpha)|\le \epsilon_\tau,
\]
\[
|\widehat c(\alpha)-c(\alpha)|\le \epsilon_c,\qquad
|\widehat r_t(\alpha)-r_t(\alpha)|\le \epsilon_r.
\]
Then
\[
J_t(\alpha_t^\star)-J_t(\widehat\alpha_t)
\le
2\bigl(\epsilon_A+\epsilon_\tau+\eta_c\epsilon_c+\eta_r\epsilon_r\bigr).
\]
\end{corollary}

\begin{proof}
The termwise bounds imply
\[
|\widehat J_t(\alpha)-J_t(\alpha)|
\le
\epsilon_A+\epsilon_\tau+\eta_c\epsilon_c+\eta_r\epsilon_r
\]
uniformly over $\alpha$. Apply Proposition~\ref{prop:plugin-selector-regret}.
\end{proof}

\subsection{Idealized Specializations}
\label{app:idealized-specializations}

These two specializations connect the receiver-relative accounting to factorized losses and conditional mutual information under the stated assumptions.

\begin{proposition}[Factorized Target under Additive Loss]
\label{prop:factorized-target}
Suppose $Z_t^\star=(Z_t^{\rm art},Z_t^{\rm act})$, the loss decomposes additively across the two factors, and the admissible receiver class factorizes into independent artifact and action predictors. Then the class-optimal risk and the receiver-relative gain decompose into artifact and action components.
\end{proposition}

\begin{proof}
Under additive loss and a product receiver class, optimizing the joint predictor separates into optimizing the artifact predictor and the action predictor. The class-optimal joint risk is therefore the sum of the two class-optimal factor risks. Subtracting the no-message and message-conditioned versions gives the same additive decomposition for the receiver-relative gain.
\end{proof}

\begin{proposition}[Variational Lower Bound on Idealized Benefit]
\label{prop:variational-benefit}
Under the zero-slack specialization,
\[
A_{C,\log}^\star(M_t;Z_t^\star)
=
I(M_t;Z_t^\star\mid X_t^j).
\]
Hence, for any uniformly bounded critic class $\mathcal T$,
\[
A_{\mathcal T}(M_t;Z_t^\star\mid X_t^j)
:=
\sup_{T\in\mathcal T}
\left\{
\E_{p(X,M,Z)}[T]
-
\log
\E_{p(X)p(M\mid X)p(Z\mid X)}[e^T]
\right\}
\le
A_{C,\log}^\star(M_t;Z_t^\star).
\]
Consequently,
\[
\tau_{\rho,\log}(M_t)
\ge
A_{\mathcal T}(M_t;Z_t^\star\mid X_t^j)
-
U_{\rho,\log}(M_t;Z_t^\star).
\]
\end{proposition}

\begin{proof}
Under zero slack, the log-loss receiver-relative gain equals conditional mutual information:
\[
A_{C,\log}^\star(M_t;Z_t^\star)
=
I(M_t;Z_t^\star\mid X_t^j).
\]
Conditional mutual information is the KL divergence
\[
I(M_t;Z_t^\star\mid X_t^j)
=
\mathrm{KL}\!\left(
p(X_t^j,M_t,Z_t^\star)
\;\middle\|\;
p(X_t^j)p(M_t\mid X_t^j)p(Z_t^\star\mid X_t^j)
\right).
\]
The Donsker--Varadhan representation of KL, restricted to the uniformly bounded critic class $\mathcal T$, gives the stated lower bound. The tax certificate follows by rearranging Proposition~\ref{thm:communication-tax}:
\[
\tau_{\rho,\log}(M_t)
=
A_{C,\log}^\star(M_t;Z_t^\star)
-
U_{\rho,\log}(M_t;Z_t^\star).
\]
\end{proof}
\section{Additional Experimental Details and Results}
\label{app:exp-details}

This appendix extends the three main experimental claims with setup details, robustness checks, cross-family ablations, protocol-level tax knobs, and shared-state propagation studies. The main text contains the central identification results; this appendix adds the supporting evaluation sets, alternative protocols, and larger family sweeps that are useful for stress-testing the same theory under more varied conditions.

Each table states its evaluation support. Table~\ref{tab:main-results} gives the Qwen main evaluation counts, while appendix audits use their stated supports. Paired analyses use common \((\mathrm{instance},\mathrm{step})\) keys. OpenAI protocol accuracies use cell-specific counts, which can differ across conditions; those marginal rates are reported with Wilson intervals. Comparisons between tables retain each table's protocol and support definitions.

Appendix~\ref{app:benchmark-summary} records the benchmark and protocol summary.
Appendix~\ref{app:claim1-extra} adds robustness checks, token-accounting and no-overflow controls, strict-recoding diagnostics, gain-surrogate diagnostics, and additional raw/stronger-receiver controls.
Appendix~\ref{app:claim2-extra} adds replay variants and additional closure plots.
Appendix~\ref{app:claim3-extra} adds sender-replacement and cross-family threshold analyses.
Appendix~\ref{app:cross-family} collects the broader OpenAI and weak-family ablations.
Appendix~\ref{app:tax-controller} reports the fuller protocol-knob, static-baseline, and selector results beyond Figure~\ref{fig:tax-knobs-main}.
Appendix~\ref{app:shared-state} reports the shared-state perturbation and repair analyses behind Proposition~\ref{prop:conditional-propagation}.

\subsection{Benchmark and Protocol Summary}
\label{app:benchmark-summary}

\begin{table}[t]
\centering
\caption{\textbf{Benchmark and protocol summary.} The table centralizes the default sender/receiver pair, message budget, and role of each benchmark in the paper's experimental program.}
\label{tab:benchmark-summary}
\small
\resizebox{\columnwidth}{!}{%
\begin{tabular}{lcccll}
\toprule
Benchmark & Sender & Receiver & Budget & Main role & Note \\
\midrule
BFCL & Qwen3-Next-80B-A3B & Qwen3.5-9B & 128 & noise / high suff. & Main paired set. \\
ToolSandbox & Qwen3-Next-80B-A3B & Qwen3.5-9B & 128 & noise / high suff. & Main paired set. \\
SWE-Lite & Qwen3-Coder-Next-FP8 & Qwen3.5-9B & 128 & capability transfer & Main paired set. \\
$\tau$-bench & Qwen3-Next-80B-A3B & Qwen3.5-9B & 128 & weakly helpful / boundary & Main paired set. \\
ContextBench & Qwen3-Next-80B-A3B & Qwen3.5-9B & 128 & helpful / context-heavy & Main paired set. \\
RepoBench & Qwen3-Next-80B-A3B & Qwen3.5-9B & 128 & helpful / context-heavy & Main paired set. \\
LongCLI & Qwen3-Next-80B-A3B & Qwen3.5-9B & 128 & shared-state stress & Appendix trajectory set. \\
\bottomrule
\end{tabular}%
}
\end{table}

The table is the bookkeeping layer behind the main-text experimental setup.
The first six benchmarks support the three main empirical claims; LongCLI is
reserved for the dynamic shared-state analyses in Appendix~\ref{app:shared-state}.
Unless otherwise stated, runs use the default Qwen3-Next-80B-A3B sender, the
Qwen3.5-9B receiver, and a $128$-token message budget on the support stated in
the corresponding table. SWE-Lite uses Qwen3-Coder-Next-FP8 for the main MA
sender condition, as listed in Table~\ref{tab:benchmark-summary}.

\begin{table}[t]
\centering
\caption{\textbf{Benchmark-specific construction of action-sufficient targets.}
The benchmarks do not share a native next-action label space. For scoring only, we project each benchmark's native supervision into a set
$G_t=\{(Z_{t,k}^{\rm art},Z_{t,k}^{\rm act})\}_{k=1}^{K_t}$ of acceptable artifact/action targets.
\texttt{hit\_joint} is one iff the parsed receiver output matches at least one pair in $G_t$.
\texttt{hit\_artifact} matches the artifact component, and $h_{\rm action}$ is action correctness conditional on artifact correctness.
Sender, receiver, and selector never observe $G_t$.}
\label{tab:target-construction}
\footnotesize
\setlength{\tabcolsep}{4pt}
\renewcommand{\arraystretch}{1.10}
\begin{tabular}{@{}p{0.13\textwidth}p{0.25\textwidth}p{0.30\textwidth}p{0.25\textwidth}@{}}
\toprule
Benchmark & Native supervision & Scorer projection & Scoring rule \\
\midrule

BFCL
& canonical function-call answer
& artifact: function/API name; action: \texttt{tool-call} or \texttt{no-call}
& match canonical function identity and action type; native BFCL evaluation separately normalizes arguments \\

ToolSandbox
& scenario milestones over tool traces and world-state changes
& artifact: valid milestone target; action: tool call, state update, add/remove, or user response
& set-valued match against the current valid milestone frontier \\

ContextBench
& human-verified gold code context, plus our harness-stage repair label when used
& artifact: gold file, block, function, or line span; action: retrieval/read, or repair-stage edit
& normalized overlap with gold context, or normalized target-file match for repair-stage diagnostics \\

RepoBench
& retrieval target, next-line completion target, or retrieval+completion pipeline target
& artifact: snippet index, candidate file, or cursor location; action: retrieval or completion
& retrieval match for RepoBench-R/P; completion match for RepoBench-C/P \\

SWE-bench Lite
& issue, gold patch, test patch, \texttt{FAIL\_TO\_PASS}, \texttt{PASS\_TO\_PASS}
& artifact: gold-patch file/module or test ID; action: patch-stage edit or harness-stage test
& scorer-side projection from gold patch and test metadata; final correctness remains SWE-bench patch-and-test execution \\

$\tau$-bench
& user task, domain tools/policy, action trace, and final goal database state
& artifact: tool, entity, state field, or goal-state diff; action: tool call, state update, or response
& set-valued projection from accepted tool/action traces or goal-state differences \\

LongCLI
& task metadata, project specification, reference solution, and step-level test outcomes
& artifact: task folder, command target, file, workspace artifact, or requirement; action: command, edit-feature, or test
& normalized match to the reference command target, task folder, or step-level requirement target \\

\bottomrule
\end{tabular}
\end{table}

\FloatBarrier
\paragraph{Concrete scorer-side mapping cases.}
The following boxes show one real benchmark row or trace per benchmark and how its native supervision is projected into the artifact/action vocabulary used by our scorer.

\begin{casebox}[label=case:map-bfcl]{BFCL: canonical function-call row}
\textbf{Native benchmark row.}
\texttt{simple\_python\_0}: the user asks for the area of a triangle with base $10$ and height $5$.

\medskip
\textbf{Native target source.}
The canonical BFCL answer calls:
\[
\texttt{calculate\_triangle\_area(base=10,\ height=5,\ unit=\dots)}.
\]

\medskip
\textbf{Projected scorer target.}
For the artifact/action scorer, we project the row to
\[
G_t=\{(\texttt{calculate\_triangle\_area},\texttt{tool-call})\}.
\]
The argument values are retained by the native BFCL function-call evaluator, but the paper's
$Z_t^{\rm art}/Z_t^{\rm act}$ projection uses the function identity and action type.

\medskip
\textbf{Scoring examples.}
An output that emits \texttt{calculate\_triangle\_area(...)} has \texttt{hit\_artifact}=1 and \texttt{hit\_joint}=1 under this projection.
An output that emits no call has \texttt{hit\_action}=0 and \texttt{hit\_joint}=0.
\end{casebox}

\begin{casebox}[label=case:map-toolsandbox]{ToolSandbox: weather-tool milestone}
\textbf{Native benchmark trace.}
Scenario family: \texttt{find\_temperature}. The user asks: ``What's the temperature here right now?''

\medskip
\textbf{Native target source.}
The scenario exposes the weather tool \texttt{search\_weather\_around\_lat\_lon}, and the valid milestone frontier includes a tool-trace milestone that requires this weather lookup at the current location.

\medskip
\textbf{Projected scorer target.}
For the tool-call step, we project the current milestone frontier to
\[
G_t=\{(\texttt{search\_weather\_around\_lat\_lon},\texttt{tool-call})\}.
\]
If the scorer is evaluating the final user-response milestone rather than the tool-call milestone, the same scenario can instead project to a response target such as
\[
G_t=\{(\texttt{temperature},\texttt{user-response})\}.
\]

\medskip
\textbf{Scoring examples.}
An output \texttt{tool-call/search\_weather\_around\_lat\_lon} is a joint hit for the tool-call milestone.
An output \texttt{read/messages} is an artifact/action miss for that milestone, even if it is a plausible natural-language protocol step.
\end{casebox}

\begin{casebox}[label=case:map-contextbench]{ContextBench: Django cascade-deletion case}
\textbf{Native benchmark case.}
Django cascade-deletion bug from the evaluated ContextBench set. The issue concerns a \texttt{UnicodeDecodeError} during \texttt{.delete()} because cascade collection fetches unnecessary fields with invalid UTF-8.

\medskip
\textbf{Native target source.}
The gold context/repair target identifies the cascade-collection implementation path:
\[
\texttt{django/db/models/deletion.py}
\]
and the relevant collector logic, including \texttt{Collector.collect}.

\medskip
\textbf{Projected scorer targets.}
For context-retrieval scoring:
\[
G_t^{\rm retrieve}
=
\{(\texttt{django/db/models/deletion.py},\texttt{retrieve-context})\}.
\]
For the repair-stage diagnostic used in our agent harness:
\[
G_t^{\rm repair}
=
\{(\texttt{django/db/models/deletion.py},\texttt{edit})\}.
\]

\medskip
\textbf{Scoring examples.}
An output \texttt{edit/django/db/models/deletion.py} is a joint hit under the repair-stage projection.
An output \texttt{edit/django/db/models/base.py} has the right action type but the wrong artifact, so \texttt{hit\_joint}=0.
\end{casebox}

\begin{casebox}[label=case:map-repobench]{RepoBench: \texttt{sastix/cms} retrieval/completion row}
\textbf{Native benchmark row.}
RepoBench-R Java row with
\[
\texttt{repo\_name=sastix/cms}
\]
and file \path{server/src/main/java/com/sastix/cms/server/services/content/impl/ZipHandlerServiceImpl.java}.
The row's next line is:
\[
\texttt{HashedDirectoryService hashedDirectoryService;}
\]
and the native retrieval target is \texttt{gold\_snippet\_index=5}.

\medskip
\textbf{Native target source.}
RepoBench-R supplies the gold snippet index for retrieval. RepoBench-C/P additionally evaluates next-line completion at the cursor.

\medskip
\textbf{Projected scorer targets.}
For retrieval:
\[
G_t^{\rm retrieve}
=
\{(\texttt{context[5]},\texttt{retrieve-snippet})\}.
\]
For completion:
\[
G_t^{\rm complete}
=
\{(\texttt{ZipHandlerServiceImpl.java@cursor},\texttt{complete-line})\}.
\]

\medskip
\textbf{Scoring examples.}
A retrieval output that selects \texttt{context[5]} is an artifact hit for RepoBench-R/P.
A completion output at the correct cursor that emits the next-line action is scored under \texttt{complete-line}.
Selecting a different context index is an artifact miss, even if the generated line is plausible.
\end{casebox}

\begin{casebox}[label=case:map-swebench]{SWE-bench Lite: \texttt{sqlfluff\_\_sqlfluff-1625}}
\textbf{Native benchmark row.}
\texttt{sqlfluff\_\_sqlfluff-1625}: the issue states that TSQL rule L031 incorrectly triggers ``Avoid using aliases in join condition'' when no join is present.

\medskip
\textbf{Native target source.}
The gold patch modifies:
\[
\texttt{src/sqlfluff/rules/L031.py}.
\]
The test patch includes:
\[
\texttt{test/cli/commands\_test.py},
\]
and \texttt{FAIL\_TO\_PASS} contains:
\[
\texttt{commands\_test.py::test\_\_cli\_\_command\_directed}.
\]

\medskip
\textbf{Projected scorer targets.}
For the patch-stage diagnostic:
\[
G_t^{\rm patch}
=
\{(\texttt{src/sqlfluff/rules/L031.py},\texttt{edit})\}.
\]
For the test-stage diagnostic:
\[
G_t^{\rm test}
=
\{(\texttt{commands\_test.py::test\_\_cli\_\_command\_directed},\texttt{run-test})\}.
\]

\medskip
\textbf{Scoring examples.}
An output \texttt{edit/src/sqlfluff/rules/L031.py} is a diagnostic joint hit for the patch-stage target.
Running the listed failing test is a diagnostic joint hit for the test-stage target.
Final task success is still decided by the native SWE-bench patch-and-test evaluator, not by our diagnostic projection alone.
\end{casebox}

\begin{casebox}[label=case:map-taubench]{$\tau$-bench: retail exchange task}
\textbf{Native benchmark row.}
Retail task: Yusuf Rossi in ZIP code \texttt{19122} wants to exchange items in order \texttt{\#W2378156}: a mechanical keyboard for a similar clicky one and a smart thermostat compatible with Google Home.

\medskip
\textbf{Native target source.}
The public task action trace includes:
\[
\texttt{find\_user\_id\_by\_name\_zip(first\_name=Yusuf,\ last\_name=Rossi,\ zip=19122)}
\]
followed by order/product lookups and a final exchange call:
\[
\texttt{exchange\_delivered\_order\_items(order\_id=\#W2378156,\ ...)}.
\]

\medskip
\textbf{Projected scorer targets.}
For the first lookup step:
\[
G_t^{\rm lookup}
=
\{(\texttt{find\_user\_id\_by\_name\_zip},\texttt{tool-call})\}.
\]
For the exchange step:
\[
G_t^{\rm exchange}
=
\{(\texttt{exchange\_delivered\_order\_items:\#W2378156},\texttt{tool-call})\}.
\]

\medskip
\textbf{Scoring examples.}
A call to \texttt{find\_user\_id\_by\_name\_zip} with the correct user-identifying intent is a joint hit for the lookup projection.
A call to \texttt{exchange\_delivered\_order\_items} on order \texttt{\#W2378156} is a joint hit for the exchange projection.
A tool call on an unrelated order or entity is an artifact miss.
\end{casebox}

\begin{casebox}[label=case:map-longcli]{LongCLI: \texttt{61810\_cow}}
\textbf{Native benchmark task.}
LongCLI task \texttt{61810\_cow}, a long-horizon CLI programming task derived from the copy-on-write fork lab.

\medskip
\textbf{Native target source.}
The task metadata points to a project workspace and instruction file for implementing copy-on-write fork behavior. LongCLI evaluates the agent through task-level and step-level requirement checks, including fail-to-pass and pass-to-pass test outcomes.

\medskip
\textbf{Projected scorer targets.}
For the implementation-stage diagnostic:
\[
G_t^{\rm impl}
=
\{(\texttt{61810\_cow/},\texttt{edit-feature})\}.
\]
For the test-stage diagnostic:
\[
G_t^{\rm test}
=
\{(\texttt{task-test-harness},\texttt{run-test})\}.
\]

\medskip
\textbf{Scoring examples.}
An action that edits the \texttt{61810\_cow} workspace toward the copy-on-write implementation is an artifact hit under the implementation-stage projection.
An action that runs the task's test harness is an action hit for the test-stage projection.
Whether the solution ultimately succeeds is determined by LongCLI's native requirements and regression tests.
\end{casebox}
\paragraph{Per-benchmark regime.}
Under the default Qwen protocol, BFCL and ToolSandbox are high-local-sufficiency settings where communication is fragile or harmful. ContextBench and RepoBench are context-heavy settings where the receiver's local view is incomplete and sender-side curation drives the signal. SWE-bench Lite probes capability transfer, and $\tau$-bench is a weakly helpful, protocol-sensitive customer-service setting. LongCLI-Bench drives the shared-state stress test in Appendix~\ref{app:shared-state}.

\paragraph{Replay frontier regret.}
On the frozen-message replay set, $\mathrm{Regret}^{\rm replay}_r=\max_{r'} h_{j,r'}-h_{j,r}$ measures how far a receiver's $h_j$ falls below the best replay receiver on the same frozen message.

\paragraph{Run bookkeeping.}
Table~\ref{tab:run-bookkeeping} records the support convention used by each
appendix numerical family. The OpenAI entries are aggregated from the stored
per-step API records and their summary CSVs; no scorer-side targets are exposed
to the models. The table distinguishes main-text supports from table-specific
appendix diagnostic supports.

\begin{table}[t]
\centering
\caption{\textbf{Run bookkeeping for appendix numerical tables.}
The table records which support convention each appendix family uses. Main-text
estimates use the supports stated in Table~\ref{tab:main-results}. Appendix
diagnostics use table-specific supports over stored model outputs, except for
the explicitly marked OpenAI receiver-side budget sweep.}
\label{tab:run-bookkeeping}
\footnotesize
\setlength{\tabcolsep}{3pt}
\renewcommand{\arraystretch}{1.08}
\begin{tabular}{@{}p{0.12\textwidth}p{0.145\textwidth}p{0.17\textwidth}p{0.13\textwidth}p{0.20\textwidth}p{0.105\textwidth}@{}}
\toprule
Run family & Benchmarks & Records / evaluation support & Sender(s) & Receiver(s) & Decoding and budget \\
\midrule
Released Qwen table cells
& ContextBench, $\tau$-bench, ToolSandbox, BFCL
& ContextBench $n=480$; $\tau$-bench $n=72$; ToolSandbox $n=72$; BFCL $n=100$
& Qwen3-Next-80B-A3B
& Qwen3.5-9B
& det.; $B=128$ \\
Appendix diagnostic controls
& ContextBench, ToolSandbox, $\tau$-bench, BFCL; RepoBench; SWE-Lite
& diagnostic supports: ContextBench $480$; ToolSandbox/$\tau$-bench $72$; BFCL $100$; RepoBench $480$; SWE-Lite $96$
& Qwen3-Next-80B-A3B; Qwen3-Coder-Next-FP8 for SWE-Lite MA
& Qwen3.5-9B
& det.; $B=128$ \\
OpenAI model-family extension
& ContextBench, ToolSandbox, $\tau$-bench, BFCL
& stored API run summaries, $n=60$--$300$ per displayed cell
& \texttt{gpt-5.4} in MA
& six OpenAI receivers listed in Tables~\ref{tab:openai-extension-counts}--\ref{tab:openai-extension-counts-continued}
& API defaults; $B=128$ \\
Selector diagnostic ablation
& ContextBench
& stored selector-ablation file, $n=300$
& \texttt{gpt-5.4}
& \texttt{gpt-4o-mini}
& frozen selector \\
Static selector diagnostics
& ToolSandbox, $\tau$-bench
& stored selector-policy files, $n=36$ per policy
& Qwen3-Next-80B-A3B
& Qwen3.5-9B
& static selector policies \\
Budget/no-overflow diagnostics
& ContextBench, ToolSandbox
& paired diagnostic rows $480/72$; no-overflow subsets $100/39$
& Qwen3-Next-80B-A3B
& Qwen3.5-9B
& det.; threshold $8192$ input tokens \\
OpenAI message-budget sweep
& ContextBench, ToolSandbox
& hard sample: $20$ steps $\times$ $3$ receivers $\times$ $6$ budgets per benchmark
& stored Qwen messages
& \texttt{gpt-5.4-nano}, \texttt{gpt-5.4-mini}, \texttt{gpt-5.4}
& receiver-side sweep \\
\bottomrule
\end{tabular}
\end{table}
\FloatBarrier

\paragraph{RepoBench and SWE-bench Lite diagnostic breakdowns.}
Table~\ref{tab:repobench-swe-breakdown} reports appendix diagnostic
breakdowns for the two supplemental benchmark families on the stated supports.
These diagnostics examine the same regime patterns as the main text.
The SWE-Lite main sender is Qwen3-Coder-Next-FP8. Its released sender-comparison
manifest selects the last record per $(\mathrm{instance},\mathrm{step})$ key:
the main-sender slice uses seed 2 and the matched slice uses seed 1.
The SWE-Lite contrast compares the recorded sender configurations; the matched
run's exported receiver inputs predominantly contain an empty coordination-message field.

\begin{table}[t]
\centering
\caption{\textbf{RepoBench and SWE-bench Lite appendix diagnostic breakdowns.}
Reported step-level rates on the stated supports. The released sender
contrasts use common \((\mathrm{instance},\mathrm{step})\) keys after
the manifest's duplicate policy is applied; the SWE-Lite seed mapping is
specified above.}
\label{tab:repobench-swe-breakdown}
\small
\begin{tabular}{lrrrrrrrr}
\toprule
Benchmark & $n$ & RO & SF & $\mathrm{MA}_{\rm main}$ & $\mathrm{MA}_{\rm match}$ & $\Delta_{\rm comm}$ & $\Delta_{\rm cmp}$ & $\Delta_{\rm strong}$ \\
\midrule
RepoBench
& 480 & 0.642 & 0.823 & 0.833 & 0.746 & +0.192 & +0.010 & +0.088 \\
SWE-bench Lite
& 96 & 0.000 & 0.052 & 0.521 & 0.000 & +0.521 & +0.469 & +0.521 \\
\bottomrule
\end{tabular}
\end{table}
\paragraph{Paired uncertainty.}
The communication effects remained positive on ContextBench and RepoBench,
negative on BFCL and ToolSandbox, and near zero on $\tau$-bench.
Table~\ref{tab:paired-clustered} reports paired differences on strict common
$(\mathrm{instance},\mathrm{step})$ keys. We computed instance-clustered
percentile bootstrap intervals with fixed seeds and 5,000 resamples, and
two-sided exact McNemar tests on discordant pairs. ContextBench, RepoBench,
SWE-Lite, and BFCL remained significant after Bonferroni correction for six contrasts;
$\tau$-bench had $p=1.0$. ContextBench and RepoBench use larger paired
diagnostic supports than the earlier appendix audits, separate from the
main-table evaluation sets.

\begin{table}[!htbp]
\centering
\caption{\textbf{Paired communication effects with uncertainty.}
Intervals are clustered by instance. Support is the number of paired steps;
ContextBench and RepoBench contain 799 and 800 instances, respectively.
The SWE-Lite contrast pairs instance-step keys across main-sender seed 2
and matched-sender seed 1.}
\label{tab:paired-clustered}
\small
\setlength{\tabcolsep}{4pt}
\begin{tabular}{@{}llrrl@{}}
\toprule
Benchmark & Contrast & Steps & Effect & 95\% CI \\
\midrule
ContextBench & MA $-$ RO & 2,390 & $+0.1711$ & $[0.1417,0.2012]$ \\
RepoBench & MA $-$ RO & 2,396 & $+0.1611$ & $[0.1340,0.1891]$ \\
SWE-Lite & MA$_{\rm main}-$MA$_{\rm match}$ & 96 & $+0.5208$ & $[0.3854,0.6563]$ \\
$\tau$-bench & MA $-$ RO & 72 & $0.0000$ & $[-0.0833,+0.0694]$ \\
BFCL & MA $-$ RO & 100 & $-0.2300$ & $[-0.3900,-0.0800]$ \\
\bottomrule
\end{tabular}
\end{table}

The discordant wins were 499 versus 90 on ContextBench, 50 versus 0 on
SWE-Lite, and 1 versus 24 on BFCL; matched-sender accuracy on the SWE-Lite
support was zero. The clustered intervals on BFCL and $\tau$-bench are wider
than their step-level intervals, $[-0.3200,-0.1500]$ and
$[-0.0694,+0.0694]$, while retaining the same conclusions.
On ToolSandbox, the released common support contains 72 steps, with 64
correct RO outputs and 47 correct MA outputs. This gives a descriptive
RO advantage of $17/72$, or $23.6$ percentage points.
\Needspace{11\baselineskip}
\paragraph{SWE-Lite action validity.}
The zero receiver-only joint score reflected the selected action despite
accurate file recovery. All outputs on the audited 96-step slices from each
of three seeds were parse-valid and had nonempty action and artifact fields;
artifact accuracy ranged from 0.885 to 0.917. Released seeds 0 and 2 each
contained 79 \texttt{read} and 17 \texttt{run\_test} outputs on 96 steps,
with zero joint hits where the repair-stage target was \texttt{edit}.
Seed 1 contained two \texttt{edit} outputs and had 2/96 joint hits.
These are the next-action scores defined in
Table~\ref{tab:target-construction}; final SWE-bench correctness uses patch
and test execution.

\subsection{Extended Compression Results}
\label{app:claim1-extra}

\begin{table}[t]
\centering
\caption{\textbf{Receiver-scaling and SF controls on ToolSandbox.} MA uses 80B or 397B receivers; SF uses an 80B receiver. These configurations complement the 9B receiver results in Table~\ref{tab:main-results}.}
\label{tab:appendix-robustness}
\small
\begin{tabular}{lccc}
\toprule
Setting & Benchmark & Mean & Note \\
\midrule
MA 80B$\to$80B & ToolSandbox & 0.750 & stronger receiver baseline \\
MA 80B$\to$397B & ToolSandbox & 0.694 & stronger receiver baseline \\
SF (80B receiver) & ToolSandbox & 0.260 & full-context baseline (80B receiver) \\
\bottomrule
\end{tabular}
\end{table}

\begin{figure}[t]
\centering
\safeincludegraphics[width=0.88\columnwidth]{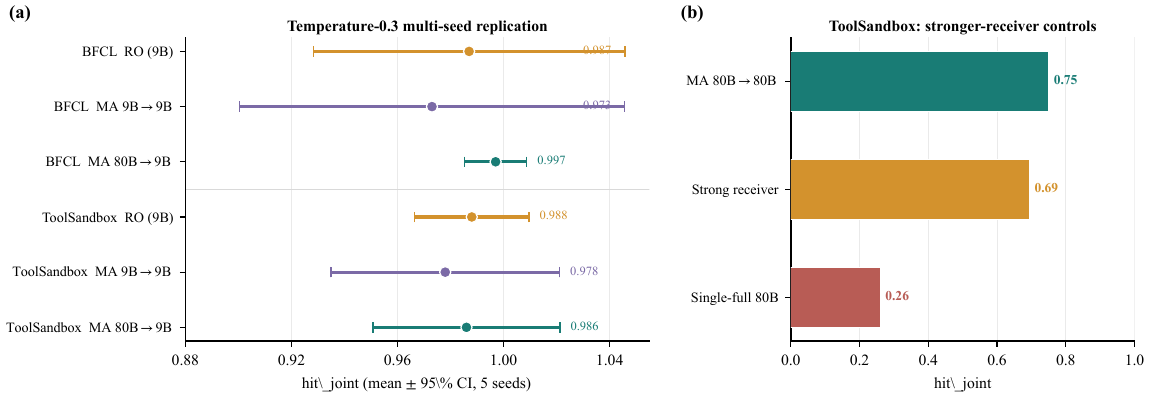}
\caption{\textbf{Additional compression-effect controls.} (a) Five-seed replication at temperature $0.3$ on BFCL and ToolSandbox. (b) Stronger-receiver and SF controls on ToolSandbox.}
\label{fig:robustness}
\end{figure}

Table~\ref{tab:appendix-robustness} and Figure~\ref{fig:robustness} examine larger receivers and raw-context access on ToolSandbox. Substantial error remains in these MA and SF configurations, including with 80B and 397B receivers. Together with the main 9B results, these controls motivate evaluating communication separately for each receiver and interface.

\paragraph{Budget accounting and no-overflow controls.}
Table~\ref{tab:budget-accounting-no-overflow} reports realized protocol-level
token accounting from stored diagnostic step records. The columns separate the
bounded sender message from the total input tokens consumed by each protocol
row, so the RO, SF, and MA rows are comparable under the same tokenizer
accounting. The final column is a no-overflow control: we pair rows by
\((\mathrm{instance},t)\), keep only examples for which all compared RO/SF/MA
rows stay below 8192 total input tokens, and recompute \texttt{hit\_joint}.
This is conservative for MA because the protocol-level total includes both
sender-side and receiver-side calls. The same sign pattern remains on this
short-context diagnostic subset. ContextBench still has MA above both RO and
SF, while ToolSandbox still has MA below RO/SF. Thus the reported communication
gains and failures are not explained only by SF context-window overload.

\begin{table}[!htbp]
\centering
\caption{\textbf{Token accounting and no-overflow controls.}
Token counts are realized counts from the stored diagnostic records. The
no-overflow column filters to paired rows where RO, SF, and MA all use at most
8192 protocol-level input tokens, giving $n=100$ ContextBench rows and $n=39$
ToolSandbox rows. These diagnostics use the stated appendix support and are not
used as replacements for the main-text estimates.}
\label{tab:budget-accounting-no-overflow}
\footnotesize
\setlength{\tabcolsep}{4pt}
\renewcommand{\arraystretch}{1.08}
\resizebox{\textwidth}{!}{%
\begin{tabular}{llrrrrrr}
\toprule
Benchmark & Protocol & $n$ & \texttt{hit\_joint} & Message tok. & Total protocol input tok. & Trunc. & No-overflow \\
& & & & mean / p90 & mean / p90 & rate & \texttt{hit\_joint} \\
\midrule
ContextBench & RO & 480 & 0.633 & 0.0 / 0.0 & 6,899 / 12,267 & 0.000 & 0.610 \\
ContextBench & SF & 480 & 0.294 & 0.0 / 0.0 & 24,486 / 51,642 & 0.000 & 0.280 \\
ContextBench & MA, $B{=}128$ & 480 & 0.775 & 127.2 / 128.0 & 16,365 / 28,160 & 0.842 & 0.790 \\
\midrule
ToolSandbox & RO & 72 & 0.889 & 0.0 / 0.0 & 1,050 / 1,672 & 0.000 & 0.923 \\
ToolSandbox & SF & 72 & 0.903 & 0.0 / 0.0 & 1,273 / 2,054 & 0.000 & 0.923 \\
ToolSandbox & MA, $B{=}128$ & 72 & 0.653 & 115.2 / 128.0 & 2,534 / 4,000 & 0.208 & 0.590 \\
\bottomrule
\end{tabular}
}
\end{table}
\paragraph{Message truncation.}
Communication improved ContextBench accuracy within both capped and uncapped
message groups. We split the 480-step, $B=128$ audit in
Table~\ref{tab:budget-accounting-no-overflow} by whether the sender message
was truncated, keeping the protocol fixed. Table~\ref{tab:truncation-groups}
shows that MA exceeded both RO and SF in each group. The groups pool to the
reported RO/SF/MA accuracies of 0.633/0.294/0.775.

\begin{table}[!htbp]
\centering
\caption{\textbf{ContextBench accuracy by message truncation.}
Both groups use the same $B=128$ protocol. The final two columns report paired
accuracy differences in percentage points.}
\label{tab:truncation-groups}
\small
\setlength{\tabcolsep}{4pt}
\begin{tabular}{@{}lrrrrrr@{}}
\toprule
Messages & Steps & RO & SF & MA & MA$-$RO & MA$-$SF \\
\midrule
Capped & 404 & 0.644 & 0.248 & 0.750 & $+10.64$ & $+50.25$ \\
Uncapped & 76 & 0.579 & 0.539 & 0.908 & $+32.89$ & $+36.84$ \\
\bottomrule
\end{tabular}
\end{table}

The paired 95\% intervals for MA$-$RO were $[+5.7,+15.8]$ pp in the capped
group and $[+22.4,+43.4]$ pp in the uncapped group. All 404 capped messages
retained \texttt{action\_type} and \texttt{target\_artifact}, and the receiver
followed both fields in 402 cases. The cap therefore removed trailing content
while preserving the decision fields. In a separate paired Qwen run on 72
ContextBench steps, doubling the message budget reduced truncation from
0.8889 to 0.2778 and increased accuracy from 0.8194 to 0.8333.
\Needspace{9\baselineskip}
\paragraph{Message form and receiver instruction.}
On a common ToolSandbox slice, canonical JSON and a decision-header template
both reached 0.750 accuracy, compared with 0.556 for prose, a 19.4 pp
difference. These comparisons varied the user-turn content while keeping
\texttt{RECEIVER\_SYSTEM\_PROMPT} fixed. The released sequential agent and
four-action threshold controller share this receiver instruction. The
five-action scorer described in Appendix~\ref{app:tax-controller}
uses user-message probes.

\paragraph{OpenAI receiver-side message-budget sweep.}
We reran the receiver-side budget sweep with stored Qwen sender messages, varying only the visible coordination-message budget. The hard sample prioritized stored sender failures, already-truncated messages, and longer protocol inputs. The sweep covered three receiver aliases, \texttt{gpt-5.4-nano},
\texttt{gpt-5.4-mini}, and \texttt{gpt-5.4}, on 20 sampled steps per benchmark and six budgets. Table~\ref{tab:openai-budget-sweep-hard} shows distinct budget responses. ContextBench improves from \(B=4\) to the \(B\ge 16\) range, while ToolSandbox degrades as more of the sender message is exposed. The effect of message budget therefore depends on the receiver's task: additional content can supply useful context or add protocol tax.

\begin{table}[!htbp]
\centering
\caption{\textbf{OpenAI receiver-side hard-sample budget sweep.}
Rows report \texttt{hit\_joint} on 20 hard-sampled steps per benchmark and
receiver alias, using stored Qwen sender messages and varying only the visible
message budget. The hard sample prioritizes stored sender failures,
sender-message truncation, and longer protocol inputs. The run completed 720
API calls with no failed calls.}
\label{tab:openai-budget-sweep-hard}
\small
\begin{tabular}{lrrrrr}
\toprule
Benchmark & $B$ & \texttt{nano} & \texttt{mini} & \texttt{gpt-5.4} & Mean \\
\midrule
ContextBench & 4 & 0.700 & 0.600 & 0.400 & 0.567 \\
ContextBench & 8 & 0.650 & 0.650 & 0.600 & 0.633 \\
ContextBench & 16 & 0.700 & 0.650 & 0.750 & 0.700 \\
ContextBench & 32 & 0.700 & 0.700 & 0.700 & 0.700 \\
ContextBench & 64 & 0.650 & 0.700 & 0.700 & 0.683 \\
ContextBench & 128 & 0.700 & 0.700 & 0.700 & 0.700 \\
\midrule
ToolSandbox & 4 & 0.900 & 0.950 & 0.800 & 0.883 \\
ToolSandbox & 8 & 0.550 & 0.700 & 0.750 & 0.667 \\
ToolSandbox & 16 & 0.350 & 0.600 & 0.800 & 0.583 \\
ToolSandbox & 32 & 0.350 & 0.400 & 0.450 & 0.400 \\
ToolSandbox & 64 & 0.250 & 0.300 & 0.250 & 0.267 \\
ToolSandbox & 128 & 0.250 & 0.250 & 0.250 & 0.250 \\
\bottomrule
\end{tabular}
\end{table}

\paragraph{Strict recoding probe.}
The strict-recoding probe holds the scorer-side artifact/action target fixed
while changing only the receiver-facing surface form of the sender
recommendation. The direct condition exposes canonical field names and action
words; the strict condition replaces them with semantically indirect labels
that preserve the benchmark target for an unbounded decoder but stress the
deployed receiver interface. Table~\ref{tab:strict-recoding-consistency}
reports the stored direct-versus-strict diagnostic comparison. BFCL is
unaffected because its function-call schema is already highly canonicalized,
while ContextBench, ToolSandbox, and especially \(\tau\)-bench lose joint
accuracy under the recoded surface. This is the empirical counterpart of
Theorem~\ref{thm:sufficiency-not-enough}: preserving the Bayes target is not
enough when the receiver class is representation-sensitive.

\begin{table}[!htbp]
\centering
\caption{\textbf{Direct versus strict recoded sender messages.}
The evaluated target \(G_t\) is unchanged across conditions; only the
receiver-facing representation changes. Values are recomputed from the stored
recoding-probe records on the stated diagnostic supports.}
\label{tab:strict-recoding-consistency}
\small
\begin{tabular}{lrrrr}
\toprule
Benchmark & $n$ & Direct & Strict recode & Drop \\
\midrule
BFCL & 200 & 0.985 & 0.985 & 0.0 pp \\
ContextBench & 100 & 1.000 & 0.940 & 6.0 pp \\
ToolSandbox & 200 & 1.000 & 0.965 & 3.5 pp \\
\(\tau\)-bench & 155 & 1.000 & 0.865 & 13.5 pp \\
\bottomrule
\end{tabular}
\end{table}

\begin{figure}[!htbp]
\centering
\safeincludegraphics[width=0.95\columnwidth]{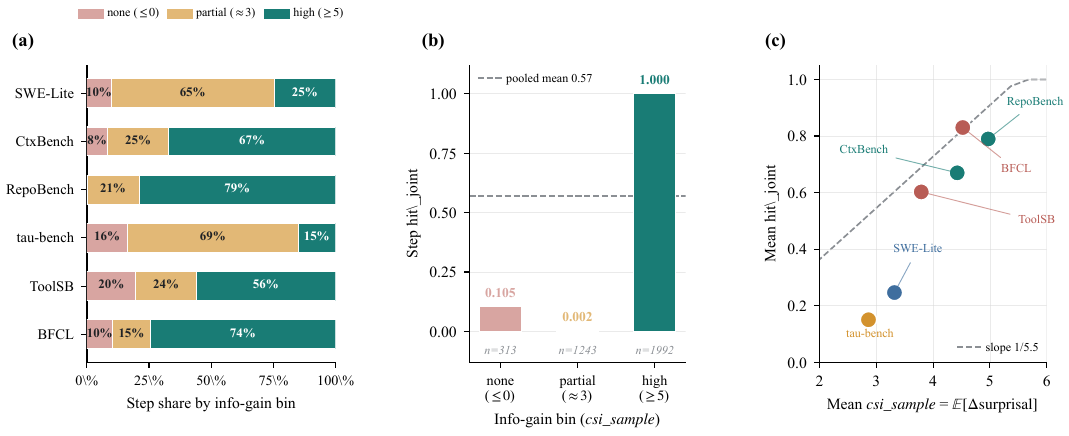}
\caption{\textbf{Gain-surrogate view of message utility.}
(a) Per-benchmark composition of the step-level hit-based proxy gain surrogate $\texttt{csi\_sample}$, split into none, partial, and high bins.
(b) Step-level \texttt{hit\_joint} conditional on gain-surrogate bin.
(c) Benchmark-level mean gain surrogate versus mean \texttt{hit\_joint}. High-surrogate steps occur in both helpful and harmful communication regimes.}
\label{fig:infogain}
\end{figure}

\paragraph{Gain-surrogate diagnostic.}
The main text uses
\[
\hat A:=\texttt{surprisal\_no\_msg}-\texttt{surprisal\_bits}
\]
as a surrogate for receiver-relative gain. Both terms are hit-based proxy scores, not measured log-probabilities. The scorer sets \texttt{surprisal\_bits} to $0.5$, $3$, or $6$ bits when the receiver's output is a joint hit, a partial hit (action or artifact only), or a miss, and sets \texttt{surprisal\_no\_msg} in the same way from a receiver call without the message. $\hat A$ therefore records how the hit outcome changes when the message is exposed, on a coarse surprisal scale, rather than a measured log-loss reduction; under an idealized zero-slack calibrated log-loss specialization, the expected log-loss reduction would correspond to conditional mutual information. Figure~\ref{fig:infogain} compares the distribution of $\hat A$, conditional joint accuracy, and benchmark averages. The high-$\hat A$ bucket contains 74\% of BFCL steps, 56\% of ToolSandbox steps, 67\% of ContextBench steps, and 25\% of SWE-Lite steps. High $\hat A$ therefore occurs in both helpful and harmful communication regimes. The surrogate describes the gain component; assessing deployed utility also requires the protocol tax.

These diagnostics complement the format comparisons in Table~\ref{tab:ablations}. The selector in Section~\ref{subsec:selector} compares each bounded-message action with \texttt{raw} using both gain and tax.

\FloatBarrier

\subsection{Extended Closure Results}
\label{app:claim2-extra}

\begin{figure}[!htbp]
\centering
\safeincludegraphics[width=0.95\columnwidth]{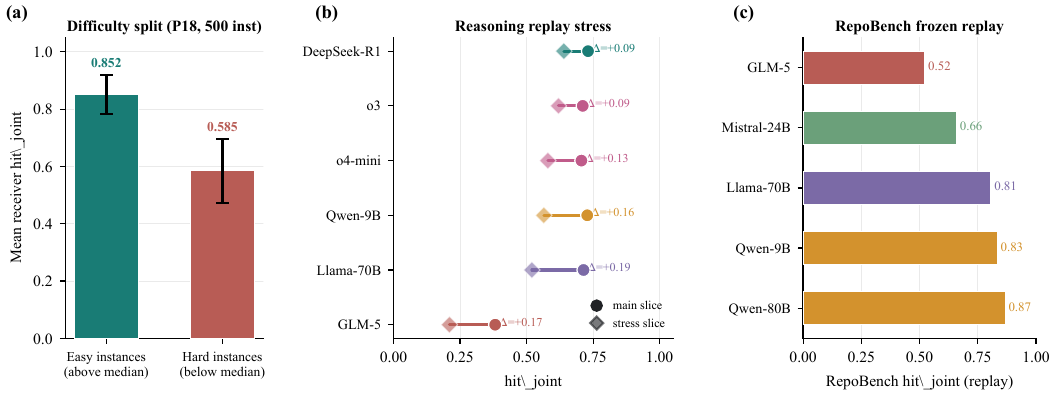}
\caption{\textbf{Additional replay and closure analyses.}
Panels report easy/hard replay partitions, replay stress across receivers, and RepoBench replay accuracy under frozen messages. The same decomposition remains evident: some receivers are primarily content-limited, whereas the main outliers are primarily closure-limited.}
\label{fig:absorption-extra}
\end{figure}

The main text uses fixed-message replay and the OpenAI full-context rows to examine action-closure loss. Figure~\ref{fig:absorption-extra} extends this in three ways. First, it shows that the content/closure split remains stable under harder/easier partitions of the same replay set. Second, it adds replay stress on additional receivers, confirming that the main outliers remain outliers under a frozen message. Third, it applies the same logic to RepoBench, where the absolute accuracy scale changes but the interpretation of the bottleneck axes remains the same.

With the sender message fixed, these additional receivers and task partitions show that post-message error can concentrate in action closure, as described by Proposition~\ref{prop:replay-isolation} and Theorem~\ref{thm:bottleneck-chain}.
\paragraph{Parsing and action choice.}
The closure failures persisted after valid outputs recovered the correct
artifact. We re-parsed a separate replay set with a \texttt{gpt-5.4} sender
using robust and permissive parsers. Table~\ref{tab:replay-parse-audit} shows
that permissive parsing changed validity by at most 0.2 pp, while wrong-action
rates remained 40.8--94.0\% among valid outputs with the correct artifact.
GLM-5 and MiniMax-M2.5 named \path{astropy/modeling/separable.py}, and
\texttt{o3} named \path{astropy/table/table.py}, while explicitly selecting
\texttt{read} where the evaluated action was \texttt{edit}.

\begin{table}[!htbp]
\centering
\caption{\textbf{Parsing audit under a \texttt{gpt-5.4} sender replay.}
Validity columns use all audited outputs. Wrong-action rates condition on
valid parsing and a correct artifact. This is a separate sender condition
from Table~\ref{tab:replay-summary}. All entries are percentages.}
\label{tab:replay-parse-audit}
\small
\setlength{\tabcolsep}{5pt}
\begin{tabular}{@{}lrrr@{}}
\toprule
Receiver & Robust valid & Permissive valid & Wrong action \\
\midrule
GLM-5 & 99.6 & 99.6 & 44.5 \\
MiniMax-M2.5 & 86.8 & 87.0 & 94.0 \\
\texttt{o3} & 97.3 & 97.3 & 40.8 \\
\bottomrule
\end{tabular}
\end{table}

An audit of all 300 paired \texttt{o3} RO/SF steps reproduced accuracies
0.223/0.133. Robust parse validity was 99.33\% for RO and 89.00\% for SF.
The discordant counts were 52 RO wins versus 25 SF wins; 41 of the 52 SF
losses retained the correct artifact and selected the wrong action.

\FloatBarrier

\subsection{Extended Stronger-Sender Results}
\label{app:claim3-extra}

\begin{figure}[!htbp]
\centering
\safeincludegraphics[width=0.95\columnwidth]{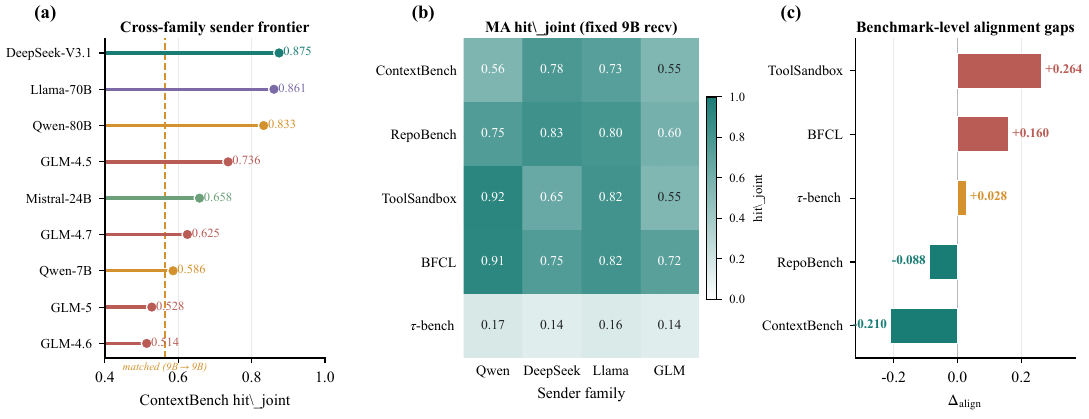}
\caption{\textbf{Additional sender-threshold analyses.}
Panels show a ContextBench sender frontier, a sender-family accuracy matrix, and benchmark-level effects with a fixed receiver. Panel (c) uses $\Delta_{\rm align}=-\Delta_{\rm strong}$, so positive values favor the matched sender.}
\label{fig:alignment-extra}
\end{figure}

Figure~\ref{fig:alignment-extra} extends the main-text sender-gap story in two directions. First, it shows that the strong-sender effect is not a same-family phenomenon: across additional sender families, the sign and rough ranking remain task-dependent. Second, it makes clear that the threshold is not only a benchmark-level average, but also visible in sender-family matrices and pilot-set frontier plots.

\begin{figure}[!htbp]
\centering
\safeincludegraphics[width=0.95\columnwidth]{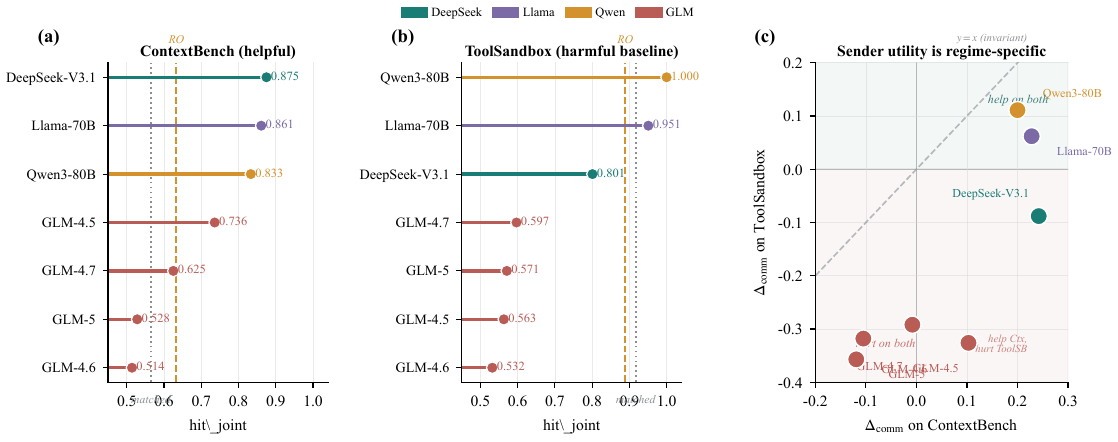}
\caption{\textbf{Sender-replacement ablation (fixed Qwen3.5-9B receiver).}
(a) ContextBench sender ranking against RO and matched-sender references.
(b) The same sender set on ToolSandbox.
(c) Per-sender signed communication effect on ContextBench versus ToolSandbox.
A sender that helps in one setting can hurt in another, so sender choice is a task-level protocol decision rather than a fixed ranking of models.}
\label{fig:sender-swap}
\end{figure}

\paragraph{Sender-replacement ablation.}
The sender-replacement study in Figure~\ref{fig:sender-swap} fixes the Qwen3.5-9B receiver and swaps the sender across seven candidates spanning multiple families and scales. This is the most direct test of the sender-gap theorem under a fixed receiver. On ContextBench, the non-GLM strong senders remain above the RO baseline, while the weakest GLM senders fall below it. On ToolSandbox, where local sufficiency is already high, even some strong senders that help on ContextBench fall below the RO threshold. The point is not that one sender family is universally better, but that sender helpfulness is regime-specific because both the extra receiver-relative gain and the induced tax vary across task settings.

This ablation complements Figure~\ref{fig:alignment}. Figure~\ref{fig:alignment} fixes the sender pair and varies the benchmark regime. Figure~\ref{fig:sender-swap} fixes the receiver and benchmark condition and varies the sender identity. Together, they show that the threshold binds to the sender--receiver--task combination, not only to a coarse family label.

\paragraph{Statistical sender-threshold fit.}
The sender-upgrade theorem predicts a sign change in the value of a stronger sender as receiver-local sufficiency increases: when RO is low, the upstream sender has room to add receiver-relative gain; when RO is already high, additional sender complexity is more likely to appear as protocol tax. We fit this relationship on the released benchmark-cell anchors, using \(\Delta_{\rm strong}=\mathrm{MA}_{\rm strong}-\mathrm{MA}_{\rm matched}\) as the response and RO accuracy as the local-sufficiency proxy. Across 22 anchors, consisting of 16 OpenAI-family cells and 6 Qwen cells, the linear fit is
\[
\widehat\Delta_{\rm strong}=0.286-0.449\,\mathrm{RO},
\qquad R^2=0.492.
\]
The fitted zero-crossing is \(\mathrm{RO}=0.638\), with bootstrap 95\% interval \([0.491,0.780]\). A segmented hinge fit gives a similar \(R^2\) but a wider interval because the released anchors contain few very-low-RO cells. We therefore use the linear fit as the appendix summary statistic and keep the theorem's condition as a sufficient-condition interpretation rather than a universal scaling law.

\begin{table}[!htbp]
\centering
\caption{\textbf{Sender-upgrade threshold fit.}
The response is \(\Delta_{\rm strong}=\mathrm{MA}_{\rm strong}-\mathrm{MA}_{\rm matched}\); RO accuracy is used as the observable local-sufficiency proxy. The negative fitted slope supports the theorem's regime prediction: stronger senders help where local sufficiency is low and become tax-dominated where RO is high.}
\label{tab:sender-threshold-fit}
\small
\begin{tabular}{lrrrrr}
\toprule
Fit & Anchors & Slope on RO & $R^2$ & Crossover RO & 95\% CI \\
\midrule
Linear & 22 & -0.449 & 0.492 & 0.638 & [0.491, 0.780] \\
Segmented / hinge & 22 & piecewise & 0.503 & 0.300 & [0.300, 0.900] \\
\bottomrule
\end{tabular}
\end{table}
\paragraph{Expanded sender-swap support.}
Adding the stronger-sender condition to every receiver/benchmark cell that
lacked one expanded the paired anchor set from 22 to 41. The fitted crossover
remained close to the original estimate, while its bootstrap interval
narrowed from $[0.491,0.780]$ to $[0.552,0.706]$
(Table~\ref{tab:expanded-sender-threshold}). Among 38 anchors with nonzero
effects, 34 (89.5\%) agreed with one positive-to-negative crossing as RO
accuracy increased. A hierarchical model with anchor random effects over the
12,540 paired steps preferred this structure to the linear/no-threshold and
two-crossing alternatives on all 10 folds separated by anchor.

\begin{table}[!htbp]
\centering
\caption{\textbf{Expanded sender-upgrade crossover analysis.}
The response is $\Delta_{\rm strong}$ and the predictor is receiver-only
accuracy, as in Table~\ref{tab:sender-threshold-fit}.}
\label{tab:expanded-sender-threshold}
\small
\setlength{\tabcolsep}{5pt}
\begin{tabular}{@{}lrrl@{}}
\toprule
Analysis & Anchor set & Crossover & 95\% CI \\
\midrule
Submitted fit & 22 & 0.6376 & $[0.491,0.780]$ \\
Expanded fit & 41 & 0.6289 & $[0.552,0.706]$ \\
Leave one benchmark family out & 41 & 0.6451 & \\
Excluding SWE-Lite & 41 & 0.6154 & \\
\bottomrule
\end{tabular}
\end{table}

\paragraph{Within-family identity check.}
Table~\ref{tab:within-tier-identity} records the OpenAI same-family self-pair controls available in the stored run. These controls separate the effect of changing the sender identity from the effect of adding a same-family message channel.

\begin{table}[!htbp]
\centering
\caption{\textbf{Within-family sender identity check.}
Rows use the same-family MA self-pairs from the stored OpenAI run. Deltas compare each self-pair against the receiver's RO value for the same benchmark condition.}
\label{tab:within-tier-identity}
\small
\begin{tabular}{lrrrr}
\toprule
Sender $\to$ receiver & ContextBench acc. & $\Delta$ vs RO & ToolSandbox acc. & $\Delta$ vs RO \\
\midrule
\texttt{gpt-4o-mini}$\to$\texttt{gpt-4o-mini} & 0.867 & +0.217 & 0.960 & +0.157 \\
\texttt{gpt-5.4-nano}$\to$\texttt{gpt-5.4-nano} & 0.873 & +0.380 & 0.960 & +0.203 \\
\texttt{gpt-5.4-mini}$\to$\texttt{gpt-5.4-mini} & 0.867 & +0.253 & 0.913 & +0.217 \\
\bottomrule
\end{tabular}
\end{table}

\FloatBarrier

\subsection{Cross-Family Model Ablations}
\label{app:cross-family}

These model-family sweeps compare compression, action closure, and sender upgrades beyond the default Qwen protocol.

\paragraph{OpenAI model-family extension.}
On four of the six benchmarks, we evaluated the API model aliases
\texttt{gpt-4o-mini}, \texttt{gpt-5.4-nano}, \texttt{gpt-5.4-mini},
\texttt{gpt-5.4}, \texttt{o3}, and \texttt{o4-mini} in SF, RO, and the
\texttt{gpt-5.4}$\to$receiver MA configuration.
The model names identify the recorded API aliases; Llama-70B denotes \path{meta-llama/Llama-3.3-70B-Instruct-Turbo}. The OpenAI extension preserves the ContextBench communication gains and reveals a different ToolSandbox response: MA improves over RO for these receivers, whereas it reduces accuracy for the default Qwen receiver. Tables~\ref{tab:openai-extension-counts}
and~\ref{tab:openai-extension-counts-continued} report the stored step counts and Wilson intervals for the stated diagnostic supports.

\begin{table}[!htbp]
\centering
\caption{\textbf{OpenAI model-family extension counts and Wilson intervals, ContextBench and ToolSandbox.}
Each entry is \texttt{hit\_joint} with Wilson $95\%$ interval over the stored
step records for that diagnostic cell. MA uses \texttt{gpt-5.4} as the sender;
the stated counts are table-specific appendix supports.}
\label{tab:openai-extension-counts}
\footnotesize
\setlength{\tabcolsep}{2.4pt}
\renewcommand{\arraystretch}{1.08}
\resizebox{\textwidth}{!}{%
\begin{tabular}{@{}llclll@{}}
\toprule
Benchmark & Receiver & $n_{\rm RO}/n_{\rm SF}/n_{\rm MA}$ & RO $h_j$ & SF $h_j$ & MA $h_j$ \\
\midrule
ContextBench & \texttt{gpt-4o-mini} & 300/300/300 & 0.650 [0.594,0.702] & 0.743 [0.691,0.789] & 0.787 [0.737,0.829] \\
ContextBench & \texttt{gpt-5.4-nano} & 300/300/300 & 0.493 [0.437,0.550] & 0.637 [0.581,0.689] & 0.783 [0.733,0.826] \\
ContextBench & \texttt{gpt-5.4-mini} & 300/300/300 & 0.613 [0.557,0.667] & 0.770 [0.719,0.814] & 0.787 [0.737,0.829] \\
ContextBench & \texttt{gpt-5.4} & 300/300/150 & 0.600 [0.544,0.654] & 0.777 [0.726,0.820] & 0.853 [0.788,0.901] \\
ContextBench & \texttt{o3} & 300/300/150 & 0.223 [0.180,0.274] & 0.133 [0.099,0.176] & 0.493 [0.414,0.573] \\
ContextBench & \texttt{o4-mini} & 300/300/300 & 0.500 [0.444,0.556] & 0.617 [0.561,0.670] & 0.743 [0.691,0.790] \\
\midrule
ToolSandbox & \texttt{gpt-4o-mini} & 300/300/300 & 0.803 [0.755,0.844] & 0.957 [0.927,0.975] & 1.000 [0.987,1.000] \\
ToolSandbox & \texttt{gpt-5.4-nano} & 300/300/300 & 0.757 [0.705,0.802] & 0.927 [0.891,0.951] & 1.000 [0.987,1.000] \\
ToolSandbox & \texttt{gpt-5.4-mini} & 300/300/300 & 0.697 [0.642,0.746] & 0.890 [0.850,0.921] & 1.000 [0.987,1.000] \\
ToolSandbox & \texttt{gpt-5.4} & 300/300/150 & 0.653 [0.598,0.705] & 1.000 [0.987,1.000] & 0.993 [0.963,0.999] \\
ToolSandbox & \texttt{o3} & 150/150/150 & 0.740 [0.664,0.804] & 0.980 [0.943,0.993] & 1.000 [0.975,1.000] \\
ToolSandbox & \texttt{o4-mini} & 300/300/150 & 0.677 [0.622,0.727] & 0.733 [0.681,0.780] & 1.000 [0.975,1.000] \\
\bottomrule
\end{tabular}
}
\end{table}

\begin{table}[!htbp]
\centering
\caption{\textbf{OpenAI model-family extension counts and Wilson intervals, $\tau$-bench and BFCL.}
Each entry is \texttt{hit\_joint} with Wilson $95\%$ interval over the stored
step records for that diagnostic cell. MA uses \texttt{gpt-5.4} as the sender;
the stated counts are table-specific appendix supports.}
\label{tab:openai-extension-counts-continued}
\footnotesize
\setlength{\tabcolsep}{2.4pt}
\renewcommand{\arraystretch}{1.08}
\resizebox{\textwidth}{!}{%
\begin{tabular}{@{}llclll@{}}
\toprule
Benchmark & Receiver & $n_{\rm RO}/n_{\rm SF}/n_{\rm MA}$ & RO $h_j$ & SF $h_j$ & MA $h_j$ \\
\midrule
$\tau$-bench & \texttt{gpt-4o-mini} & 150/150/150 & 0.253 [0.190,0.328] & 0.020 [0.007,0.057] & 0.373 [0.300,0.453] \\
$\tau$-bench & \texttt{gpt-5.4-nano} & 150/150/150 & 0.313 [0.245,0.391] & 0.233 [0.173,0.307] & 0.300 [0.232,0.378] \\
$\tau$-bench & \texttt{gpt-5.4-mini} & 150/150/150 & 0.293 [0.226,0.371] & 0.233 [0.173,0.307] & 0.327 [0.257,0.405] \\
$\tau$-bench & \texttt{gpt-5.4} & 150/150/75 & 0.240 [0.179,0.314] & 0.180 [0.127,0.249] & 0.147 [0.084,0.244] \\
$\tau$-bench & \texttt{o3} & 75/75/75 & 0.200 [0.125,0.304] & 0.173 [0.104,0.274] & 0.160 [0.094,0.259] \\
$\tau$-bench & \texttt{o4-mini} & 150/150/75 & 0.273 [0.208,0.350] & 0.280 [0.214,0.357] & 0.160 [0.094,0.259] \\
\midrule
BFCL & \texttt{gpt-4o-mini} & 180/180/180 & 0.878 [0.822,0.918] & 0.722 [0.653,0.782] & 0.833 [0.772,0.881] \\
BFCL & \texttt{gpt-5.4-nano} & 180/180/180 & 0.883 [0.828,0.922] & 0.811 [0.748,0.862] & 0.839 [0.778,0.885] \\
BFCL & \texttt{gpt-5.4-mini} & 180/180/180 & 0.839 [0.778,0.885] & 0.833 [0.772,0.881] & 0.844 [0.784,0.890] \\
BFCL & \texttt{gpt-5.4} & 180/180/60 & 0.861 [0.803,0.904] & 0.833 [0.772,0.881] & 1.000 [0.940,1.000] \\
BFCL & \texttt{o3} & 60/60/60 & 1.000 [0.940,1.000] & 0.983 [0.911,0.997] & 1.000 [0.940,1.000] \\
BFCL & \texttt{o4-mini} & 180/180/60 & 0.844 [0.784,0.890] & 0.889 [0.835,0.927] & 1.000 [0.940,1.000] \\
\bottomrule
\end{tabular}
}
\end{table}

\begin{figure}[!htbp]
\centering
\safeincludegraphics[width=0.98\textwidth]{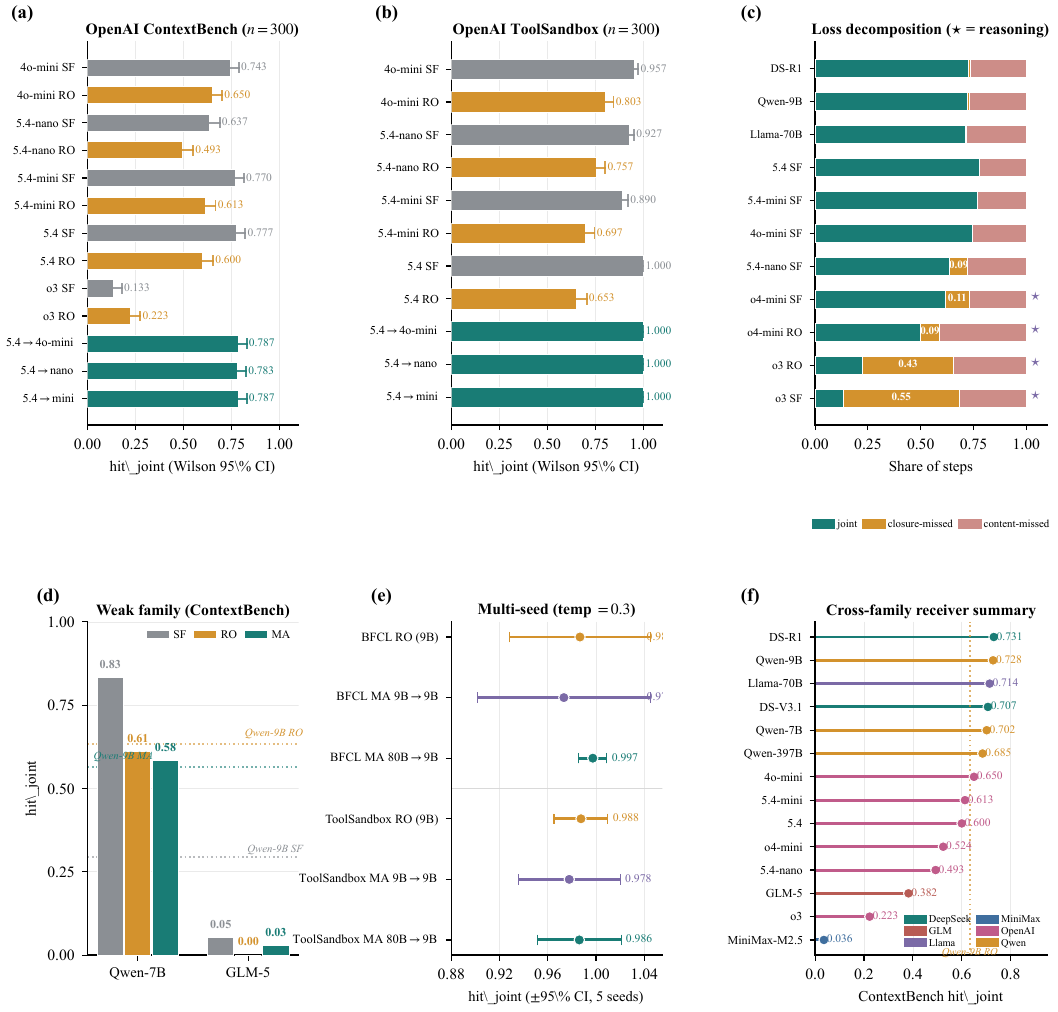}
\caption{\textbf{Cross-family ablations.}
(a) OpenAI model extensions on ContextBench.
(b) OpenAI ToolSandbox.
(c) ContextBench content/closure diagnostics across OpenAI and replay receivers.
(d) Weak-family full-protocol baselines.
(e) Five-seed replication at temperature $0.3$ on BFCL and ToolSandbox.
(f) Cross-family receiver summary on ContextBench.
Each panel extends one of the three main claims under a different family, receiver scale, or protocol stress.}
\label{fig:full-ablation}
\end{figure}

Panels~(a) and (b) compare compressed and full-context execution. On ContextBench, OpenAI MA exceeds SF for each displayed receiver. On ToolSandbox, the SF and MA endpoints often approach ceiling, and MA improves over RO. This differs from the default Qwen condition, where MA reduces accuracy. The comparison shows that the communication regime depends on the receiver and protocol as well as the benchmark.

Panel~(c) extends the closure diagnostic. Non-reasoning OpenAI receivers remain largely content-limited, while reasoning receivers continue to exhibit closure-sensitive behavior, the cross-family analog of Figure~\ref{fig:absorption}.

Panel~(d) adds weak-family baselines: at a smaller scale, matched MA sits below RO even on ContextBench, matching the sender-gap theorem when the extra gain fails to cover the protocol tax. Panel~(e) reports five-seed BFCL and ToolSandbox results at temperature $0.3$. Panel~(f) aligns all receivers on a single ContextBench axis, summarizing the preceding panels.

\FloatBarrier

\subsection{Protocol-Level Tax Knobs}
\label{app:tax-controller}

\paragraph{Calibration and execution.}
The early gate compares a receiver-sufficiency rate with a fixed threshold
of $0.80$. The rate is the mean joint-hit indicator on held-out RO records
for the deployed interface. The gate selects \texttt{skip} when the rate
meets or exceeds the threshold, with its configuration fixed before testing.
This path returns one receiver generation before any sender or scorer calls.

After the gate, the five-action path generates and caches two real message
candidates: \texttt{compress} summarizes upstream context under the message
budget, and \texttt{refresh} summarizes the updated shared state.
The \texttt{raw} condition appends full upstream context to the receiver
observation. \texttt{escalate} reuses the compressed message and directly
invokes the stronger receiver. Scoring uses five no-message and four
conditioned probes, each capped at four tokens. Each probe pair uses the
receiver that executes its action; conditioned probes use the actual
candidate input. The gain surrogate clips the conditioned-minus-unconditioned
leading-token log-probability difference at zero. The selected action reuses
its cached candidate for one final receiver generation.
Table~\ref{tab:selector-call-counts} includes both candidate generations,
all scoring probes, and the selected receiver call in each decision's cost.

The default calibration and routing split assigns 20\% of instances to
calibration or router training and 80\% to testing, grouped by
\texttt{instance\_id}. Repeated-split studies use 20 partitions with split
seeds 0--19 and a fixed receiver, interface, and action set. Metrics are
computed within each test split and averaged equally across the 20 splits.
Selector decoding defaults to temperature $0$.
Paired calibration records
provide an action-level protocol-burden offset. This offset can absorb gain
effects as well as protocol burden, so the per-instance gain and cost terms
supply the instance-specific ranking signal. The selector reads no
benchmark label, gold target, hit field, or realized correctness at test time.
The static score uses $\eta_c=10^{-4}$ on raw token counts and $\eta_r=0$
for its one-step horizon. The shared-state experiments evaluate residual
propagation and targeted refresh separately.

\paragraph{Selector ablation records.}
Table~\ref{tab:selector-term-ablation} reports the ContextBench selector
ablation for the OpenAI \texttt{gpt-4o-mini} receiver cell on the stated
appendix diagnostic support. Gain and tax removals are obtained from the stored
ablation file; the cost-removal row is recomputed from the same stored
RO/SF/MA records, with the cost coefficient set to zero. The file records
\texttt{lambda\_c}$=10^{-4}$, which is the code-level $\eta_c$ applied to raw
token counts. It disables finite-horizon residual propagation on this static
one-step condition, so $\eta_r=0$ and the residual-removal row equals the full
score. Cost is displayed as mean token count per step divided by the
always-send baseline in the same file.

\begin{table}[!htbp]
\centering
\caption{\textbf{Selector ablation on ContextBench.}
Rows are computed from stored OpenAI RO/SF/MA records for the
\texttt{gpt-4o-mini} receiver and \texttt{gpt-5.4} sender on the stated
diagnostic support. The comparison is an accuracy--cost frontier diagnostic:
always-send has the highest raw accuracy in this static cell, while the full
selector uses less than $0.54\times$ the always-send cost and remains above
always-skip.}
\label{tab:selector-term-ablation}
\small
\begin{tabular}{lrrrrrr}
\toprule
Selector variant & $n$ & Acc. & Cost / send & Skip & Compress & Raw \\
\midrule
Full score & 300 & 0.737 & 0.536 & 175 & 11 & 114 \\
No gain term & 300 & 0.650 & 0.402 & 300 & 0 & 0 \\
No tax term & 300 & 0.737 & 0.536 & 175 & 11 & 114 \\
No cost term & 300 & 0.730 & 0.550 & 175 & 0 & 125 \\
No residual term & 300 & 0.737 & 0.536 & 175 & 11 & 114 \\
Always skip & 300 & 0.650 & 0.402 & 300 & 0 & 0 \\
Always send & 300 & 0.787 & 1.000 & 0 & 300 & 0 \\
Always raw & 300 & 0.743 & 0.669 & 0 & 0 & 300 \\
\bottomrule
\end{tabular}
\end{table}

Removing the gain term makes the selector choose \texttt{skip} on every step, reducing accuracy from 0.737 to 0.650. Removing the tax term leaves every selected action unchanged on this evaluation set. The residual-removal row also matches the full score because the static configuration sets $\eta_r=0$. Removing the cost term changes the mixture from 11 compressed messages and 114 raw-context actions to 125 raw-context actions, with 175 skips; accuracy is 0.730. Always-send reaches the highest accuracy, 0.787, at the highest cost; the full selector uses 0.536 times that cost.

\paragraph{Static selector baselines.}
Table~\ref{tab:selector-static-baselines} compares the released four-action threshold controller with fixed policies on ToolSandbox and \(\tau\)-bench. Here, \texttt{skip} suppresses communication, \texttt{send} uses the bounded MA message, and \texttt{rewrite} forwards a normalized evidence bundle. The adaptive controller uses local sufficiency, parsing, budget pressure, and schema fidelity to choose an action. On ToolSandbox, it attains 1.000 accuracy at 581 input tokens per step, close to the skip policy's cost. On \(\tau\)-bench, rewriting achieves higher accuracy than the adaptive controller (0.139 versus 0.083) at nearly identical cost (2,506 versus 2,504 tokens).

\begin{table}[!htbp]
\centering
\caption{\textbf{Static selector diagnostics on protocol-sensitive benchmarks.}
Rows are aggregated from the stored selector-policy records on the stated
diagnostic supports. Token cost is realized total input tokens per step. These
baselines complement the ContextBench term ablation in
Table~\ref{tab:selector-term-ablation}.}
\label{tab:selector-static-baselines}
\small
\begin{tabular}{llrr}
\toprule
Benchmark & Policy & \texttt{hit\_joint} & Mean input tokens \\
\midrule
ToolSandbox & adaptive selector & 1.000 & 581 \\
ToolSandbox & always skip & 0.972 & 577 \\
ToolSandbox & always send & 0.806 & 2,474 \\
ToolSandbox & always rewrite & 0.833 & 1,930 \\
\midrule
\(\tau\)-bench & adaptive selector & 0.083 & 2,504 \\
\(\tau\)-bench & always skip & 0.056 & 711 \\
\(\tau\)-bench & always send & 0.056 & 3,087 \\
\(\tau\)-bench & always rewrite & 0.139 & 2,506 \\
\bottomrule
\end{tabular}
\end{table}

\begin{table}[!htbp]
\centering
\caption{\textbf{Representative protocol ablations on a harmful and a helpful benchmark.}
The rows compare \texttt{skip}, bounded \texttt{compress}, evidence rewriting, and suppression under rewritten prompts and per-knob cost accounting. The evidence-rewrite row forwards a compact, normalized evidence bundle; its input and token cost differ from the full-context SF protocol in Table~\ref{tab:main-results}.}
\label{tab:ablations}
\small
\begin{tabular}{@{}lccc@{\hspace{14pt}}ccc@{}}
\toprule
& \multicolumn{3}{c}{ToolSandbox} & \multicolumn{3}{c}{ContextBench} \\
\cmidrule(lr){2-4}\cmidrule(lr){5-7}
Condition & Acc. & $\Delta$ vs RO & Cost & Acc. & $\Delta$ vs RO & Cost \\
\midrule
Receiver-only & 0.889 & 0.000 & 1,374 & 0.633 & 0.000 & 7,251 \\
MA (default) & 0.653 & -0.236 & 2,995 & 0.775 & 0.142 & 16,801 \\
Prose & 0.556 & -0.333 & 2,905 & 0.694 & 0.061 & 4,183 \\
Evidence-rewrite (knob set) & --- & --- & --- & 0.778 & 0.144 & 1,853 \\
Canonical JSON & 0.750 & -0.139 & 3,004 & 0.764 & 0.131 & 4,162 \\
Budget $=$16 & 0.896 & 0.007 & 3,676 & 0.736 & 0.103 & 3,705 \\
Budget $=$32 & 1.000 & 0.111 & 3,717 & 0.833 & 0.200 & 3,729 \\
Selective suppression & 1.000 & 0.111 & 919 & 0.750 & 0.117 & 3,960 \\
\bottomrule
\end{tabular}
\end{table}

\begin{figure}[!htbp]
\centering
\safeincludegraphics[width=0.95\columnwidth]{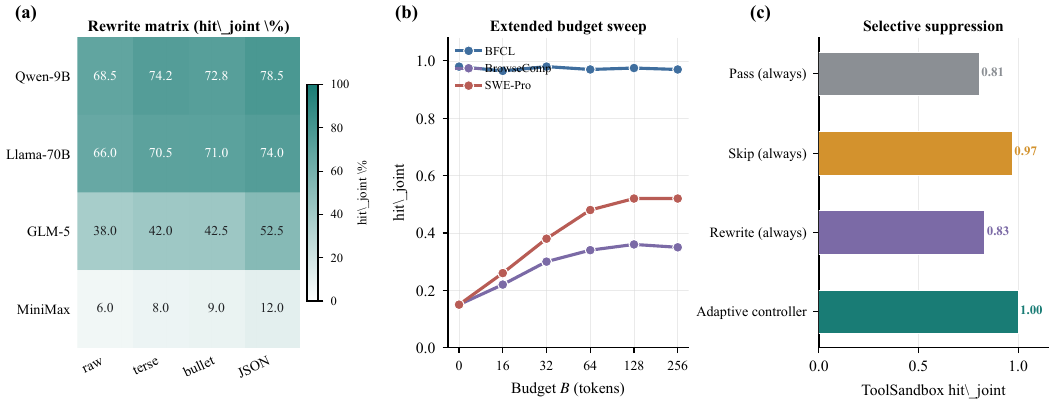}
\caption{\textbf{Expanded protocol-knob analyses.}
Panels report the grounded rewrite matrix, broader budget sweeps, and selective-suppression controls.}
\label{fig:tax-controller}
\end{figure}

Table~\ref{tab:ablations} reports concrete protocol changes, including \texttt{skip}, bounded \texttt{compress}, evidence rewriting, and selective suppression. The \texttt{raw} action retains its full-context meaning. Figure~\ref{fig:tax-controller} gives the corresponding format and budget sweeps, complementing the static and online comparisons in Figure~\ref{fig:tax-knobs-main}.

Theorem~\ref{thm:sufficiency-not-enough} motivates including \texttt{raw} in protocol comparisons. Rewriting and budget changes vary preserved content, decoder compatibility, and protocol tax.
\paragraph{Calibration sensitivity.}
The selector changed little across calibration fractions from 5\% to 40\%.
We evaluated 20 instance-disjoint splits per fraction with split seeds
0--19, for 100 splits in total. Mean accuracy spanned 0.90 pp and cost relative to always-send spanned
0.30 pp (Table~\ref{tab:selector-calibration}). Pooling the ContextBench and
ToolSandbox offset records without benchmark labels changed ContextBench
accuracy from 0.7425 to 0.7389 and preserved all 36/36 ToolSandbox
\texttt{skip} decisions. On the separate 300-step ablation support in
Table~\ref{tab:selector-term-ablation}, removing tax changed 0/300 choices,
while removing gain selected \texttt{skip} on all 300 steps and reduced
accuracy from 0.7367 to 0.6500. The full score used 1,735.0 tokens per step;
always-send reached 0.7867 accuracy at 3,235.4 tokens.

\begin{table}[!htbp]
\centering
\caption{\textbf{Selector sensitivity to calibration size.}
Each column averages 20 instance-disjoint calibration/test splits.}
\label{tab:selector-calibration}
\small
\setlength{\tabcolsep}{5pt}
\begin{tabular}{@{}lrrrrr@{}}
\toprule
Calibration fraction & 5\% & 10\% & 20\% & 30\% & 40\% \\
\midrule
Mean accuracy & 0.7368 & 0.7415 & 0.7425 & 0.7452 & 0.7458 \\
Cost / always-send & 0.5358 & 0.5366 & 0.5356 & 0.5372 & 0.5386 \\
\bottomrule
\end{tabular}
\end{table}
\paragraph{Learned routing comparison.}
The receiver-relative selector achieved the highest accuracy minus token
cost among the evaluated routing policies (Table~\ref{tab:learned-router-comparison}).
All policies used \texttt{skip}/\texttt{raw}/\texttt{compress} and the same
20 instance-disjoint calibration/test partitions described above.
The learned routers used receiver-local input
length and step index as test features. The selector reached 0.7425 accuracy
at 1,732.9 tokens per step; always-send reached 0.7917 accuracy but used
3,227.0 tokens. Utilities apply the same $10^{-4}$ cost weight.

\begin{table}[!htbp]
\centering
\caption{\textbf{ContextBench routing policies on a shared action set.}
Utility is accuracy minus $10^{-4}$ times mean tokens.}
\label{tab:learned-router-comparison}
\small
\setlength{\tabcolsep}{5pt}
\begin{tabular}{@{}lrrr@{}}
\toprule
Policy & Accuracy & Tokens/step & Utility \\
\midrule
Receiver-relative selector & 0.7425 & 1,732.9 & 0.5692 \\
Always raw & 0.7517 & 2,151.7 & 0.5365 \\
Multinomial logistic (always skip) & 0.6525 & 1,300.0 & 0.5225 \\
Gradient boosted & 0.6617 & 1,425.3 & 0.5191 \\
Learned centroid & 0.7292 & 2,181.5 & 0.5110 \\
Always send & 0.7917 & 3,227.0 & 0.4690 \\
\bottomrule
\end{tabular}
\end{table}
\paragraph{Reported MasRouter configuration.}
The MasRouter comparison matches sender and receiver identities, receiver
prompt, evaluation support, split seeds, message budget, decoding, and token
accounting with the receiver-relative selector. MasRouter uses an always-send
communication policy.
Table~\ref{tab:masrouter-comparison} gives its recorded accuracy and cost.

\begin{table}[!htbp]
\centering
\caption{\textbf{Separately reported MasRouter measurements.}
Token counts are per step; the utility column uses ContextBench accuracy and
the $10^{-4}$ token weight.}
\label{tab:masrouter-comparison}
\small
\setlength{\tabcolsep}{4pt}
\begin{tabular}{@{}lrrr rr@{}}
\toprule
& \multicolumn{3}{c}{ContextBench} & \multicolumn{2}{c}{ToolSandbox} \\
\cmidrule(lr){2-4}\cmidrule(l){5-6}
Policy & Accuracy & Tokens & Utility & Accuracy & Tokens \\
\midrule
MasRouter (always-send) & 0.6317 & 3,227 & 0.3090 & 0.7056 & 3,085 \\
\bottomrule
\end{tabular}
\end{table}
\paragraph{Executed cost and latency.}
The pre-sender gate reduced the executed ToolSandbox path to one receiver
generation. On the 36-step online support, it reduced recorded tokens by
70.2\% and latency by 31.0\% relative to always-send
(Table~\ref{tab:gate-runtime}). A five-seed endpoint comparison with 720
step-runs per policy reduced tokens by 57.9\% and latency by 26.0\%, with
711/720 correct SKIP outputs versus 710/720 for always-send. The online and
endpoint supports differ; the online slice was harder for always-send.

\begin{table}[!htbp]
\centering
\caption{\textbf{ToolSandbox runtime measurements.}
Recorded tokens include prompt and completion tokens on the executed path.
The 36-step input-only counts in Table~\ref{tab:selector-static-baselines}
are 581 for the selector and 2,474 for always-send. The endpoint study used
144 steps across five seeds per policy.}
\label{tab:gate-runtime}
\small
\setlength{\tabcolsep}{5pt}
\begin{tabular}{@{}lrrrr@{}}
\toprule
Policy & Step-runs & Accuracy & Tokens/step & Seconds/step \\
\midrule
Pre-sender gate & 36 & 1.000 & 919.4 & 6.60 \\
Always send & 36 & 0.8056 & 3,085.4 & 9.56 \\
SKIP endpoint & 720 & 0.9875 & 1,801.4 & 14.37 \\
Always send & 720 & 0.9861 & 4,282.4 & 19.42 \\
\bottomrule
\end{tabular}
\end{table}

\begin{table}[!htbp]
\centering
\caption{\textbf{Model calls for the gate and five-action selector.}
The gate precedes candidate construction. Each post-gate decision generates
and caches both message candidates, uses nine probes capped at four tokens,
and executes one receiver generation. \texttt{escalate} uses the stronger
receiver; other actions use the base receiver.}
\label{tab:selector-call-counts}
\small
\setlength{\tabcolsep}{5pt}
\begin{tabular}{@{}lrrr@{}}
\toprule
Action & Scorer probes & Sender calls & Receiver generations \\
\midrule
\texttt{skip} through gate & 0 & 0 & 1 \\
\texttt{skip} after scoring & 9 & 2 & 1 \\
\texttt{raw} & 9 & 2 & 1 \\
\texttt{compress} & 9 & 2 & 1 \\
\texttt{refresh} & 9 & 2 & 1 \\
\texttt{escalate} & 9 & 2 & 1 \\
\bottomrule
\end{tabular}
\end{table}

\FloatBarrier

\subsection{Conditional Propagation in Shared-State Tasks}
\label{app:shared-state}

\begin{figure}[!htbp]
\centering
\safeincludegraphics[width=0.95\columnwidth]{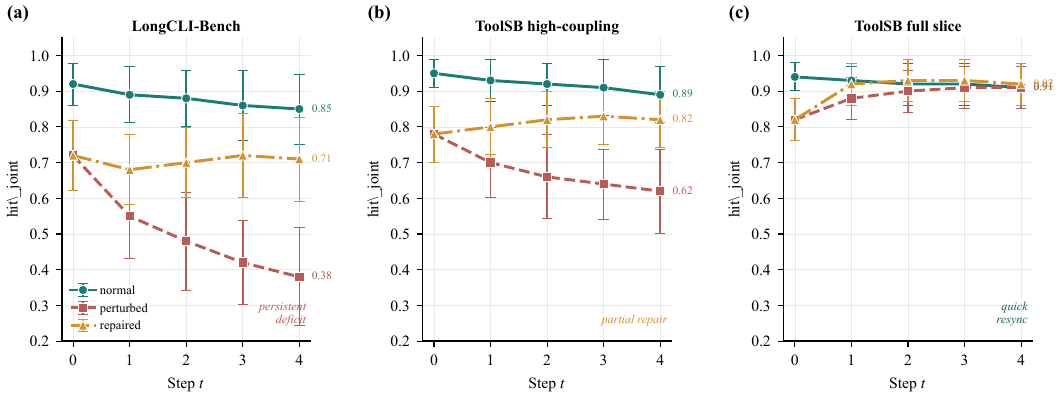}
\caption{\textbf{Conditional propagation in shared-state tasks.}
(a) LongCLI-Bench: perturbations produce a persistent downstream deficit, and repair only partially closes it.
(b) A higher-coupling ToolSandbox condition: downstream effects are visible, and repair helps but does not fully erase the gap.
(c) A broader ToolSandbox condition: the deficit is largely local, and the trajectory quickly resynchronizes.}
\label{fig:accumulation}
\end{figure}

\begin{figure}[!htbp]
\centering
\safeincludegraphics[width=0.95\columnwidth]{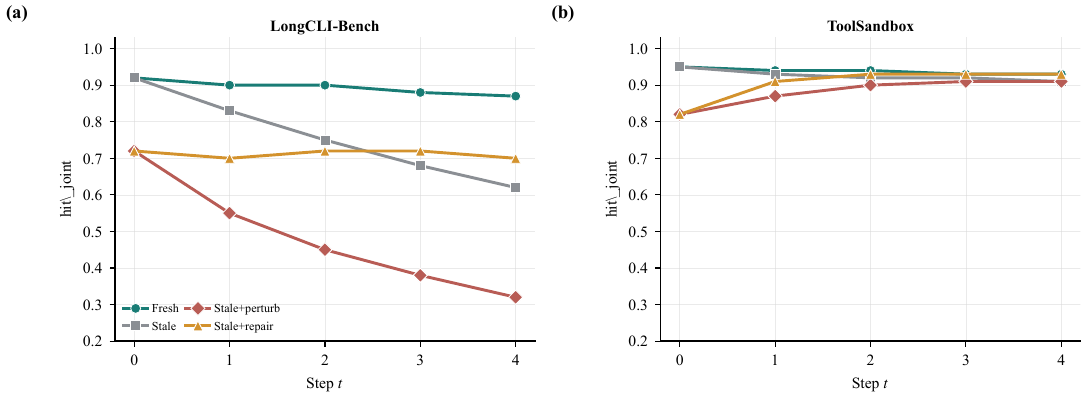}
\caption{\textbf{Fresh-observation ablation.}
(a) LongCLI-Bench: stale observations degrade downstream performance even without an injected perturbation, and targeted repair helps most under stale conditions.
(b) ToolSandbox: fresh observations largely neutralize downstream perturbation effects in broader conditions, reinforcing the resynchronization interpretation.}
\label{fig:stale-obs}
\end{figure}

These studies test the coupling dependence in Proposition~\ref{prop:conditional-propagation}. LongCLI and the higher-coupling ToolSandbox condition behave like positive-coupling settings: a local perturbation leaves a downstream deficit, and repair only partially closes it. The broader ToolSandbox condition behaves like a resynchronizing regime: the same intervention is mostly local, and the trajectory re-aligns once fresh observations arrive.

The fresh-observation ablation in Figure~\ref{fig:stale-obs} makes the same point from the opposite direction. If a benchmark continually refreshes the local view, residual coupling is weak, and the perturbation dissipates quickly. If not, the same local error may persist for several steps. This is why the main-text selector includes a residual-risk term rather than optimizing one-step utility alone.

\FloatBarrier
\clearpage

\section{Illustrative Case Studies}
\label{app:case-studies}

The following traces are extracted from actual benchmark runs. Each case shows the receiver's local observation, the sender's message (when present), and the resulting action. These records illustrate the mechanisms in the main text.

\begin{casebox}[label=case:noise]{Message overrides a correct local view (noise regime)}
\textbf{Benchmark:} ToolSandbox \quad \textbf{Task:} \texttt{find\_temperature} \quad \textbf{Step:} 1

\medskip
\textbf{Receiver's local observation:}\\
\texttt{Goal: What's the temperature here right now}\\
\texttt{Available tools: [search\_weather\_around\_lat\_lon]}\\
\texttt{Recent calls: [search\_weather invoked in Step 0]}

\medskip
\textbf{Recorded gold label:} \texttt{run\_test} / \texttt{search\_weather\_around\_lat\_lon}

\medskip
\textbf{Without sender (\texttt{receiver\_only}):} \texttt{run\_test} / \texttt{search\_weather\_around\_lat\_lon} \hfill \colorbox{green!15}{\textbf{Correct}}

\medskip
\textbf{Sender message (80B):}\\
\texttt{\{``action\_type'': ``read'', ``target\_artifact'': ``messages'',}\\
\texttt{~``rationale'': ``The weather tool was already invoked in Step 0;}\\
\texttt{~~now we must read the response, which should be stored in}\\
\texttt{~~messages as the tool's output.''\}}

\medskip
\textbf{With sender (\texttt{ma\_seq}):} \texttt{read} / \texttt{messages} \hfill \colorbox{red!15}{\textbf{Wrong}}

\medskip
\textbf{Mechanism:} The receiver-only output names the weather tool recorded as the target. The sender instead directs the receiver to \texttt{messages}, and the receiver adopts that target. This artifact mismatch identifies the harmful redirection independently of the trace's literal \texttt{run\_test} action label.
\end{casebox}

\begin{casebox}[label=case:positive]{Sender provides cross-module context the receiver lacks (helpful coordination)}
\textbf{Benchmark:} ContextBench \quad \textbf{Task:} Django cascade deletion bug \quad \textbf{Step:} 0

\medskip
\textbf{Bug:} \texttt{UnicodeDecodeError} during \texttt{.delete()} because cascade collection fetches fields with invalid UTF-8.

\medskip
\textbf{Without sender (\texttt{receiver\_only}):}\\
Action: \texttt{edit} / \texttt{django/db/models/base.py} \hfill \colorbox{red!15}{\textbf{Wrong file}}\\
{\small The receiver edits \texttt{base.py}, a plausible module that does not contain the cascade logic.}

\medskip
\textbf{Sender message (80B):}\\
\texttt{\{``action\_type'': ``edit'',}\\
\texttt{~``target\_artifact'': ``django/db/models/deletion.py'',}\\
\texttt{~``rationale'': ``The UnicodeDecodeError occurs because .delete()}\\
\texttt{~~fetches unnecessary fields (e.g., text\_log\_error.line) during}\\
\texttt{~~cascade collection. We must modify Collector.collect() to only}\\
\texttt{~~retrieve fields required for deletion.''\}}

\medskip
\textbf{With sender (\texttt{ma\_seq}):}\\
Action: \texttt{edit} / \texttt{django/db/models/deletion.py} \hfill \colorbox{green!15}{\textbf{Correct}}

\medskip
\textbf{Mechanism:} The receiver's local view does not reveal which module implements cascade collection. The sender, with access to the full repository context, identifies the correct file (\texttt{deletion.py}) and the specific function (\texttt{Collector.collect}). This is the helpful-coordination regime: the message transfers cross-module knowledge that the receiver cannot derive from its local observation alone.
\end{casebox}

\begin{casebox}[label=case:absorption]{Same message and different action compliance (conversion bottleneck)}
\textbf{Benchmark:} ContextBench \quad \textbf{Task:} Astropy mask propagation bug \quad \textbf{Step:} 0\\
\textbf{Protocol:} Same-message replay (frozen 80B sender output)

\medskip
\textbf{Frozen sender message:}\\
\texttt{\{``action\_type'': ``edit'',}\\
\texttt{~``target\_artifact'': ``astropy/nddata/mixins/ndarithmetic.py'',}\\
\texttt{~``rationale'': ``Mask propagation fails when one operand has no}\\
\texttt{~~mask (None) and handle\_mask=np.bitwise\_or is used. Must}\\
\texttt{~~pre-process masks to use zeros for missing masks.''\}}

\medskip
\textbf{Qwen-9B receiver output:}\\
\texttt{\{``action\_type'': ``edit'',}\\
\texttt{~``target\_artifact'': ``astropy/.../ndarithmetic.py''\}} \hfill \colorbox{green!15}{\textbf{Correct}}

\medskip
\textbf{GLM-5 receiver output:}\\
\texttt{\{``action\_type'': ``read'',}\\
\texttt{~``target\_artifact'': ``astropy/.../ndarithmetic.py'',}\\
\texttt{~``rationale'': ``Need to examine the current mask handling}\\
\texttt{~~implementation before making the fix.''\}} \hfill \colorbox{red!15}{\textbf{Wrong action}}

\medskip
\textbf{Decomposition:}\\
\begin{tabular}{lcc}
\toprule
& Qwen-9B & GLM-5 \\
\midrule
Artifact correct (\texttt{hit\_artifact}) & \cmark & \cmark \\
Action correct (\texttt{hit\_action}) & \cmark & \xmark \\
Joint correct (\texttt{hit\_joint}) & \cmark & \xmark \\
\bottomrule
\end{tabular}

\medskip
\textbf{Mechanism:} Both receivers process the identical frozen message. Both identify the correct target file. But GLM-5 emits \texttt{read} instead of \texttt{edit}. The information was recovered at the artifact level, but the action did not close correctly. This is why the fixed-message bottleneck in the main text appears as a conversion loss rather than as a failure of raw content recovery.
\end{casebox}

\section{Scope and Limitations}
\label{app:limitations}

Our account is step-level and protocol-relative: the framework makes a sharp prediction at this granularity and isolates the message-conditioning effect from trajectory-level confounders. Extending to end-to-end task success requires modeling cross-step error correlation; a natural extension would estimate fresh discrepancies and carry coefficients separately rather than folding them into one-step contrasts. The selector is analysis-guided; its gains are strongest when closure failure or protocol tax dominates, and richer online estimators, together with receiver-side adaptation, will extend its reach to other regimes. Paired MA, RO, and SF contrasts measure deployed protocol effects, and instance-clustered intervals quantify their uncertainty. The theoretical gain and tax depend additionally on the bounded decoder class; empirical calibration offsets combine these effects. Shared-state results hold when trajectory coupling is non-trivial; in loosely coupled settings, trajectories resynchronize regardless of intervention, a direct consequence of the dynamic framework.

\paragraph{Broader impacts.}
Reducing unnecessary sender calls can lower inference cost and latency, making multi-agent assistance more accessible. The same communication mechanisms can spread incorrect directives or expose sensitive context to additional model services. Our observed target overrides and action-closure failures identify concrete failure modes for deployments. Applications should limit shared context to task-relevant information and validate consequential actions against the user's request and available tool state. Lower communication cost also makes automated workflows easier to scale, including workflows used for harmful purposes; efficiency and task accuracy therefore need to be assessed alongside the intended use.

\newpage
\section*{NeurIPS Paper Checklist}

\begin{enumerate}

\item {\bf Claims}
    \item[] Question: Do the main claims made in the abstract and introduction accurately reflect the paper's contributions and scope?
    \item[] Answer: \answerYes{}
    \item[] Justification: The abstract and Section~\ref{sec:intro} state the step-level, receiver-relative framework, its formal results, and the selector; Sections~\ref{sec:theory}--\ref{sec:experiments} develop and test them, and Appendix~\ref{app:limitations} states the scope.

\item {\bf Limitations}
    \item[] Question: Does the paper discuss the limitations of the work performed by the authors?
    \item[] Answer: \answerYes{}
    \item[] Justification: Appendix~\ref{app:limitations}, referenced from Section~\ref{sec:conclusion}, discusses the step-level scope, the extension to end-to-end task success, the regimes where the analysis-guided selector helps most, and the dependence of gain and tax on the bounded decoder class.

\item {\bf Theory assumptions and proofs}
    \item[] Question: For each theoretical result, does the paper provide the full set of assumptions and a complete (and correct) proof?
    \item[] Answer: \answerYes{}
    \item[] Justification: Each result in Sections~\ref{sec:theory} and~\ref{sec:implications} states its assumptions locally (log-loss, proper Bayes-sufficient compression, bottleneck factorization, single crossing, shared carry sequences), and Appendix~\ref{app:theory} gives the complete proofs.

    \item {\bf Experimental result reproducibility}
    \item[] Question: Does the paper fully disclose all the information needed to reproduce the main experimental results of the paper to the extent that it affects the main claims and/or conclusions of the paper (regardless of whether the code and data are provided or not)?
    \item[] Answer: \answerYes{}
    \item[] Justification: Section~\ref{sec:exp-setup} and Appendix~\ref{app:exp-details} specify the benchmarks, sender and receiver models, message budget, paired contrasts, scorer targets, replay protocol, and selector calibration; Table~\ref{tab:run-bookkeeping} records the support used by each numerical family.

\item {\bf Open access to data and code}
    \item[] Question: Does the paper provide open access to the data and code, with sufficient instructions to faithfully reproduce the main experimental results, as described in supplemental material?
    \item[] Answer: \answerYes{}
    \item[] Justification: Code and the stored per-step records are released at \coderepo, with slice manifests and scripts that recompute the reported table cells from the records; the repository README lists the components it does not include, such as the full benchmark-running harness.

\item {\bf Experimental setting/details}
    \item[] Question: Does the paper specify all the training and test details (e.g., data splits, hyperparameters, how they were chosen, type of optimizer) necessary to understand the results?
    \item[] Answer: \answerYes{}
    \item[] Justification: The work is inference-only; Section~\ref{sec:exp-setup} gives the default models, message budget, contrasts, and metrics, and Appendices~\ref{app:exp-details} and~\ref{app:tax-controller} give per-benchmark supports, decoding settings, calibration splits, and selector weights.

\item {\bf Experiment statistical significance}
    \item[] Question: Does the paper report error bars suitably and correctly defined or other appropriate information about the statistical significance of the experiments?
    \item[] Answer: \answerYes{}
    \item[] Justification: Table~\ref{tab:paired-clustered} reports instance-clustered percentile bootstrap 95\% intervals (5,000 resamples) and exact McNemar tests for the paired effects; the OpenAI extension reports Wilson 95\% intervals (Tables~\ref{tab:openai-extension-counts} and~\ref{tab:openai-extension-counts-continued}), and the sender threshold reports a bootstrap interval (Section~\ref{sec:exp-claim3}).

\item {\bf Experiments compute resources}
    \item[] Question: For each experiment, does the paper provide sufficient information on the computer resources (type of compute workers, memory, time of execution) needed to reproduce the experiments?
    \item[] Answer: \answerYes{}
    \item[] Justification: All experiments are inference-only with no model training; open-weight Qwen models were served on a 2$\times$NVIDIA H200 node, OpenAI models were queried through the API at a cumulative cost of approximately 300 USD, and Appendix~\ref{app:exp-details} reports the evaluated step counts that determine reproduction cost.
    
\item {\bf Code of ethics}
    \item[] Question: Does the research conducted in the paper conform, in every respect, with the NeurIPS Code of Ethics \url{https://neurips.cc/public/EthicsGuidelines}?
    \item[] Answer: \answerYes{}
    \item[] Justification: The paper evaluates communication protocols on existing public benchmarks through model APIs and open-weight models; it involves no human subjects, deception, private data, or high-risk artifacts.

\item {\bf Broader impacts}
    \item[] Question: Does the paper discuss both potential positive societal impacts and negative societal impacts of the work performed?
    \item[] Answer: \answerYes{}
    \item[] Justification: The broader-impacts paragraph in Appendix~\ref{app:limitations} discusses lower inference cost and latency as well as the risks of propagating incorrect directives, exposing sensitive context, and scaling harmful workflows, together with mitigations.
    
\item {\bf Safeguards}
    \item[] Question: Does the paper describe safeguards that have been put in place for responsible release of data or models that have a high risk for misuse (e.g., pre-trained language models, image generators, or scraped datasets)?
    \item[] Answer: \answerNA{}
    \item[] Justification: The paper releases no pretrained model, generative system, or scraped dataset; the released records are model outputs on existing public benchmarks.

\item {\bf Licenses for existing assets}
    \item[] Question: Are the creators or original owners of assets (e.g., code, data, models), used in the paper, properly credited and are the license and terms of use explicitly mentioned and properly respected?
    \item[] Answer: \answerYes{}
    \item[] Justification: The six benchmark families are cited at first use in Section~\ref{sec:exp-setup} and used under their published terms for research; models are identified by their public names and accessed through their providers' APIs or open-weight releases under the corresponding terms of use.

\item {\bf New assets}
    \item[] Question: Are new assets introduced in the paper well documented and is the documentation provided alongside the assets?
    \item[] Answer: \answerYes{}
    \item[] Justification: The released code and per-step records are documented in the repository README and data card and are distributed under the MIT License.

\item {\bf Crowdsourcing and research with human subjects}
    \item[] Question: For crowdsourcing experiments and research with human subjects, does the paper include the full text of instructions given to participants and screenshots, if applicable, as well as details about compensation (if any)? 
    \item[] Answer: \answerNA{}
    \item[] Justification: The work does not involve crowdsourcing or research with human subjects.

\item {\bf Institutional review board (IRB) approvals or equivalent for research with human subjects}
    \item[] Question: Does the paper describe potential risks incurred by study participants, whether such risks were disclosed to the subjects, and whether Institutional Review Board (IRB) approvals (or an equivalent approval/review based on the requirements of your country or institution) were obtained?
    \item[] Answer: \answerNA{}
    \item[] Justification: The work does not involve human subjects, so IRB approval is not required.

\item {\bf Declaration of LLM usage}
    \item[] Question: Does the paper describe the usage of LLMs if it is an important, original, or non-standard component of the core methods in this research? Note that if the LLM is used only for writing, editing, or formatting purposes and does \emph{not} impact the core methodology, scientific rigor, or originality of the research, declaration is not required.
    \item[] Answer: \answerYes{}
    \item[] Justification: LLMs are the object of study: the experiments measure communication between sender and receiver LLMs from the Qwen, OpenAI, DeepSeek, Llama, GLM, and MiniMax families, whose identities, decoding settings, and prompts are reported in Section~\ref{sec:exp-setup} and Appendix~\ref{app:exp-details}.

\end{enumerate}

\end{document}